\documentclass[]{grounding-template}
\input{compile_compat.tex}
\setcitestyle{sort}
\usepackage[shortlabels,inline]{enumitem} %
\usepackage{amsmath}
\usepackage{amssymb}

\usepackage{etoc}
\usepackage{wrapfig}

\usepackage{needspace}
\newsavebox{\ArxivNaturalTable}
\newcount\ArxivPreviousLines
\newcount\ArxivWrapSteps
\makeatletter
\newcommand{\FinishArxivWrap}{\par
  \ArxivWrapSteps=0
  \loop\ifnum\c@WF@wrappedlines>1
    \ArxivPreviousLines=\c@WF@wrappedlines
    \noindent\strut\par
    \advance\ArxivWrapSteps by 1
    \ifnum\c@WF@wrappedlines<\ArxivPreviousLines\else\WFclear\fi
    \ifnum\ArxivWrapSteps>80\WFclear\fi
  \repeat\WFclear}
\makeatother

\makeatletter
\newcommand{\CompactContents}{%
  \small\setlength{\parskip}{0pt}%
  \renewcommand*\l@section[2]{\@dottedtocline{1}{0em}{1.6em}{\bfseries ##1}{##2}}%
  \renewcommand*\l@subsection[2]{\@dottedtocline{2}{1.6em}{2.5em}{##1}{##2}}%
  \renewcommand*\l@subsubsection[2]{\@dottedtocline{3}{4.1em}{3.3em}{##1}{##2}}%
}
\makeatother

\makeatletter
\patchcmd{\WF@putfigmaybe}
  {\ifdim\dimen@>\@tempdimb\global\WF@floatfalse\pagebreak\fi}
  {\ifWF@float\else\pagebreak\fi}
  {}{\PackageError{arxiv-layout}{Could not install wrap fit guard}{}}
\apptocmd{\WF@startwrapping}{\ifinner\else\suppressfloats\fi}{}{}
\apptocmd{\@floatplacement}{\ifvoid\WF@box\else\global\@topnum\z@\global\@botnum\z@\fi}{}{}
\newcommand{\FinishPlacedArxivWrap}{\ifvoid\WF@box\FinishArxivWrap\fi}
\makeatother

\newif\ifArxivBlockPending
\newdimen\ArxivBlockRoom
\newdimen\ArxivBlockHeight
\makeatletter
\apptocmd{\@floatplacement}{\ifArxivBlockPending\global\@topnum\z@\global\@botnum\z@\global\@colnum\z@\fi}{}{}
\newcommand{\PlaceArxivInlineBox}{%
  \par\global\ArxivBlockPendingtrue\suppressfloats
  \ArxivBlockHeight=\dimexpr\ht\ArxivInlineBox+\dp\ArxivInlineBox+\baselineskip\relax
  \ArxivBlockRoom=\@colroom\advance\ArxivBlockRoom-\pagetotal
  \ifdim\ArxivBlockHeight>\ArxivBlockRoom\newpage\fi
  \suppressfloats\noindent\usebox{\ArxivInlineBox}\par
  \global\ArxivBlockPendingfalse
}
\makeatother
\newsavebox{\ArxivInlineBox}

\newcommand{\FrontContents}{%
  \etocdepthtag.toc{main}%
  \phantomsection\addcontentsline{toc}{section}{Abstract}%
  {\setlength{\parindent}{0pt}%
    \section*{Abstract}\abstractlist\par}%
  {\CompactContents
    \etocsettagdepth{main}{subsubsection}%
    \etocsettagdepth{appendix}{none}%
    \etocsettocstyle{\section*{Contents}}{}%
    \tableofcontents}%
  \clearpage
  {\CompactContents
    \etocsettagdepth{main}{none}%
    \etocsettagdepth{appendix}{subsubsection}%
    \etocsettocstyle{\section*{Appendix Contents}}{}%
    \tableofcontents}%
  \clearpage
}

\newcommand{\question}[2]{
    \begin{tcolorbox}[
        enhanced,
        frame hidden,
        colback=blue!5,
        borderline west={2pt}{0pt}{blue!70!black},
        sharp corners,
        boxsep=0pt,
        left=8pt,
        right=4pt,
        top=8pt,
        bottom=8pt,
        before skip=12pt plus 2pt minus 2pt,
        after skip=8pt plus 2pt minus 2pt,
    ]
        \phantomsection\label{question:#1}%
        \noindent\textcolor{blue!70!black}{\textbf{\textit{Question #1:}}} #2
    \end{tcolorbox}
}

\newcommand{\finding}[2]{
    \begin{tcolorbox}[
        colback=blue!5,
        colframe=blue!70!black,
        arc=5pt,
        boxsep=5pt,
        left=2pt,
        right=2pt,
        top=2pt,
        bottom=2pt,
        boxrule=0.8pt,
        drop shadow=gray!50!white,
        enhanced jigsaw,
        before skip=12pt plus 2pt minus 2pt, %
        after skip=12pt plus 2pt minus 2pt,  %
    ]
        \phantomsection\label{finding:#1}%
        \noindent\textbf{\textit{Finding #1:}} #2
    \end{tcolorbox}
}

\newcommand{\takeaway}[2]{
    \begin{tcolorbox}[
        colback=blue!5,
        colframe=blue!70!black,
        arc=5pt,
        boxsep=5pt,
        left=2pt,
        right=2pt,
        top=2pt,
        bottom=2pt,
        boxrule=0.8pt,
        drop shadow=gray!50!white,
        enhanced jigsaw,
        before skip=12pt plus 2pt minus 2pt, %
        after skip=12pt plus 2pt minus 2pt,  %
    ]
        \phantomsection\label{takeaway:#1}%
        \noindent\textbf{\textit{Takeaway #1:}} #2
    \end{tcolorbox}
}
\title{GroundAnything: Reconciling Parallel Decoding with Precise Visual Grounding at Flash Speed}

\author[1,2,*,\ddagger]{Qize~Yu}
\author[1,*]{Lianrui~Fan}
\author[2,*]{Bowen~Ping}
\author[1,*]{Xini~Ding}
\author[1,2,*]{Zetian~Song}
\author[2]{Junbo~Niu}
\author[3]{Kaixuan~Wang}
\author[3]{Tianxing~Chen}
\author[2]{Yue~Chen}
\author[2]{Minghua~He}
\author[2,6]{Yuran~Wang}
\author[2]{Jie~Huang}
\author[1]{Haojun~Zhang}
\author[1]{Min~Chen}
\author[1]{Hao~Li}
\author[7]{Wenxuan~Song}
\author[4]{Ruihai~Wu}
\author[1]{Xianming~Liu}
\author[5]{Shilong~Liu}
\author[1]{Shuchang~Zhou}
\author[3,\dagger]{Ping~Luo}
\author[1,\dagger]{Shiyu~Huang}

\affiliation[1]{XPeng Inc.}
\affiliation[2]{Peking University}
\affiliation[3]{The University of Hong Kong}
\affiliation[4]{University of California, Berkeley}
\affiliation[5]{Princeton University}
\affiliation[6]{National University of Singapore}
\affiliation[7]{HKUST (GZ)}
\contribution[*]{Equal contribution}
\contribution[\ddagger]{Project lead}
\contribution[\dagger]{Corresponding authors}

\project{\href{https://groundingpi.github.io/groundanything/}{\sffamily \fontsize{8.8pt}{11pt}\selectfont \texttt{https://groundingpi.github.io/groundanything/}}}
\code{\href{https://github.com/groundingpi/GroundAnything}{\sffamily \fontsize{8.8pt}{11pt}\selectfont \texttt{https://github.com/groundingpi/GroundAnything}}}
\model{\mbox{\href{https://huggingface.co/GroundingPI/GroundAnything}{\sffamily \fontsize{7pt}{11pt}\selectfont \texttt{https://huggingface.co/GroundingPI/GroundAnything}}\hspace{5pt}\href{https://huggingface.co/GroundingPI/GroundAnything-VLM}{\sffamily \fontsize{7pt}{11pt}\selectfont \texttt{https://huggingface.co/GroundingPI/GroundAnything-VLM}}}}

\titleteaser{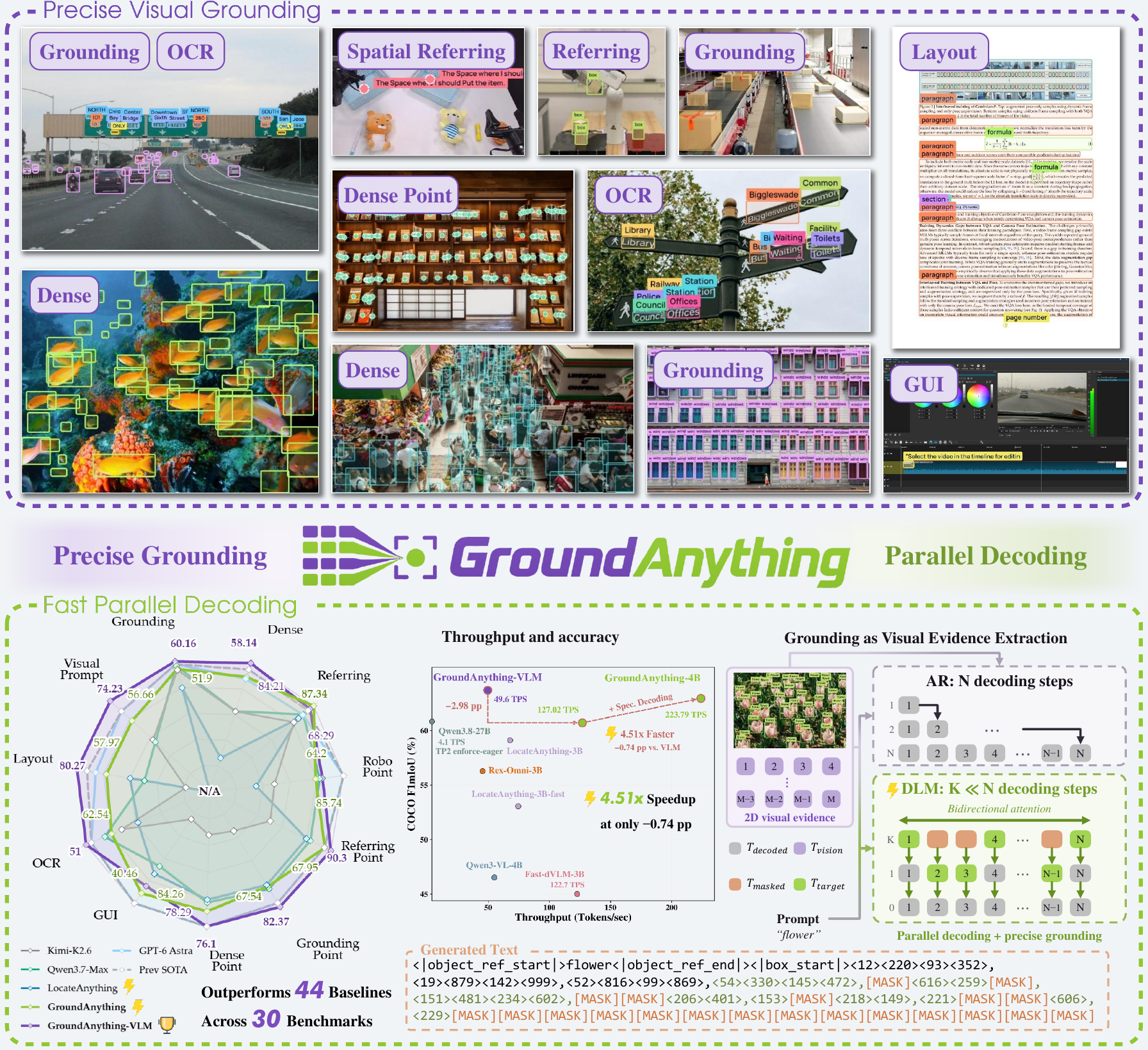}

\newlength{\titleauthorstoteaserskip}
\newlength{\titleteasertocaptionskip}
\newlength{\titlecaptiontoruleskip}
\abstract{Autoregressive (AR) grounding models serialize spatial predictions, introducing sequential latency and imposing a causal order on output tokens. We view grounding as visual evidence extraction: objects, locations, and spatial relations are jointly constrained by the image and query, yet their dependencies do not imply an intrinsic left-to-right generation order. This distinction makes bidirectional diffusion a natural fit, allowing spatial hypotheses to emerge in parallel and be jointly refined through iterative denoising. We introduce GroundAnything, a 4B-parameter grounding foundation model that reconciles fast parallel decoding with precise localization through blockwise denoising. Training combines grounding pretraining from public datasets and dedicated data engines, direct AR-to-diffusion conversion with joint AR and diffusion objectives, supervised fine-tuning, and GRPO-based reinforcement post-training. Across 30 grounding benchmarks, our autoregressive variant, GroundAnything-VLM, establishes a new overall state of the art among similarly sized models at 72.42\%, remaining competitive with GPT-6 Astra (71.35\%). With entropy-guided decoding, GroundAnything also surpasses the prior state of the art at this scale, averaging 61.75\% versus 53.32\% for the fast MTP-based LocateAnything model. We further explore decoding strategies, showing that an optional self-speculative mode achieves a 4.51$\times$ speedup over the AR counterpart with a 0.74 percentage-point drop in COCO F1mIoU. Infrastructure experiments show that progressive inference optimizations translate parallel decoding into practical speedups. These gains support efficient visual grounding in latency-sensitive real-world systems.

}

\begin{document}
\makeatletter
\renewcommand{\mymaketitle}{%
  \tcbset{enhanced,frame hidden}%
  \tcbset{left=0.5cm}%
  \tcbset{right=0.5cm}%
  \tcbset{top=0.26cm}%
  \tcbset{bottom=0.26cm}%
  \tcbset{arc=10pt}%
  \tcbset{colback=templatebackground}%
  \tcbset{before skip=0pt}%
  \tcbset{grow to left by=1.5pt}%
  \tcbset{grow to right by=1.5pt}%
  \begin{tcolorbox}
    \hypersetup{linkcolor=templatetitle,citecolor=templatetitle,urlcolor=templatetitle}%
    \setlength{\parindent}{0cm}%
    \setlength{\parskip}{0.5cm}%
    {\setlength{\parskip}{0cm}%
      \raggedright
      \nohyphens
      {\setstretch{1.618}\titlelist\par}%
      \vskip 0.12cm
      \authorlist\par
      \vskip 0.12cm
      \affiliationlist\par
      \contributionlist\par
    }%
    \ifdefempty{\teaserfile}{}{%
      {\setlength{\parskip}{0pt}%
        {\centering
          \includegraphics[width=\linewidth,page=1]{figures/teaser.pdf}\par
          {\captionsetup{type=figure,skip=5pt}%
            \caption{\textbf{GroundAnything}: broad, precise grounding with fast parallel decoding. \textbf{Top}: diverse tasks challenge localization in dense scenes and complex layouts. \textbf{Bottom}: given a 2D image and prompt, bidirectional diffusion extracts visual evidence into 1D structured grounding tokens in parallel. Self-speculative decoding achieves a 4.51$\times$ speedup over GroundAnything-VLM with only a 0.74 pp COCO F1mIoU drop.}
            \label{fig:teaser}}%
        }%
        \templatedashedrule
      }%
    }%
    \ifdefempty{\metadatalist}{}{%
      \vskip 0.04cm
      \noindent
      \begin{minipage}[c]{\linewidth}
        \setlength{\parskip}{0cm}%
        \raggedright
        \metadatalist
      \end{minipage}\par
      \vskip 0.04cm
    }%
  \end{tcolorbox}%
  \tcbset{reset}%
  \FloatBarrier
}

\makeatother
\maketitle
\clearpage
\FrontContents

\newtcolorbox{introquestions}{
    enhanced,
    frame hidden,
    colback=blue!5,
    borderline west={2pt}{0pt}{blue!70!black},
    sharp corners,
    boxsep=0pt,
    left=8pt,
    right=8pt,
    top=6pt,
    bottom=4pt,
    before skip=7pt plus 2pt minus 1pt,
    after skip=10pt plus 2pt minus 2pt,
}
\newcommand{\introquestion}[2]{%
    \noindent\textcolor{blue!70!black}{\textbf{\textit{Question #1:}}} #2\par\vskip3pt\relax%
}

\section{Introduction}
\label{sec:intro}
Visual grounding turns visual inputs into structured spatial predictions across object detection, referring expression comprehension, OCR, GUI interaction, and embodied perception. Next-token prediction has become a mainstream paradigm for generative grounding, with models such as Pix2Seq, Florence-2, and Rex-Omni expressing visual predictions as structured token sequences \citep{rexomni,chen2022pix2seqlanguagemodelingframework,xiao2024florence}. However, autoregressive (AR) decoding serializes labels and coordinates, introducing sequential latency and restricting output interactions to the generated prefix \citep{chen2022pix2seqlanguagemodelingframework,cheng2024parallel}. Dense scenes and complex layouts make these limitations especially costly for real-time grounding.

To reduce AR decoding latency, LocateAnything uses multi-token prediction (MTP) for parallel box decoding \citep{locateanything}. Diffusion VLMs support parallel generation and bidirectional refinement, with advances in multimodal understanding \citep{you2026llada,FastdVLM}, GUI grounding \citep{kumbhar2026guiagentsvisionlanguagediffusion}, and document OCR \citep{dong2026mineru}. Yet these grounding applications remain domain-specific, leaving diffusion's broader potential for unified structured prediction across diverse visual tasks largely untapped.

To unlock this potential and reconcile fast parallel decoding with broad and precise visual grounding, we start from a familiar observation: we may notice objects or read separate signs in different orders, yet reach the same judgment about what is present and where. We therefore view grounding as structured visual evidence extraction: spatial variables are jointly constrained by the image and query, without an intrinsic left-to-right generation order. Bidirectional masked diffusion naturally supports this process, allowing hypotheses to emerge in parallel and be refined using shared visual evidence and partially recovered structure (\Cref{fig:teaser}).

We introduce \textbf{GroundAnything}, a 4B-parameter grounding foundation model combining broad perceptual capabilities with blockwise parallel decoding. A shared vocabulary with quantized coordinates unifies diverse grounding tasks. Its decoder combines bidirectional attention within blocks with causal dependencies across blocks, enabling parallel prediction and KV-cache reuse \citep{arriola2025block}.

Grounding pretraining draws on public datasets and dedicated data engines. We build GroundAnything-VLM and GroundAnything from a shared pretrained checkpoint, applying AR-to-diffusion conversion with joint AR and diffusion objectives to the latter. Both variants then undergo supervised fine-tuning and GRPO-based reinforcement post-training.

Across 30 grounding benchmarks, GroundAnything-VLM establishes a new overall state of the art among similarly sized models at 72.42\%, remaining competitive with GPT-6 Astra (71.35\%). With entropy-guided decoding, GroundAnything is, to our knowledge, the first diffusion language model to surpass the prior overall AR state of the art at this scale, averaging 61.75\% at over 2$\times$ the speed of GroundAnything-VLM. We use this mode throughout the main benchmarks to evaluate pure diffusion decoding without AR verification. 	Additional ablations examine key model and training design choices throughout the study.

We further conduct a systematic speed analysis of decoding strategies and their trade-offs, including comparisons with MTP-based generation \citep{locateanything}. An optional self-speculative mode combines diffusion drafting with AR verification and improves both speed and accuracy over entropy-guided decoding on COCO, achieving a 4.51$\times$ speedup over GroundAnything-VLM with only a 0.74 pp F1mIoU drop. Systematic infrastructure experiments with SGLang, CUDA Graph, and FP8 demonstrate further practical speedups through progressive inference optimizations, supporting efficient visual grounding in real-world applications.

In summary, our major contributions are as follows:
\begin{itemize}[leftmargin=1.5em]

\item \textbf{Unlocking diffusion for broad visual grounding.} We introduce GroundAnything, a 4B foundation model that unifies diverse grounding tasks with precise localization and parallel decoding, trained through grounding pretraining, AR-to-diffusion conversion with joint objectives, supervised fine-tuning, and GRPO-based reinforcement learning.

\item \textbf{State-of-the-art grounding at 4B scale.} Across 30 benchmarks, GroundAnything surpasses the prior overall state of the art among similarly sized AR models and outperforms the larger Qwen3.7-Max. GroundAnything-VLM establishes a new overall state of the art at this scale and remains competitive with GPT-6 Astra.

\item \textbf{Systematic acceleration studies.} We analyze decoding strategies in detail, compare with MTP-based generation, and evaluate progressive infrastructure optimizations for practical deployment.

\end{itemize}

\section{Related Work}
\label{sec:related_work}

\subsection{Visual Grounding}

Object detectors such as YOLO establish efficient localization within predefined category vocabularies \citep{yolo}. Grounded vision--language pretraining extends detection to language-conditioned, open-set localization \citep{glip,groundingdino}. Generative approaches formulate detection as autoregressive next-token prediction \citep{chen2022pix2seqlanguagemodelingframework} and integrate spatial grounding into multimodal language modeling \citep{peng2023kosmos2groundingmultimodallarge}. Structured generation unifies diverse perception tasks \citep{xiao2024florence}, while quantized coordinates, large-scale grounding data, and reinforcement post-training strengthen localization \citep{rexomni}. However, AR generation serializes spatial predictions, increasing decoding costs for dense outputs. MTP-based parallel box decoding reduces this overhead by treating boxes and points as atomic generation units \citep{locateanything}. This progression motivates exploring alternative parallel generation mechanisms for broad and precise grounding.

\subsection{Diffusion Vision-Language Models}

Masked diffusion offers a complementary route to parallel generation through bidirectional conditioning and iterative token recovery. Diffusion VLMs demonstrate its potential for visual instruction following and multimodal understanding \citep{you2026llada,li2026lavida}. Direct AR-to-diffusion conversion adapts pretrained VLMs for efficient blockwise generation \citep{FastdVLM}. Beyond general understanding, specialized applications include GUI grounding with geometry-aware masking \citep{kumbhar2026guiagentsvisionlanguagediffusion} and structured document OCR with blockwise denoising \citep{dong2026mineru}. These studies demonstrate the promise of diffusion in both general multimodal understanding and specialized visual tasks. Nevertheless, a unified diffusion grounding foundation that preserves precise localization across heterogeneous tasks remains underexplored. GroundAnything addresses this gap through grounding-focused training and a shared structured output interface, combining broad perceptual capabilities with efficient parallel decoding.

\subsection{Parallel Decoding}

Prior work explores blockwise generation \citep{arriola2025block}, adaptive parallel sampling \citep{fastdllm,ben2026accelerated}, and diffusion-based speculative decoding \citep{liu2026tidar,FastdVLM}. Further work emphasizes practical acceleration through optimized inference infrastructure \citep{WeDLM}. Motivated by these advances, we examine decoding choices for GroundAnything in detail, studying their speed--accuracy trade-offs and extending the analysis to comparisons with MTP-based generation and progressive infrastructure optimizations.

\section{Method}
\label{sec:method}

\subsection{Problem Formulation: Grounding as Structured Visual Evidence Extraction}
\label{sec:problem_formulation}

Given an image $I$ and a task query $Q$, grounding extracts semantic identifiers and their spatial evidence, represented by boxes, points, or ordered point sequences. We serialize this evidence as $Y=(y_1,\ldots,y_L)$ using a shared vocabulary of text, structural markers, and quantized coordinates~\citep{rexomni}; input/output conventions are detailed in \Cref{app:groundanything_protocol}.

These variables are jointly constrained by the image and query, but need not be recovered in a fixed left-to-right order (\Cref{fig:teaser}). We therefore use blockwise masked diffusion~\citep{arriola2025block,FastdVLM,you2026affordancewamaffordanceawarejointworldaction,yu2026affordancevlavisionlanguageactionmodelempowering}, modeling
\begin{equation}
  p_\theta(Y\mid I,Q)=\prod_{k=1}^{K}p_\theta\!\left(Y^{(k)}\mid I,Q,Y^{(<k)}\right)
\end{equation}
over $K$ consecutive response blocks. Bidirectional attention supports parallel token recovery within each block, while completed blocks form a causal prefix. Semantic ordering, including trajectory time order, is preserved.

\subsection{Model Architecture}
\label{sec:model_architecture}

\begin{figure}[htbp]
  \centering
  \includegraphics[width=1.0\linewidth]{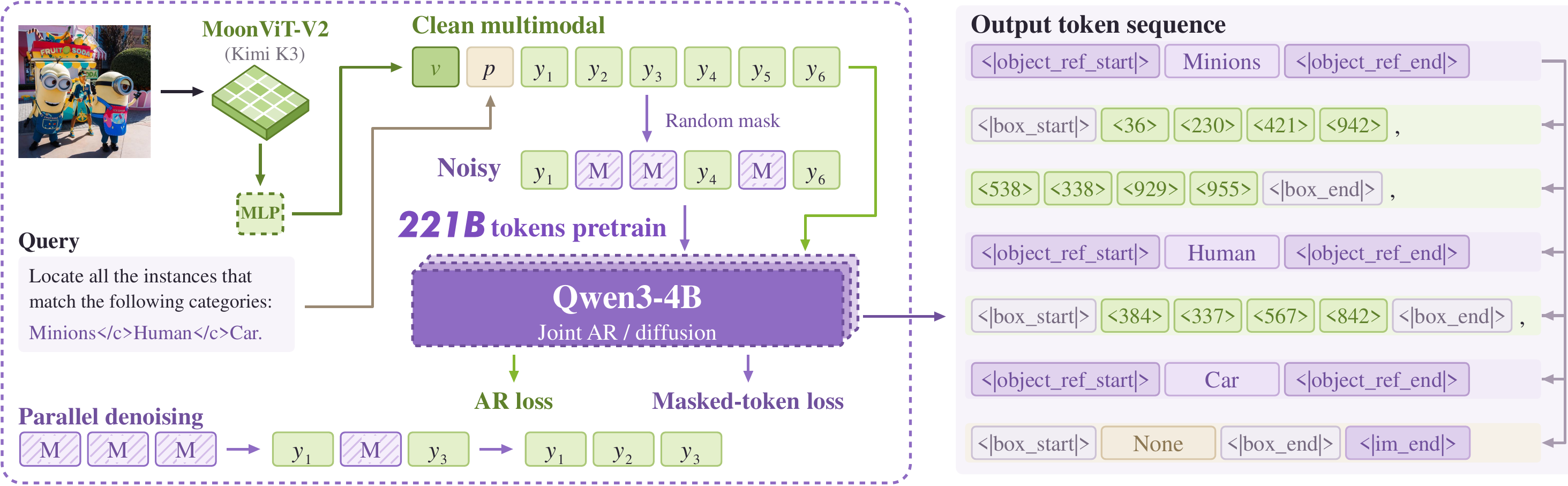}
  \caption{\textbf{GroundAnything architecture.} Clean multimodal inputs and response-corrupted text views share one visual encoding and jointly supervise causal prediction and masked denoising. One noisy view is illustrated, with its uncorrupted prompt copy omitted for clarity. The bottom-left panel illustrates parallel denoising, and the grounding example includes two minions and one human and an absent car category. Architecture specifications and parameter counts are provided in \Cref{app:groundanything_architecture}.}
  \label{fig:groundanything-architecture}
\end{figure}

GroundAnything is a diffusion grounding model obtained by directly converting GroundAnything-VLM, an autoregressive grounding model trained through multimodal and spatial pretraining, supervised fine-tuning, and reinforcement post-training. It combines a MoonViT-V2 (Kimi K3) visual encoder~\citep{KimiK3}, a projector with $2\times2$ spatial aggregation and a two-layer MLP, and a Qwen3-4B language decoder~\citep{Qwen34BLLM} (\Cref{fig:groundanything-architecture}).
Its variable-length output interface uses a shared vocabulary for semantic labels, protocol markers, and 1,000 quantized coordinate tokens (\texttt{<0>}--\texttt{<999>}); absent queried categories receive a None payload. The conversion introduces a mask token and jointly trains causal prediction and blockwise denoising with $B=32$; both modes share the decoder and vocabulary head, enabling parallel grounding and optional self-speculative decoding (\Cref{sec:ar_to_diffusion}).

\subsection{GroundAnything Data}
\label{sec:groundanything_data}

\label{sec:public_datasets}
\label{sec:data_engines}

Training data combines public datasets with annotations produced by our data engine.

\begin{figure}[htbp]
  \centering
  \includegraphics[width=1.0\linewidth]{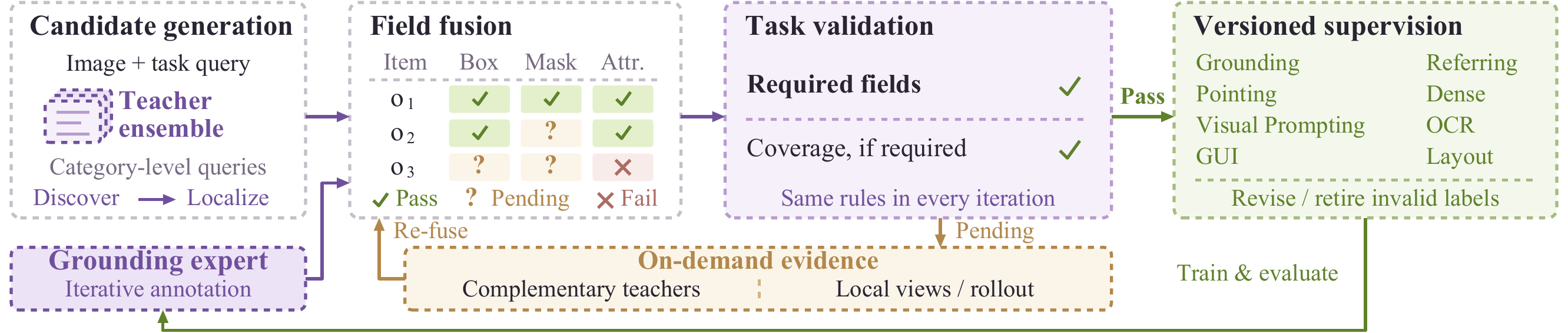}
  \caption{\textbf{Data engine.} Field-level teacher fusion and task-specific validation guide expert iteration, with targeted observations resolving uncertain annotations.}
  \label{fig:data-engine}
  \label{fig:public-datasets}
\end{figure}

Our data engine (\Cref{fig:data-engine}) fuses multi-teacher annotations at the field level and applies task-specific validation. Accepted labels train a unified grounding expert for iterative annotation; complementary teachers and local observations resolve uncertain cases. This process expands supervision while refining existing labels. Field validity and query-level coverage are checked separately, so an empty result alone does not establish absence. Fusion, validation, and expert iteration are detailed in \Cref{app:groundanything_data_engine}.

\FloatBarrier
\subsection{Training Design}
\label{sec:training_design}
\subsubsection{Base VLM Training}
\label{sec:base_vlm_training}

Base pretraining has three stages. \emph{Vision--language connector alignment} trains the projector on image--text pairs with both backbones frozen. \emph{Joint multimodal pretraining} updates all modules using text and multimodal supervision. \emph{General visual and video understanding} develops instruction following and image/video understanding. This autoregressive foundation supports subsequent spatial specialization through supervised fine-tuning (SFT) and reinforcement post-training. GroundAnything additionally undergoes the conversion below; training settings are provided in \Cref{app:groundanything_vlm_training,app:groundanything_training}.

\subsubsection{AR-to-Diffusion Conversion}
\label{sec:ar_to_diffusion}

\FinishArxivWrap
\begin{wrapfigure}{r}{0.4\textwidth}

  \centering
  \includegraphics[width=\linewidth]{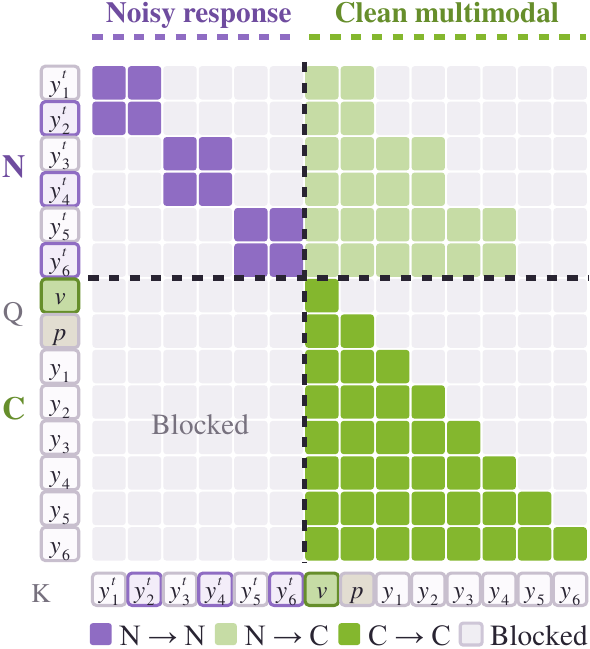}
  \caption{\textbf{Conversion attention mask.} Rows/columns are queries/keys. Purple denotes within-block bidirectionality, light green the permitted clean prefix, and green clean-stream causality. Other entries are blocked. The symbols $v$ and $p$ condense visual and prompt tokens. One noisy view with two-token blocks is shown; noisy prompt copies are omitted. Training uses $B=32$.}
  \label{fig:groundanything-training-attention}

\end{wrapfigure}

Following multimodal AR-to-diffusion conversion~\citep{FastdVLM,dong2026mineru}, we retain the pretrained backbone and grounding vocabulary and add a mask token. For a clean sequence $x=[V,P,Y]$ of visual embeddings, text prompt, and response, we partition $Y$ into $B=32$ token blocks, independently of geometric-tuple boundaries. Sampling $t_b\sim\mathcal{U}(0,1)$ independently per block, we corrupt only response positions to form complementary noisy views $w^{(a)}=[P,\widetilde{Y}^{(a)}]$, $a\in\{1,2\}$; EOS is masked in both. The joint input $[w^{(1)};w^{(2)};x]$ contains visual embeddings only in $x$.

Under the mask in \Cref{fig:groundanything-training-attention}, each noisy response block attends bidirectionally within itself and reads the clean image/prompt context and strictly preceding clean response blocks. Clean tokens use causal attention and cannot read noisy views. The shared decoder minimizes
\begin{equation}
\mathcal{L}=0.5\mathcal{L}_{\mathrm{MDM}}+0.5\mathcal{L}_{\mathrm{AR}},
\label{eq:groundanything-conversion-loss}
\end{equation}
where $\mathcal{L}_{\mathrm{MDM}}$ supervises masked response targets and $\mathcal{L}_{\mathrm{AR}}$ supervises clean next-token prediction. Both retain the pretrained token shift.
Full loss definitions and training settings appear in \Cref{app:groundanything_conversion,app:groundanything_training}.

\subsubsection{Reinforcement Post-Training}
\label{sec:reinforcement_posttraining}

\begin{figure}[htbp]
  \centering
  \includegraphics[width=1.0\linewidth]{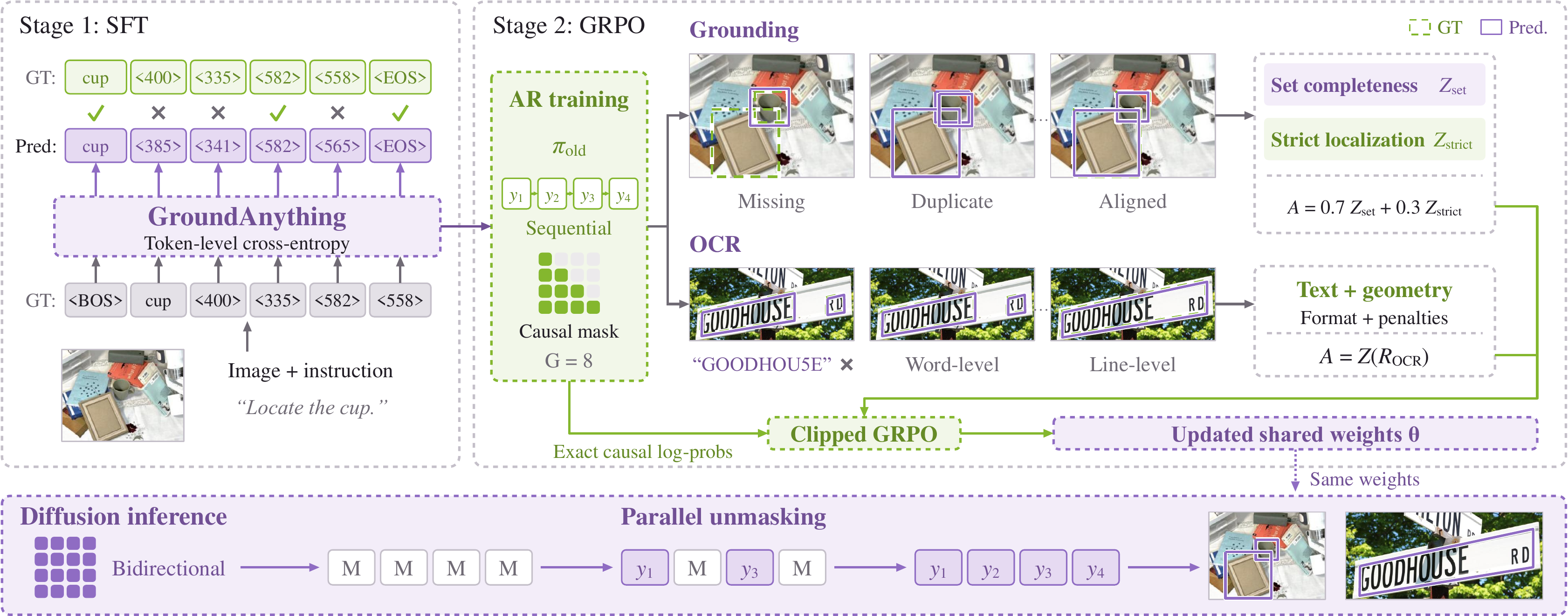}
  \caption{\textbf{Training and inference pipeline.} SFT and causal GRPO update shared parameters reused for blockwise diffusion inference. The schematic illustrates token-level supervision and grounding/OCR reward aggregation; the final advantage includes the additional group normalization in \Cref{eq:groundanything-grounding-advantage}. Rollouts and unmasking states are illustrative.}
  \label{fig:groundanything-sft-rl}
\end{figure}

We apply GRPO~\citep{GRPO,ping2026longactharnessingintrinsicactivation} using causal rollouts and exact causal token likelihoods (\Cref{fig:groundanything-sft-rl}). Box grounding combines set-completeness $R_{\mathrm{set}}$ and strict localization $R_{\mathrm{strict}}$; OCR scores annotation-aware text--geometry agreement; pointing scores location, count, label, and format. For each prompt group, advantages are $A_i=Z_G(\sum_c w_c Z_G(R_{c,i}))$, where $Z_G$ standardizes within the group. Box-reward weights are $(0.7,0.3)$; OCR and pointing each provide one scalar reward. We optimize the clipped objective
\FinishArxivWrap
\begin{equation}
  \mathcal{L}_{\mathrm{GRPO}}=-\mathbb{E}_i\!\left[|\mathcal{T}_i|^{-1}\!\sum\nolimits_{t\in\mathcal{T}_i}\min\!\bigl(r_{i,t}A_i,\operatorname{clip}(r_{i,t},1-\epsilon,1+\epsilon)A_i\bigr)\right],
  \label{eq:groundanything-grpo-main}
\end{equation}
where $r_{i,t}$ is the current-to-old causal token-probability ratio at the rollout temperature, $\mathcal{T}_i$ contains valid response tokens including EOS, and $\epsilon=0.2$. One update per rollout trains the projector and language parameters with the vision encoder frozen and no reference-policy KL term. The updated weights are reused for diffusion inference. Reward definitions and variant-specific reinforcement settings appear in \Cref{app:groundanything_rewards,app:groundanything_rl,app:groundanything_vlm_rl}.

\subsection{Efficient Parallel Inference}
\label{sec:parallel_inference}

\paragraph{Parallel diffusion decoding.}
After image/query prefill, GroundAnything denoises response blocks with bidirectional attention and caches the completed prefix. Sub-blocks are processed from left to right. Our default \emph{Entropy-Guided Decoding}~\citep{WeDLM} commits masked positions with $H_j\leq\tau$, where $H_j=-\sum_v p_j(v)\log p_j(v)$ is the entropy of the unmodified token distribution. If none qualifies, the lowest-entropy position is committed to ensure progress. Committed tokens remain fixed, and a causal pass constructs the completed block's cache without AR verification. Dynamic and Static Decoding instead use confidence thresholds and fixed quotas; commitment and cache construction are detailed in \Cref{app:groundanything_decoding}.

\begin{figure}[htbp]
  \centering
  \includegraphics[width=1.0\linewidth]{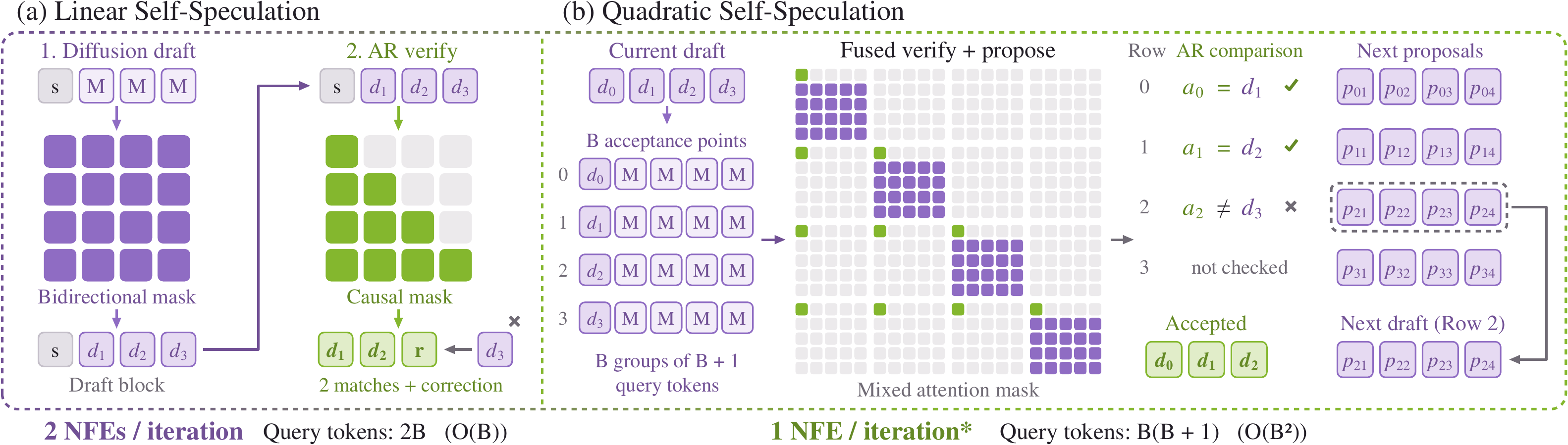}
  \caption{\textbf{Self-speculative decoding with shared model weights ($B=4$).} Linear decoding separates drafting from causal verification: $d_1$ and $d_2$ are accepted, while the mismatched $d_3$ is replaced by the AR correction $r$. Quadratic decoding compares $a_i$ with $d_{i+1}$: $a_0=d_1$, $a_1=d_2$, and $a_2\ne d_3$, thereby committing $[d_0,d_1,d_2]$ and reusing Row~2 as the next still-unverified proposal. Purple denotes bidirectional proposals and green causal verification. Matrix rows and columns denote queries and keys, respectively; the cached prefix is omitted. $M$ denotes \texttt{[MASK]}, $s$ the last accepted seed, and $p_{ij}$ proposal token $j$ from row $i$. One NFE per quadratic round applies after initialization.}
  \label{fig:self-speculative-decoding}
\end{figure}

\paragraph{Optional self-speculative decoding.}
The shared weights also support diffusion drafting with causal verification~\citep{FastdVLM}, accepting the longest matching prefix (\Cref{fig:self-speculative-decoding}). Linear decoding uses two network function evaluations (NFEs) and $2B$ query tokens per round; quadratic decoding fuses verification and proposal generation in one NFE after initialization, using $B(B+1)$ query tokens. These counts describe queries, not Transformer FLOPs. Acceptance and proposal reuse are detailed in \Cref{app:groundanything_speculative}; CUDA Graph and selective FP8 optimizations appear in \Cref{app:groundanything_execution}.

\FinishArxivWrap

\Needspace{14\baselineskip}
\section{Experiments}
\label{sec:experiments}

\ifdefined\BenchmarkTableFont\else
\newcommand{\BenchmarkTableFont}{\normalfont}
\newcommand{\BenchHead}[1]{\begin{tabular}{@{}c@{}}#1\end{tabular}}
\colorlet{benchmarktype}{black!8}
\definecolor{benchmarkpurple}{HTML}{F5F1FA}
\definecolor{benchmarkgreen}{HTML}{F0F3E8}
\fi

\ifdefined\GPIBoxHelpersLoaded\else
\def\GPIBoxHelpersLoaded{1}
\ifdefined\tcolorbox
  \providecommand{\finding}[2]{%
    \begin{tcolorbox}[colback=blue!4,colframe=blue!55!black,
      boxrule=0.6pt,arc=2pt,left=5pt,right=5pt,top=3pt,bottom=3pt,
      before skip=5pt,after skip=5pt]\small
      \textbf{\textit{Finding #1:}} #2
    \end{tcolorbox}}
  \providecommand{\takeaway}[2]{%
    \begin{tcolorbox}[colback=blue!4,colframe=blue!55!black,
      boxrule=0.6pt,arc=2pt,left=5pt,right=5pt,top=3pt,bottom=3pt,
      before skip=5pt,after skip=5pt]\small
      \textbf{\textit{Takeaway #1:}} #2
    \end{tcolorbox}}
\else
  \providecommand{\finding}[2]{\begin{gpiquestions}\textbf{\textit{Finding #1:}} #2\end{gpiquestions}}
  \providecommand{\takeaway}[2]{\begin{gpiquestions}\textbf{\textit{Takeaway #1:}} #2\end{gpiquestions}}
\fi
\fi

In this section, we ask:
\begin{introquestions}
  \introquestion{1}{\textbf{Grounding capability.} How far can GroundAnything push the breadth and precision of visual grounding?}
  \introquestion{2}{\textbf{Speed.} How far can parallel grounding accelerate inference with minimal loss of precision, and what makes its speedups practical?}
  \introquestion{3}{\textbf{Design.} Which model, training, and output choices support precise and efficient parallel grounding?}
\end{introquestions}

\subsection{Implementation Details and Evaluation Setup.}
\label{sec:implementation_evaluation}

We evaluate GroundAnything-VLM and GroundAnything against 44 baselines on 30 benchmarks spanning 11 perceptual capabilities, covering specialized detectors, general-purpose VLMs, grounding specialists, and embodied foundations. All numerical results and conclusions involving our models are based on the mean of ten runs, using five random seeds with two runs per seed. Owing to the high computational cost, other baselines that we evaluate locally for the main leaderboard are averaged over three runs, using three random seeds with one run per seed. GPT-6 Astra uses thinking effort \texttt{High}.
We report F1mIoU for box grounding and OCR, F1@Point for object pointing, and task-specific accuracy for spatial and GUI grounding.

For a fair comparison of diffusion language model (DLM) decoding, all GroundAnything results in the main benchmark tables (Tables~\ref{tab:main-detection}--\ref{tab:main-object-pointing}) use entropy-guided diffusion decoding, without self-speculative decoding.
The self-speculative setting provides higher throughput and quality at the reported operating point, but combines diffusion drafting with causal autoregressive (AR) verification and correction; it is therefore not pure DLM decoding.
We analyze this setting separately in Section~\ref{sec:speed_evaluation} and highlight it in the teaser (Figure~\ref{fig:teaser}).

\subsection{Main Results}
\label{sec:main_results}

\paragraph{Benchmark suite.}
Detection covers common and long-tailed objects on COCO~\citep{COCO} and LVIS~\citep{LVIS}, and dense and tiny objects on Dense200~\citep{rexomni} and VisDrone~\citep{VisDrone}.
Referring grounding uses RefCOCO, RefCOCO+, and RefCOCOg~\citep{RefCOCO,RefCOCOg,RefCOCOgUMD}; object pointing covers the four detection datasets and RefCOCOg.
Spatial and GUI grounding use RoboSpatial~\citep{RoboSpatial}, RefSpatial~\citep{RoboRefer2B}, ScreenSpot-V2~\citep{ScreenSpotV2}, ScreenSpot-Pro~\citep{ScreenSpotPro}, and OSWorld-G~\citep{JEDI3B}.
OCR uses HierText~\citep{HierText}, ICDAR2015~\citep{ICDAR2015}, TotalText~\citep{TotalText}, and SROIE~\citep{SROIE}; layout grounding uses DocLayNet~\citep{DocLayNet} and M6Doc~\citep{M6Doc}.
Exemplar-based visual prompting on FSC147~\citep{FSC147} and Dense200 is also included in the main results.

\begin{wrapfigure}{r}{0.65\textwidth}
  \raggedleft
  \includegraphics[width=\linewidth]{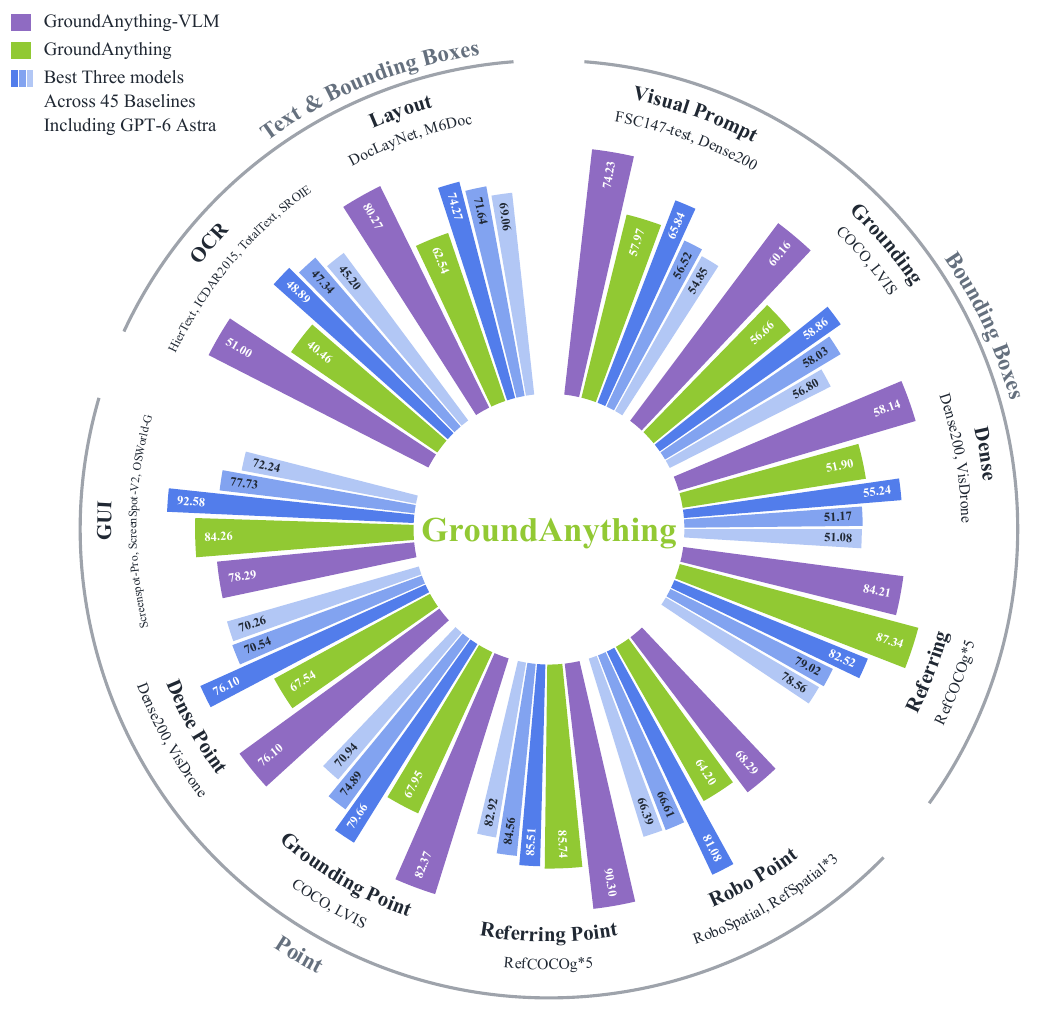}
  \caption{\textbf{Grounding across task groups.} GroundAnything-VLM, GroundAnything, and leading baselines across box, point, text, and exemplar-conditioned tasks. Complete results: \Cref{app:detailed_results}.}
  \label{fig:groundanything-benchmark-results}
\end{wrapfigure}

GroundAnything-VLM establishes a new overall best among similarly sized models at 72.42\%, remaining competitive with GPT-6 Astra (71.35\%).
GroundAnything reaches 61.75\% with pure diffusion decoding, surpassing the prior overall AR state of the art at comparable scale.
\Cref{fig:groundanything-benchmark-results} summarizes this breadth; the comparisons below examine localization precision and task-dependent diffusion--AR gaps.

\FinishArxivWrap

\paragraph{Reporting conventions.}
\label{sec:benchmark_reporting}
Scores are percentages; bold marks column bests, including ties (lower is better only for parse error).
Model-name stars denote externally reported rows; entry-level stars denote source or task-interface exceptions.
\texttt{--}, \texttt{N/A}, and \texttt{UNK} indicate unreported values, unsupported or unresolved evaluations, and unspecified zero-shot status, respectively.
Daggers flag uncertain prompt/protocol alignment, so affected scores are descriptive.
Under our protocols, GroundingDINO lacks GUI/OCR/layout interfaces, and Kimi-K3 lacks compatible pointing/OCR/layout outputs; SenseNova-Vision's GUI evaluation remains unresolved.
LocateAnything lacks a supported visual-prompt interface.
Starred SenseNova-Vision HierText/ICDAR2015 scores follow \citet{SenseNovaVision7BMoT}; its other OCR scores are local.
Complete sources and exceptions appear in \Cref{app:detailed_results}.

\paragraph{Detection and referring grounding.}
GroundAnything reaches 70.04 F1mIoU on Dense200 and 91.61/91.10 on RefCOCOg val/test (\Cref{tab:main-detection}), extending precise diffusion grounding to crowded scenes and language-conditioned targets, although tiny-object localization remains challenging.
RefCOCO avg is the unweighted mean of RefCOCO, RefCOCOg, and RefCOCO+.
\FinishArxivWrap
\begin{table}[htbp]
  \centering
  \BenchmarkTableFont
  \caption{\textbf{Detection and referring grounding (F1mIoU).} External rows follow \citet{rexomni}. Full results appear in \Cref{app:common_longtailed_detection,app:dense_tiny_detection,app:referring_detection}.}
  \label{tab:main-detection}
  \begingroup
  \fontsize{8}{9.7}\selectfont
  \setlength{\tabcolsep}{3pt}
  \renewcommand{\arraystretch}{1.15}
  \resizebox{\linewidth}{!}{%
  \begin{NiceTabular}{@{}>{\raggedright\arraybackslash}p{177pt}ccccccc@{}}
  \CodeBefore
    \rowcolor{benchmarktype}{3}
    \rowcolor{benchmarktype}{6}
    \rowcolor{benchmarktype}{8}
    \rowcolor{benchmarkpurple}{16}
    \rowcolor{benchmarkgreen}{17}
    \rowcolor{benchmarktype}{18}
    \rowcolor{benchmarktype}{21}
  \Body
  \toprule
  \textbf{Model} & \multicolumn{1}{c}{\textbf{Common}} & \multicolumn{1}{c}{\textbf{Long-tailed}} & \multicolumn{2}{c}{\textbf{Dense \& Tiny}} & \multicolumn{3}{c}{\textbf{Referring grounding}} \\
  \cmidrule(lr){2-2}\cmidrule(lr){3-3}\cmidrule(lr){4-5}\cmidrule(lr){6-8}
   & \BenchHead{COCO} & \BenchHead{LVIS} & \BenchHead{Dense200} & \BenchHead{VisDrone} & \BenchHead{RefCOCOg\\val} & \BenchHead{RefCOCOg\\test} & \BenchHead{RefCOCO\\avg} \\
  \midrule
  \multicolumn{8}{@{}l}{\hspace{0.4em}\strut\textbf{Closed-set Specialized Detectors}} \\
  DINO-R50\textsuperscript{*}\hspace{0.35em}\citep{DINOR50} & 55.60 & -- & -- & -- & -- & -- & -- \\
  DETR-R50\textsuperscript{*}\hspace{0.35em}\citep{DETRR50} & 48.30 & -- & -- & -- & -- & -- & -- \\
  \addlinespace[2pt]
  \multicolumn{8}{@{}l}{\hspace{0.4em}\strut\textbf{Open-set Specialized Detectors}} \\
  GroundingDINO\hspace{0.35em}\citep{groundingdino} & 60.56 & 52.61 & 24.92 & 34.47 & 49.77 & 50.43 & 45.15 \\
  \addlinespace[2pt]
  \multicolumn{8}{@{}l}{\hspace{0.4em}\strut\textbf{Vision-Language Models (<10B)}} \\
  Qwen3-VL-4B\hspace{0.35em}\citep{qwen3vl} & 46.53 & 49.86 & 14.02 & 31.14 & 75.27 & 75.88 & 75.24 \\
  Qwen3.5-9B\hspace{0.35em}\citep{qwen35} & 51.99 & 48.87 & 30.02 & 32.84 & 76.20 & 76.28 & 76.08 \\
  Rex-Omni\hspace{0.35em}\citep{rexomni} & 56.28 & 46.74 & 53.29 & 27.19 & 73.90 & 74.76 & 69.29 \\
  LocateAnything Fast\hspace{0.35em}\citep{locateanything} & 53.06 & 42.90 & 20.70 & 9.82 & 75.30 & 76.50 & 76.62 \\
  LocateAnything Hybrid\hspace{0.35em}\citep{locateanything} & 59.12 & 49.56 & 50.07 & 28.57 & 76.43 & 77.67 & 78.06 \\
  SenseNova-Vision\hspace{0.35em}\citep{SenseNovaVision7BMoT} & 57.49 & 56.12 & 68.13 & \textbf{42.35} & 78.69 & 79.48 & 77.55 \\
  RynnBrain1.1\hspace{0.35em}\citep{RynnBrain112B} & 35.15 & 26.01 & 0.12 & 8.29 & 67.66 & 68.24 & 63.63 \\
  GroundAnything-VLM & \textbf{63.70} & \textbf{56.63} & \textbf{75.55} & 40.74 & 84.37 & 83.56 & 83.43 \\
  GroundAnything & 60.72 & 52.59 & 70.04 & 33.76 & \textbf{91.61} & \textbf{91.10} & \textbf{85.33} \\
  \addlinespace[2pt]
  \multicolumn{8}{@{}l}{\hspace{0.4em}\strut\textbf{Vision-Language Models (10B--1T)}} \\
  SEED1.5-VL\textsuperscript{*}\hspace{0.35em}\citep{SEED15VL} & 51.40 & 46.70 & 53.20 & 27.40 & 71.90 & 73.20 & -- \\
  Qwen3.8-27B\hspace{0.35em}\citep{qwen38} & 60.84 & 49.99 & 34.30 & 33.33 & 76.03 & 77.37 & 77.89 \\
  \addlinespace[2pt]
  \multicolumn{8}{@{}l}{\hspace{0.4em}\strut\textbf{Vision-Language Models (>1T)}} \\
  Qwen3.7-Max\hspace{0.35em}\citep{qwen37} & 62.79 & 53.28 & 31.20 & 41.59 & 80.54 & 81.71 & 80.20 \\
  Kimi-K2.6 & 61.16 & 51.19 & 42.67 & 29.95 & 70.33 & 71.87 & 69.66 \\
  Kimi-K3\hspace{0.35em}\citep{KimiK3} & 60.89 & 47.73 & 51.64 & 30.31 & 73.39 & 73.99 & 72.86 \\
  GPT-6 Astra & 62.75 & 54.97 & 65.04 & 37.12 & 74.98 & 78.91 & 77.81 \\
  \addlinespace[2pt]
  \bottomrule
  
  \end{NiceTabular}%
  }
  \endgroup
\end{table}

\paragraph{Robot, spatial, and GUI grounding.}
GroundAnything matches its AR counterpart on RefSpatial Unseen (62.34\%) and improves ScreenSpot-Pro from 65.34\% to 75.96\% (\Cref{tab:main-spatial-gui}), while Astra remains stronger on RefSpatial and GUI tasks.
RefSpatial avg averages the Location and Placement splits.
\begin{table}[htbp]
  \centering
  \BenchmarkTableFont
  \caption{\textbf{Robot, spatial, and GUI grounding (accuracy).} External RefSpatial and JEDI/UI-R1 scores follow \citet{rexomni}; GUI-Owl scores follow \citet{locateanything}. Full results appear in \Cref{app:robot_spatial_pointing,app:gui_grounding}.}
  \label{tab:main-spatial-gui}
  \begingroup
  \fontsize{8}{9.7}\selectfont
  \setlength{\tabcolsep}{3pt}
  \renewcommand{\arraystretch}{1.15}
  \resizebox{\linewidth}{!}{%
  \begin{NiceTabular}{@{}>{\raggedright\arraybackslash}p{177pt}cccccc@{}}
  \CodeBefore
    \rowcolor{benchmarktype}{3}
    \rowcolor{benchmarktype}{5}
    \rowcolor{benchmarkpurple}{16}
    \rowcolor{benchmarkgreen}{17}
    \rowcolor{benchmarktype}{18}
    \rowcolor{benchmarktype}{24}
  \Body
  \toprule
  \textbf{Model} & \multicolumn{3}{c}{\textbf{Robot and spatial pointing}} & \multicolumn{3}{c}{\textbf{GUI grounding}} \\
  \cmidrule(lr){2-4}\cmidrule(lr){5-7}
   & \BenchHead{RefSpatial\\(avg)} & \BenchHead{RefSpatial\\Unseen} & \BenchHead{RoboSpatial\\Context} & \BenchHead{ScreenSpot-Pro} & \BenchHead{ScreenSpot-V2} & \BenchHead{OSWorld-G} \\
  \midrule
  \multicolumn{7}{@{}l}{\hspace{0.4em}\strut\textbf{Open-set Specialized Detectors}} \\
  GroundingDINO\hspace{0.35em}\citep{groundingdino} & 14.25 & 4.33 & 4.92 & N/A\textsuperscript{*} & N/A\textsuperscript{*} & N/A\textsuperscript{*} \\
  \addlinespace[2pt]
  \multicolumn{7}{@{}l}{\hspace{0.4em}\strut\textbf{Vision-Language Models (<10B)}} \\
  JEDI\textsuperscript{*}\hspace{0.35em}\citep{JEDI3B} & -- & -- & -- & 36.10 & 88.60 & -- \\
  UI-R1\textsuperscript{*}\hspace{0.35em}\citep{UIR13B} & -- & -- & -- & 17.80 & 85.40 & -- \\
  Qwen3-VL-4B\hspace{0.35em}\citep{qwen3vl} & 49.00 & 27.27 & 64.75 & 56.74 & 92.30 & 56.91 \\
  Qwen3.5-9B\hspace{0.35em}\citep{qwen35} & 55.92 & 37.01 & 60.66 & 53.13 & 90.57 & 60.99 \\
  Rex-Omni\hspace{0.35em}\citep{rexomni} & 51.75 & 37.01 & 59.02 & 36.75 & 88.29 & 46.10 \\
  LocateAnything Fast\hspace{0.35em}\citep{locateanything} & 36.00 & 16.88 & 15.57 & 56.04 & 88.44 & 59.93 \\
  LocateAnything Hybrid\hspace{0.35em}\citep{locateanything} & 36.17 & 20.78 & 14.75 & 57.05 & 89.94 & 60.46 \\
  SenseNova-Vision\hspace{0.35em}\citep{SenseNovaVision7BMoT} & 20.38 & 8.54 & 0.82 & N/A\textsuperscript{\textdagger} & N/A\textsuperscript{\textdagger} & N/A\textsuperscript{\textdagger} \\
  RynnBrain1.1\hspace{0.35em}\citep{RynnBrain112B} & 50.60 & 36.90 & 54.10 & 34.66 & 70.44 & 33.33 \\
  RoboRefer\textsuperscript{*}\hspace{0.35em}\citep{RoboRefer2B} & 50.00 & 39.00 & -- & -- & -- & -- \\
  GroundAnything-VLM & 69.00 & 62.34 & \textbf{72.13} & 65.34 & 94.89 & 74.65 \\
  GroundAnything & 63.00 & 62.34 & 69.67 & 75.96 & 95.60 & 81.21 \\
  \addlinespace[2pt]
  \multicolumn{7}{@{}l}{\hspace{0.4em}\strut\textbf{Vision-Language Models (10B--1T)}} \\
  RoboPoint\textsuperscript{*}\hspace{0.35em}\citep{RoboPoint13B} & 16.10 & 8.40 & -- & -- & -- & -- \\
  GUI-Owl\textsuperscript{*}\hspace{0.35em}\citep{GUIOwl32B} & -- & -- & -- & 58.00 & -- & -- \\
  Gemini-2.5-Pro\textsuperscript{*}\hspace{0.35em}\citep{Gemini2.5Pro} & 35.60 & 27.10 & -- & -- & -- & -- \\
  Molmo-72B\textsuperscript{*}\hspace{0.35em}\citep{Molmo} & 30.25 & 21.20 & -- & -- & -- & -- \\
  Qwen3.8-27B\hspace{0.35em}\citep{qwen38} & 60.00 & 46.75 & 63.93 & 58.76 & 94.50 & 63.48 \\
  \addlinespace[2pt]
  \multicolumn{7}{@{}l}{\hspace{0.4em}\strut\textbf{Vision-Language Models (>1T)}} \\
  Qwen3.7-Max\hspace{0.35em}\citep{qwen37} & 68.75 & 57.14 & 69.67 & 55.06 & 81.13 & 49.29 \\
  Kimi-K2.6 & 41.33 & 41.56 & 30.33 & 6.07\textsuperscript{\textdagger} & 52.36\textsuperscript{\textdagger} & 10.11\textsuperscript{\textdagger} \\
  Kimi-K3\hspace{0.35em}\citep{KimiK3} & 58.92 & 54.98 & 54.92 & 25.36\textsuperscript{\textdagger} & 82.70\textsuperscript{\textdagger} & 68.26\textsuperscript{\textdagger} \\
  GPT-6 Astra & \textbf{85.93} & \textbf{81.93} & 65.69 & \textbf{93.17} & \textbf{97.88} & \textbf{86.70} \\
  \addlinespace[2pt]
  \bottomrule
  
  \end{NiceTabular}%
  }
  \endgroup
\end{table}

\paragraph{OCR, layout, and visual prompting.}
The shared interface extends to text, document regions, and exemplar-matched instances: GroundAnything-VLM reaches 85.78 on DocLayNet and 74.76 on M6Doc, while GroundAnything scores 63.52 on Dense200 visual prompting (\Cref{tab:main-ocr-layout}).
OCR and layout exhibit larger diffusion--AR gaps than referring and GUI grounding, identifying where parallel generation still sacrifices precision.
\begin{table}[htbp]
  \centering
  \BenchmarkTableFont
  \caption{\textbf{OCR, layout, and visual prompting (F1mIoU).} External rows follow \citet{rexomni}. SenseNova-Vision uses published HierText/ICDAR2015 scores~\citep{SenseNovaVision7BMoT} and locally evaluated TotalText/SROIE scores. Full results appear in \Cref{app:ocr,app:layout_grounding,app:visual_prompting}.}
  \label{tab:main-ocr-layout}
  \begingroup
  \fontsize{8}{9.7}\selectfont
  \setlength{\tabcolsep}{3pt}
  \renewcommand{\arraystretch}{1.15}
  \resizebox{\linewidth}{!}{%
  \begin{NiceTabular}{@{}>{\raggedright\arraybackslash}p{177pt}cccccccc@{}}
  \CodeBefore
    \rowcolor{benchmarktype}{3}
    \rowcolor{benchmarktype}{6}
    \rowcolor{benchmarkpurple}{14}
    \rowcolor{benchmarkgreen}{15}
    \rowcolor{benchmarktype}{16}
    \rowcolor{benchmarktype}{19}
  \Body
  \toprule
  \textbf{Model} & \multicolumn{4}{c}{\textbf{OCR}} & \multicolumn{2}{c}{\textbf{Layout grounding}} & \multicolumn{2}{c}{\textbf{Visual prompting}} \\
  \cmidrule(lr){2-5}\cmidrule(lr){6-7}\cmidrule(lr){8-9}
   & \BenchHead{HierText} & \BenchHead{ICDAR2015} & \BenchHead{TotalText} & \BenchHead{SROIE} & \BenchHead{DocLayNet} & \BenchHead{M6Doc} & \BenchHead{FSC147} & \BenchHead{Dense200} \\
  \midrule
  \multicolumn{9}{@{}l}{\hspace{0.4em}\strut\textbf{Closed-set Specialized Detectors}} \\
  DocLayout-YOLO\textsuperscript{*}\hspace{0.35em}\citep{DocLayoutYOLO} & -- & -- & -- & -- & 81.10 & -- & -- & -- \\
  PaddleOCRv5\textsuperscript{*}\hspace{0.35em}\citep{PaddleOCRv5} & 30.50 & 25.60 & 25.70 & 58.60 & -- & -- & -- & -- \\
  \addlinespace[2pt]
  \multicolumn{9}{@{}l}{\hspace{0.4em}\strut\textbf{Vision-Language Models (<10B)}} \\
  Qwen3-VL-4B\hspace{0.35em}\citep{qwen3vl} & 23.48 & 28.41 & 38.35 & 40.41 & 40.81 & 24.73 & 18.90 & 2.49 \\
  Qwen3.5-9B\hspace{0.35em}\citep{qwen35} & 29.63 & 29.90 & 37.26 & 26.74 & 34.65 & 17.35 & 20.56 & 26.93 \\
  Rex-Omni\hspace{0.35em}\citep{rexomni} & 34.46 & 45.65 & 52.35 & 48.35 & 68.06 & 54.95 & 57.15 & 55.50 \\
  LocateAnything Fast\hspace{0.35em}\citep{locateanything} & 22.59 & 26.73 & 44.40 & 24.89 & 49.37 & 47.03 & N/A\textsuperscript{*} & N/A\textsuperscript{*} \\
  LocateAnything Hybrid\hspace{0.35em}\citep{locateanything} & 26.65 & 27.48 & 45.49 & 30.05 & 77.34 & 65.94 & N/A\textsuperscript{*} & N/A\textsuperscript{*} \\
  SenseNova-Vision\hspace{0.35em}\citep{SenseNovaVision7BMoT} & 31.20\textsuperscript{*} & \textbf{49.50}\textsuperscript{*} & 11.40 & 36.26 & 85.53 & 35.62 & \textbf{62.51} & 62.84 \\
  RynnBrain1.1\hspace{0.35em}\citep{RynnBrain112B} & 2.07 & 17.81 & 16.90 & 3.59 & 6.02 & 4.53 & 4.78 & 1.45 \\
  GroundAnything-VLM & 41.33 & 42.50 & 49.03 & \textbf{71.15} & \textbf{85.78} & \textbf{74.76} & 60.29 & \textbf{75.26} \\
  GroundAnything & 33.31 & 41.81 & 43.08 & 43.64 & 68.38 & 56.69 & 52.43 & 63.52 \\
  \addlinespace[2pt]
  \multicolumn{9}{@{}l}{\hspace{0.4em}\strut\textbf{Vision-Language Models (10B--1T)}} \\
  SEED1.5-VL\textsuperscript{*}\hspace{0.35em}\citep{SEED15VL} & 12.00 & 18.70 & 19.50 & 28.10 & 28.70 & 28.00 & -- & -- \\
  Qwen3.8-27B\hspace{0.35em}\citep{qwen38} & 32.95 & 39.72 & 41.57 & 37.24 & 42.21 & 19.92 & 16.80 & 28.19 \\
  \addlinespace[2pt]
  \multicolumn{9}{@{}l}{\hspace{0.4em}\strut\textbf{Vision-Language Models (>1T)}} \\
  Qwen3.7-Max\hspace{0.35em}\citep{qwen37} & \textbf{42.95} & 43.07 & 48.64 & 38.54 & 46.93 & 32.31 & 21.62 & 31.81 \\
  Kimi-K2.6 & 25.26\textsuperscript{\textdagger} & 32.25\textsuperscript{\textdagger} & 40.99\textsuperscript{\textdagger} & 46.83\textsuperscript{\textdagger} & 15.59\textsuperscript{\textdagger} & 13.50\textsuperscript{\textdagger} & 32.47 & 32.41 \\
  GPT-6 Astra & 39.58 & 48.87 & \textbf{53.55} & 53.57 & 77.54 & 60.59 & 61.04 & 68.94 \\
  \addlinespace[2pt]
  \bottomrule
  
  \end{NiceTabular}%
  }
  \endgroup
\end{table}

\paragraph{Object pointing.}
Following \citet{rexomni}, SAM-derived masks~\citep{SAM} determine point correctness, with F1@Point balancing misses and false positives. GroundAnything-VLM leads five of the six selected comparisons.
\begin{table}[htbp]
  \centering
  \BenchmarkTableFont
  \caption{\textbf{Object pointing (F1@Point).} External rows follow \citet{rexomni}, where Molmo denotes Molmo-7B-D. Full results appear in \Cref{app:object_pointing}.}
  \label{tab:main-object-pointing}
  \begingroup
  \fontsize{8}{9.7}\selectfont
  \setlength{\tabcolsep}{3pt}
  \renewcommand{\arraystretch}{1.15}
  \resizebox{\linewidth}{!}{%
  \begin{NiceTabular}{@{}>{\raggedright\arraybackslash}p{177pt}cccccc@{}}
  \CodeBefore
    \rowcolor{benchmarktype}{3}
    \rowcolor{benchmarktype}{5}
    \rowcolor{benchmarkpurple}{14}
    \rowcolor{benchmarkgreen}{15}
    \rowcolor{benchmarktype}{16}
    \rowcolor{benchmarktype}{19}
  \Body
  \toprule
  \textbf{Model} & \multicolumn{2}{c}{\textbf{Referring object pointing}} & \multicolumn{2}{c}{\textbf{Common / long-tailed}} & \multicolumn{2}{c}{\textbf{Dense / tiny}} \\
  \cmidrule(lr){2-3}\cmidrule(lr){4-5}\cmidrule(lr){6-7}
   & \BenchHead{RefCOCOg val} & \BenchHead{RefCOCOg test} & \BenchHead{COCO} & \BenchHead{LVIS} & \BenchHead{Dense200} & \BenchHead{VisDrone} \\
  \midrule
  \multicolumn{7}{@{}l}{\hspace{0.4em}\strut\textbf{Open-set Specialized Detectors}} \\
  GroundingDINO\hspace{0.35em}\citep{groundingdino} & 49.34 & 49.97 & 70.41 & 55.07 & 32.93 & 39.45 \\
  \addlinespace[2pt]
  \multicolumn{7}{@{}l}{\hspace{0.4em}\strut\textbf{Vision-Language Models (<10B)}} \\
  Qwen3-VL-4B\hspace{0.35em}\citep{qwen3vl} & 76.43 & 77.64 & 65.33 & 55.08 & 21.72 & 23.50 \\
  Qwen3.5-9B\hspace{0.35em}\citep{qwen35} & 77.59 & 77.85 & 72.21 & 64.00 & 65.35 & 44.73 \\
  Rex-Omni\hspace{0.35em}\citep{rexomni} & 84.96 & 85.32 & 79.74 & 70.04 & 76.66 & 51.97 \\
  LocateAnything Fast\hspace{0.35em}\citep{locateanything} & 73.84 & 74.94 & 71.53 & 62.13 & 65.63 & 56.36 \\
  LocateAnything Hybrid\hspace{0.35em}\citep{locateanything} & 75.89 & 76.65 & 73.78 & 64.89 & 78.07 & 57.30 \\
  SenseNova-Vision\hspace{0.35em}\citep{SenseNovaVision7BMoT} & 74.63 & 75.42 & 72.96 & 62.66 & 78.71 & 61.81 \\
  RynnBrain1.1\hspace{0.35em}\citep{RynnBrain112B} & 74.42 & 74.17 & 25.70 & 17.56 & 3.95 & 13.18 \\
  Molmo-7B\textsuperscript{*}\hspace{0.35em}\citep{Molmo} & 83.70 & 83.60 & 77.30 & 40.30 & 33.10 & 29.20 \\
  GroundAnything-VLM & \textbf{90.44} & \textbf{91.03} & \textbf{84.92} & \textbf{79.81} & 84.16 & \textbf{68.03} \\
  GroundAnything & 85.75 & 85.73 & 71.32 & 64.57 & 77.14 & 57.94 \\
  \addlinespace[2pt]
  \multicolumn{7}{@{}l}{\hspace{0.4em}\strut\textbf{Vision-Language Models (10B--1T)}} \\
  SEED1.5-VL\textsuperscript{*}\hspace{0.35em}\citep{SEED15VL} & 83.60 & 84.20 & 78.20 & 70.70 & 72.10 & 56.70 \\
  Qwen3.8-27B\hspace{0.35em}\citep{qwen38} & 75.84 & 75.86 & 74.01 & 67.65 & 74.55 & 51.16 \\
  \addlinespace[2pt]
  \multicolumn{7}{@{}l}{\hspace{0.4em}\strut\textbf{Vision-Language Models (>1T)}} \\
  Qwen3.7-Max\hspace{0.35em}\citep{qwen37} & 71.21 & 72.40 & 72.13 & 66.94 & 66.28 & 59.04 \\
  Kimi-K2.6 & 39.59\textsuperscript{\textdagger} & 39.60\textsuperscript{\textdagger} & 30.65\textsuperscript{\textdagger} & 25.51\textsuperscript{\textdagger} & 35.18\textsuperscript{\textdagger} & 13.63\textsuperscript{\textdagger} \\
  GPT-6 Astra & 87.80 & 84.90 & 82.17 & 77.14 & \textbf{86.57} & 65.62 \\
  \addlinespace[2pt]
  \bottomrule
  
  \end{NiceTabular}%
  }
  \endgroup
\end{table}

\FinishArxivWrap

\finding{1}{Across 30 benchmarks, GroundAnything extends broad and precise grounding to pure diffusion, surpassing the prior overall AR state of the art at comparable scale. GroundAnything-VLM sets a new overall best at this scale, remaining competitive with GPT-6 Astra.}

\subsection{Speed Evaluation}
\label{sec:speed_evaluation}

\subsubsection{Decoding Strategy Analysis}

\question{2.1}{Which decoding choices govern the speed--accuracy trade-off?}

\FinishArxivWrap
\begin{wrapfigure}{r}{0.65\textwidth}

  \centering
  \includegraphics[width=\linewidth]{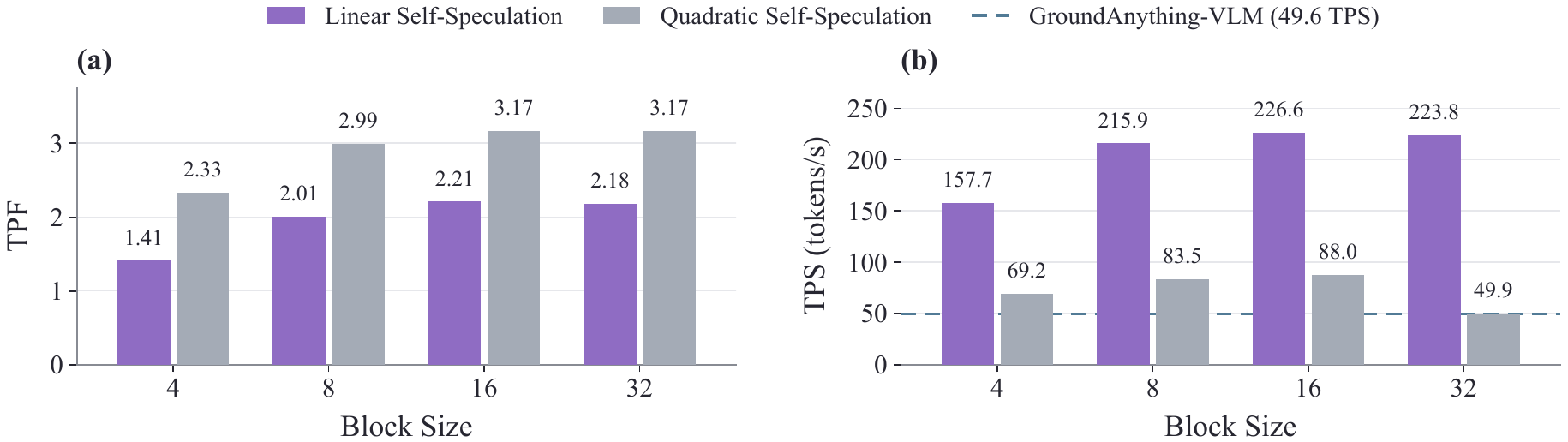}
  \caption{\textbf{Self-speculative schedules.} TPF and TPS across block sizes; dashed lines denote GroundAnything-VLM.}
  \label{fig:self-speculation-block-size}

\end{wrapfigure}

We report tokens per forward (TPF), output tokens per second (TPS), and COCO F1mIoU.
Decoding protocols and additional analysis appear in \Cref{app:speed_reporting}.

\paragraph{Forward efficiency versus latency.}
At $B=32$, quadratic self-speculation reaches higher TPF than the linear schedule (3.17 versus 2.18), yet substantially lower throughput (49.9 versus 223.8 TPS; \Cref{fig:self-speculation-block-size}).
Its $O(B^2)$ query-token cost offsets the reduction in forward calls, favoring the linear schedule at this operating point.

\begin{figure}[htbp]
  \centering
  \includegraphics[width=1.0\linewidth]{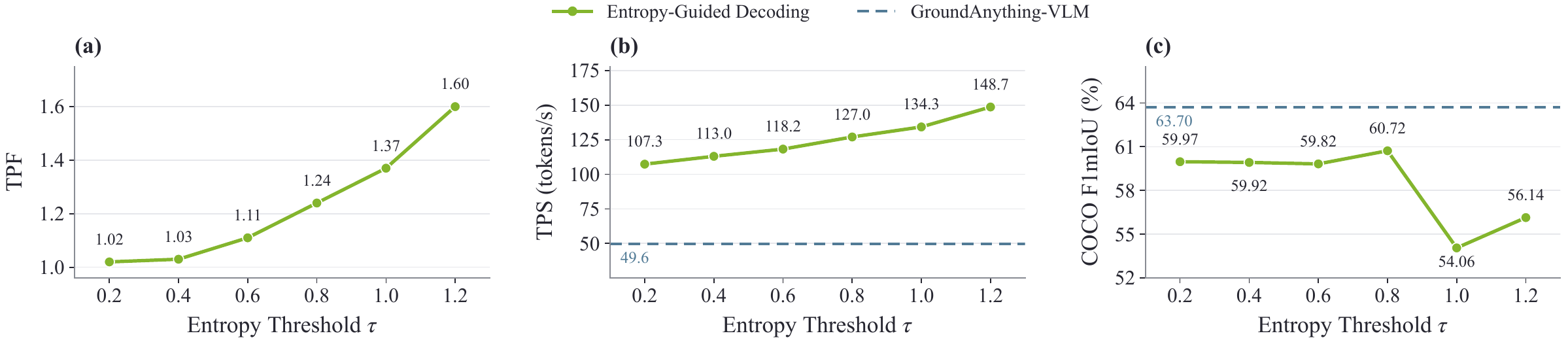}
  \caption{\textbf{Entropy-threshold sensitivity.} TPF, TPS, and COCO F1mIoU at $B=32$ with sub-block size 4. Dashed lines denote GroundAnything-VLM.}
  \label{fig:entropy-threshold}
\end{figure}

\begin{figure}[htbp]
  \centering
  \includegraphics[width=1.0\linewidth]{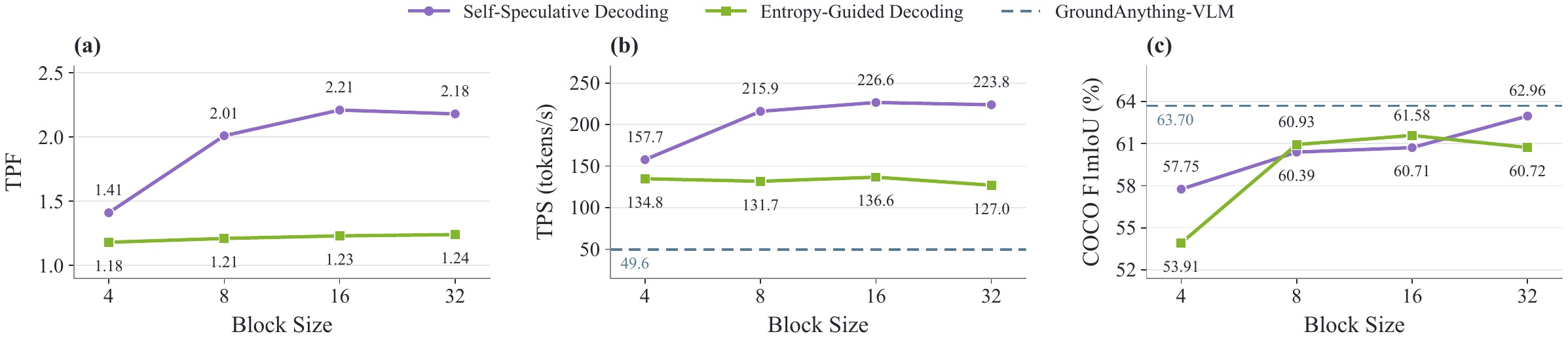}
  \caption{\textbf{Decoding strategy and block size.} Linear self-speculation and entropy guidance ($\tau=0.8$, sub-block size 4) compared on COCO. Dashed lines denote GroundAnything-VLM.}
  \label{fig:decoding-methods-block-size}
\end{figure}

\paragraph{Entropy threshold and block size.}
Relaxing the entropy threshold increases parallel commitment, but COCO quality falls sharply beyond $\tau=0.8$ (\Cref{fig:entropy-threshold}).
At $B=32$, $\tau=0.8$ achieves the highest F1mIoU in this threshold sweep (60.72) at 127.0 TPS, or $2.56\times$ GroundAnything-VLM throughput.
Block size also changes this balance: entropy-guided quality peaks at $B=16$ in this COCO sweep, whereas linear self-speculation peaks at $B=32$ (\Cref{fig:decoding-methods-block-size}); larger blocks are therefore not uniformly better.

\FinishArxivWrap

\finding{2.1}{In direct DLM decoding, the entropy threshold controls the speed--precision trade-off, providing a decoding-effort knob analogous to thinking effort. We favor $\tau=0.8$ in the $B=32$ COCO sweep. Optional self-speculation improves both quality and throughput over this direct-decoding point, reaching $4.51\times$ AR throughput with a 0.74 pp F1mIoU gap.}

\Needspace{6\baselineskip}
\subsubsection{DLM vs.\ MTP}
\label{sec:dlm-vs-mtp}

\question{2.2}{Why use diffusion rather than MTP when both can draft for the same AR verifier?}

We match visual inputs, adaptation data, tokenization, and prefixes, and freeze the same converted model's causal branch as verifier.
The primary MTP baseline is an in-house causal drafter; bidirectional block-MTP and retrained causal-DLM controls test the role of attention (\Cref{app:dlm-mtp-protocol}).

We address four questions:
\begin{introquestions}
    \introquestion{2.2.1}{\textbf{At the same candidate length, which drafter achieves higher acceptance?}}
    \introquestion{2.2.2}{\textbf{How much wall-clock time does each accepted draft token cost?}}
    \introquestion{2.2.3}{\textbf{Can additional DLM refinement repay its extra forward-pass cost?}}
    \introquestion{2.2.4}{\textbf{Which grounding failures benefit from bidirectional attention?}}
\end{introquestions}

\FinishArxivWrap
\begin{wrapfigure}{r}{0.7\textwidth}

\centering
\includegraphics[width=\linewidth]{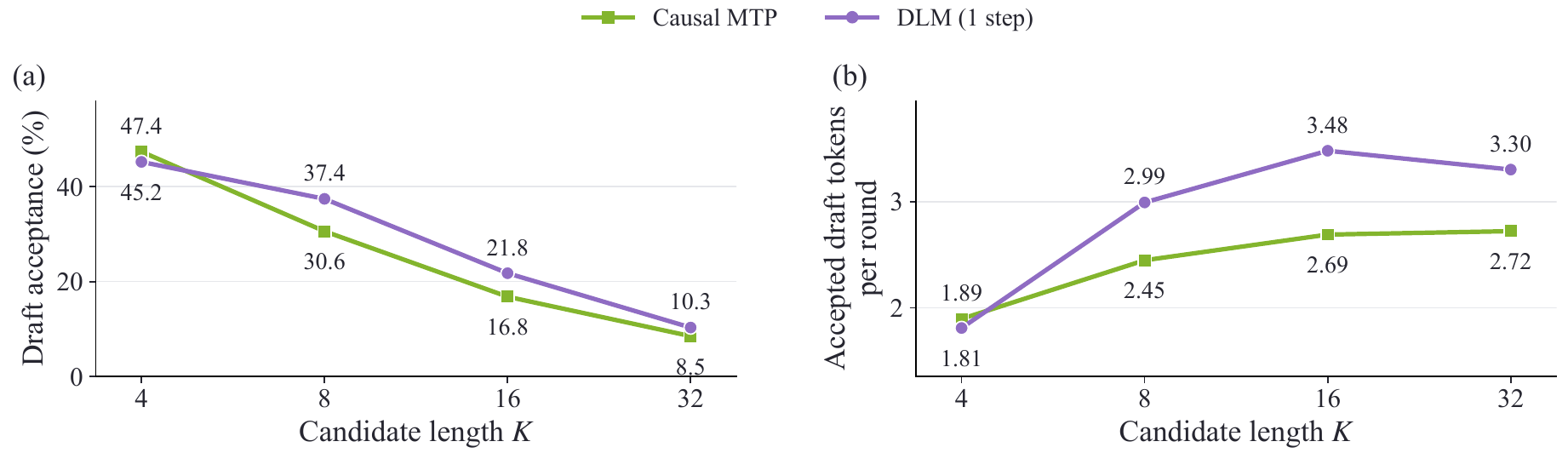}
\caption{\textbf{Acceptance at matched candidate length.} One-step DLM and causal MTP share prefixes and a frozen AR verifier. Accepted tokens exclude verifier corrections and bonuses.}
\label{fig:01_dlm_mtp_acceptance}

\centering
\includegraphics[width=\linewidth]{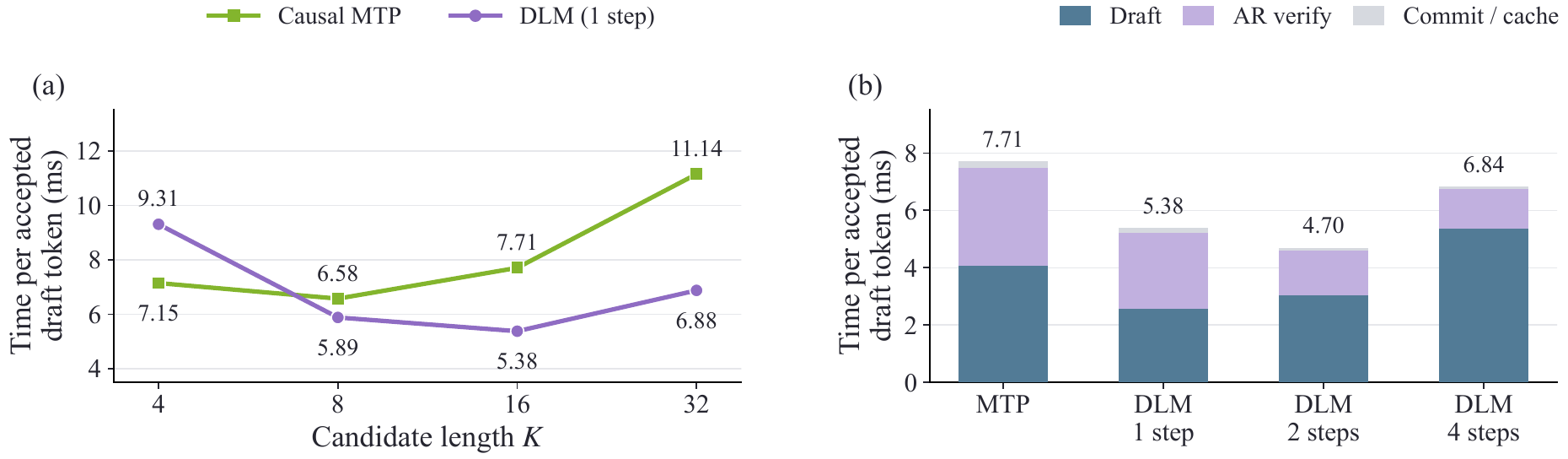}
\caption{\textbf{Cost per accepted draft token.} Full-round latency includes rejected work; the stage breakdown uses $K=16$.}
\label{fig:02_dlm_mtp_time_cost}
\vspace{-10pt}
\end{wrapfigure}

\paragraph{Draft acceptance.}
At $K=16$, one-step DLM accepts 3.48 draft tokens per round versus 2.69 for causal MTP (\Cref{fig:01_dlm_mtp_acceptance}), yielding more useful proposals at the same candidate length.

\paragraph{Cost per accepted draft token.}
Including drafting, verification, and rejected work, one-step DLM reduces full-round cost per accepted draft token from 7.71 to 5.38 ms (\Cref{fig:02_dlm_mtp_time_cost}).
Higher acceptance therefore translates into lower wall-clock cost at this operating point.

\begin{figure}[htbp]
\centering
\includegraphics[width=\linewidth]{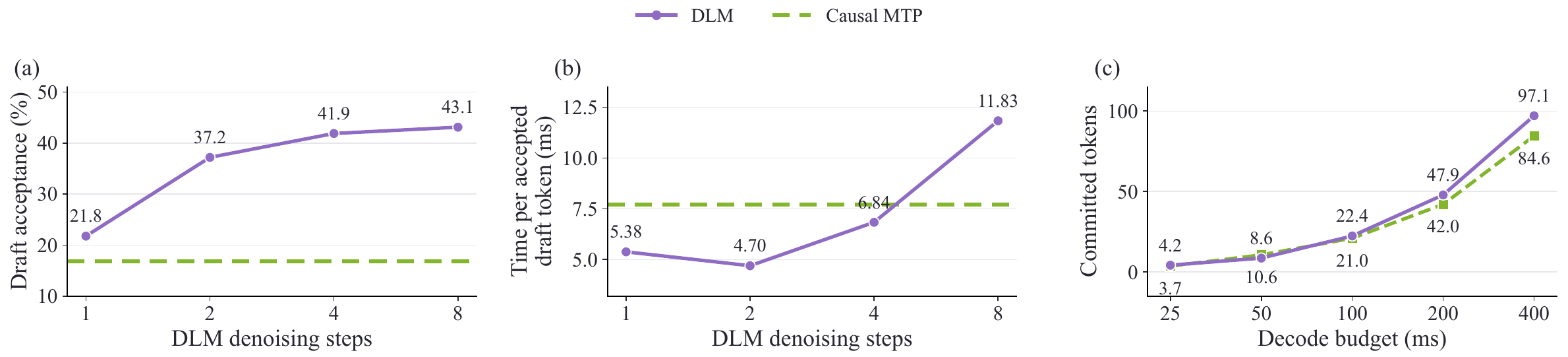}
\caption{\textbf{Refinement and available decode time.} (a,b) Varying refinement depth at $K=16$. (c) Committed tokens under equal deadlines on long-output prefixes, using validation-selected policies.}
\label{fig:03_dlm_mtp_refinement_budget}
\end{figure}

\paragraph{Refinement versus latency.}
At $K=16$, a second DLM refinement step lowers this cost to 4.70 ms; further refinement increases it.
With validation-selected policies on the long-output subset, DLM commits 97.1 versus 84.6 tokens within 400 ms, although MTP leads at 50 ms (\Cref{fig:03_dlm_mtp_refinement_budget}).
Refinement is thus useful when its acceptance gain repays its additional latency.

\begin{figure}[htbp]
\centering
\includegraphics[width=\linewidth]{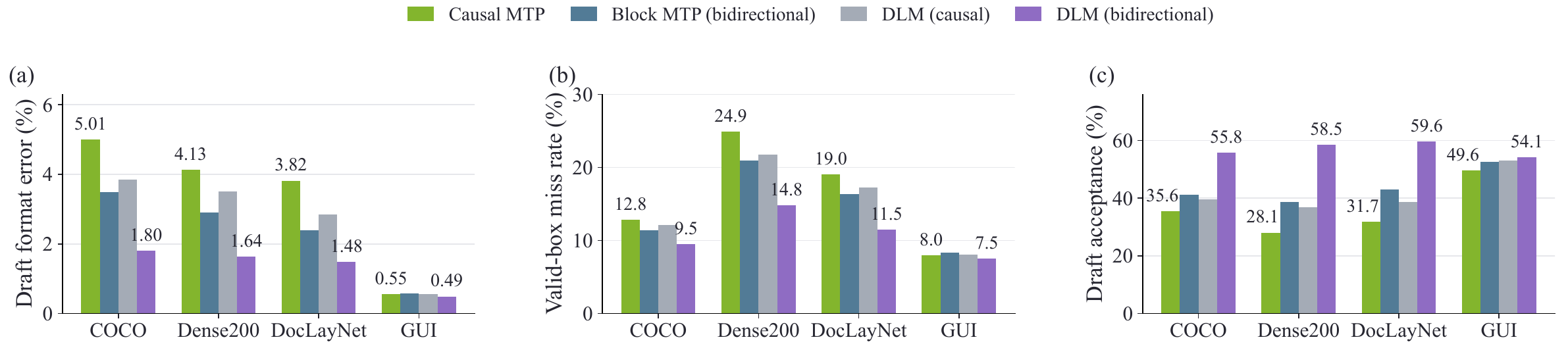}
\caption{\textbf{Draft diagnostics before verification.} Comparisons at $K=8$; both DLMs use two refinement steps. Valid-box misses count unmatched proposals, not missed ground-truth instances. GUI denotes ScreenSpot-Pro.}
\label{fig:04_dlm_mtp_task_diagnostics}
\end{figure}

\paragraph{Bidirectional attention and grounding failures.}
Bidirectional DLM reduces Dense200's unmatched valid-proposal rate from 24.89\% for causal MTP to 14.79\%; its retrained causal counterpart scores 21.74\% (\Cref{fig:04_dlm_mtp_task_diagnostics}).
This supports bidirectional attention for drafting jointly constrained spatial predictions.
Bidirectional block-MTP also improves draft quality, so the benefit is not exclusive to diffusion.

These gains concern draft efficiency and reliability; exact shared verification preserves the AR verifier's completed output.

\FinishArxivWrap

\finding{2.2}{Bidirectional diffusion with a limited number of refinement steps improves draft-generation efficiency over the tested causal MTP baseline. At $K=16$, two passes reduce cost per accepted draft token to 4.70 ms; further refinement raises the cost.}

\subsubsection{Infrastructure Analysis}
\question{2.3}{How can parallel decoding translate into practical inference speedups?}

\FinishArxivWrap
\begin{wrapfigure}{r}{0.55\textwidth}

    \centering
    \includegraphics[width=\linewidth]{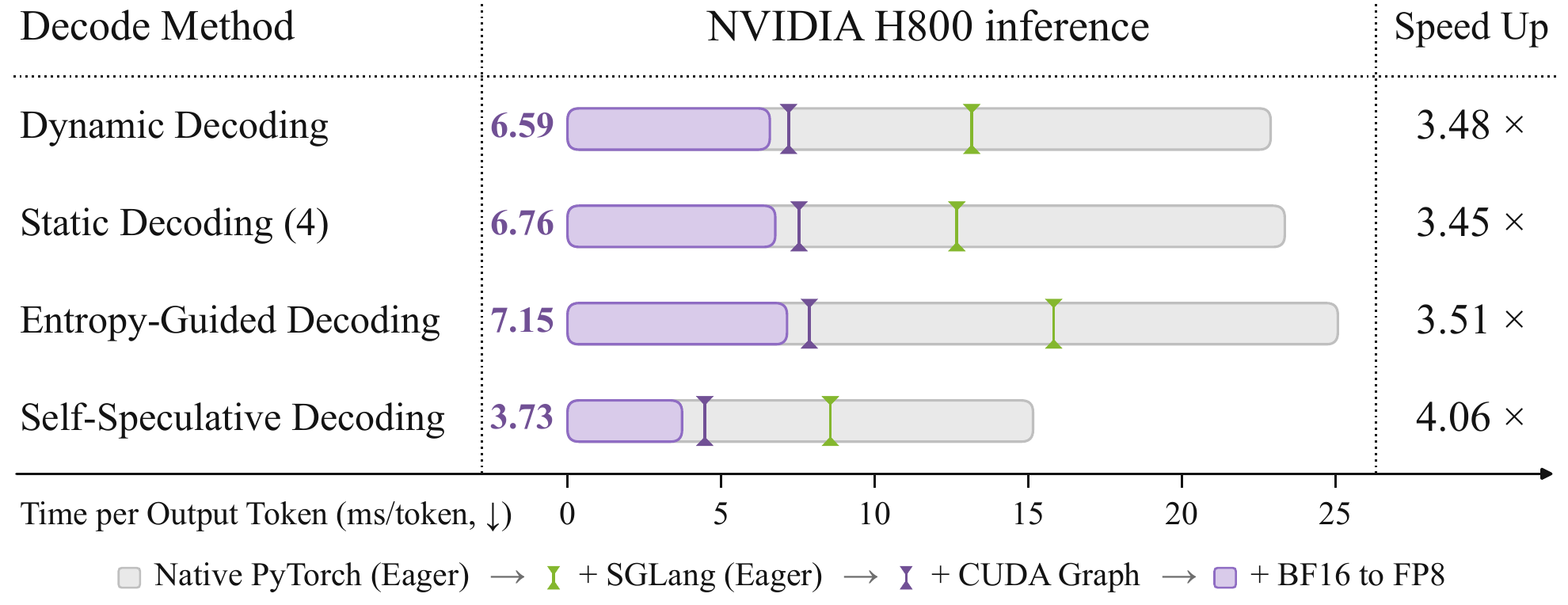}
    \caption{\textbf{Progressive inference optimization.} Time per output token ($\downarrow$) and cumulative speedup relative to each decoding mode's native PyTorch implementation.}
    \label{fig:h800-decode-methods}
    \label{fig:decode-infrastructure}
\vspace{-30pt}
\end{wrapfigure}

Progressive integration with SGLang, CUDA Graph, and FP8 reduces time per output token across all four decoding modes (\Cref{fig:decode-infrastructure}).
The complete stack yields $3.45$--$4.06\times$ speedups over each mode's native PyTorch implementation, with self-speculation reaching 3.73 ms/token.
Implementation gains are measured within each decoding mode, separately from algorithmic speedups (\Cref{app:infrastructure_analysis}).

\subsection{Ablation Studies}
\label{sec:ablation_studies}

\begin{wrapfigure}[20]{r}{0.6\textwidth}
\raggedleft
\begin{minipage}{\linewidth}

    \centering
    \includegraphics[width=\linewidth]{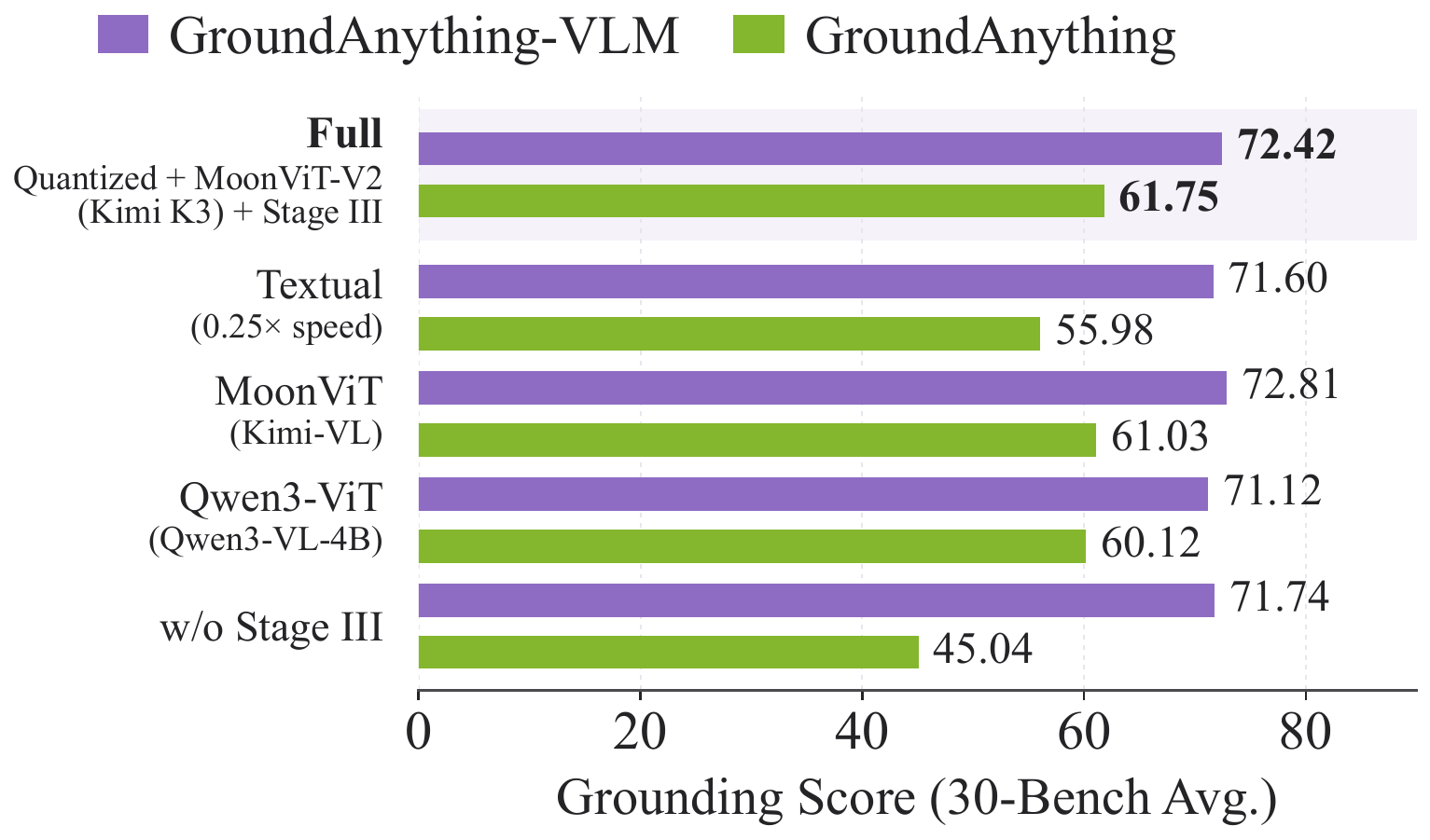}
    \caption{\textbf{Architecture and training ablations.} Coordinate representation, visual encoder, and Stage III under the ablation protocol (\Cref{app:architecture_ablations}).}
    \label{fig:grounding-model-ablations}

\end{minipage}\par

  \centering
  \BenchmarkTableFont
  \captionof{table}{\textbf{Output token efficiency.} Mean boxes and tokens per image, and tokens per box. SEED1.5-VL values are external references from \citet{rexomni}; output-length ratios are not runtime speedups.}
  \label{tab:grounding-token-efficiency}
  \begingroup
  \small
  \setlength{\tabcolsep}{5pt}
  \renewcommand{\arraystretch}{1.15}
  \definecolor{efficiencypurple}{HTML}{F5F1FA}
  \definecolor{efficiencygreen}{HTML}{F0F3E8}
  \resizebox{\linewidth}{!}{%
    \begin{NiceTabular}{@{}lcccccc@{}}
      \CodeBefore
        \rowcolor{efficiencypurple}{4}
        \rowcolor{efficiencygreen}{5}
      \Body
      \toprule
      & \multicolumn{3}{c}{\textbf{COCO}} & \multicolumn{3}{c}{\textbf{Dense200}} \\
      \cmidrule(lr){2-4}\cmidrule(lr){5-7}
      Model & Boxes/img & Tokens/img & Tokens/box & Boxes/img & Tokens/img & Tokens/box \\
      \midrule
      SEED1.5-VL\hspace{0.35em}\citep{SEED15VL} & 4.2 & 631.0 & 148.8 & 73.1 & 5446.3 & 74.5 \\
      GroundAnything-VLM & \textbf{6.1} & \textbf{46.4} & \textbf{7.6} & \textbf{87.6} & \textbf{446.8} & \textbf{5.1} \\
      GroundAnything & \textbf{6.1} & \textbf{46.4} & \textbf{7.6} & \textbf{87.6} & \textbf{446.8} & \textbf{5.1} \\
      \bottomrule
    \end{NiceTabular}%
  }
  \endgroup

\end{wrapfigure}

\paragraph{Coordinate representation.}
Quantized coordinates improve the reported ablation average by 1.43 pp for GroundAnything-VLM and 5.68 pp for GroundAnything over textual coordinates (\Cref{fig:grounding-model-ablations}).

\FinishArxivWrap

\paragraph{ViT choice.}
MoonViT-V2 gives the strongest AR and DLM scores among the tested encoders. \Cref{app:deepstack_discussion} analyzes when multi-level visual evidence can help coordinate prediction; the encoder ablation supports our choice under this recipe.

\paragraph{Training.}
Removing Stage III causes the largest DLM drop (16.62 pp).

\begin{figure}[!htbp]
  \centering
  \includegraphics[width=\linewidth]{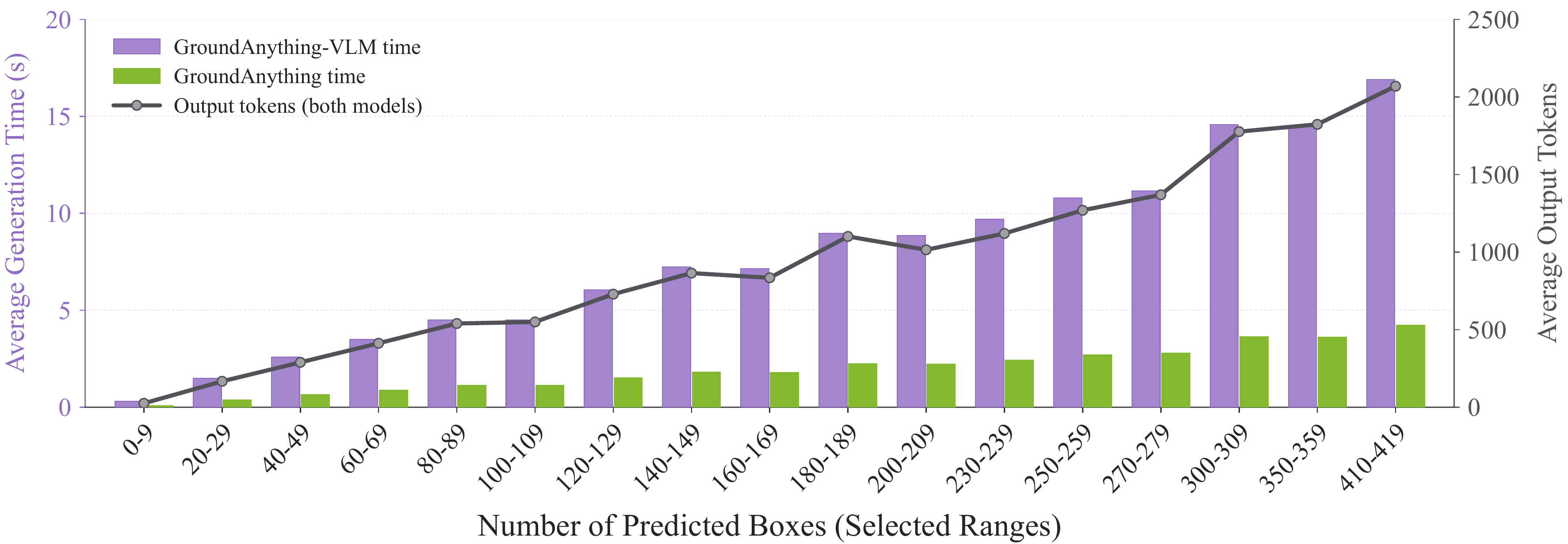}
  \caption{\textbf{Generation cost versus predicted object count.} Average generation time for GroundAnything-VLM and GroundAnything, with output-token counts across box-count ranges.}
  \label{fig:efficiency-report}
\end{figure}

\paragraph{Compact outputs and dense scenes.}
The shared output format uses 7.6 tokens/box on COCO and 5.1 on Dense200, compared with 148.8 and 74.5 for the SEED1.5-VL reference reported by \citet{rexomni} (\Cref{tab:grounding-token-efficiency}).
\Cref{fig:efficiency-report} shows how generation time and output length vary with predicted object count: GroundAnything reduces generation time across the displayed ranges, with larger absolute savings for longer outputs (\Cref{app:output_efficiency}).

\vspace{20pt}

\FinishArxivWrap
\makeatletter\global\@colnum=\c@totalnumber\relax\makeatother

\FinishArxivWrap

\subsection{Discussion}
\label{sec:experiment_discussion}

\takeaway{1}{Grounding as visual evidence extraction naturally suits bidirectional diffusion: related spatial fields share the same image evidence and can be filled jointly, then selectively refined before commitment. This connects the task's structure to parallel decoding.}

\takeaway{2}{Diffusion is a generation strategy within a Transformer backbone~\citep{FastdVLM}. Its distinctive value lies in suitable workloads, such as parallel filling and structured outputs, where joint prediction and limited refinement can repay their computational cost.}

\section{Conclusion}
\label{sec:conclusion}
We introduced GroundAnything, reconciling fast parallel decoding with broad and precise visual grounding. Structured grounding naturally suits DLMs: its predictions are jointly constrained by shared visual evidence. Diffusion complements the Transformer backbone through a generation process with distinctive strengths in parallel filling and structured outputs. Causal conditioning supports long outputs across blocks, while diffusion accelerates prediction within them. GroundAnything thus highlights a broader principle: generation should follow task structure, and spatial evidence need not be inferred in the order it is serialized.

\subsection*{AI Use Statement}
In this work, we used generative AI tools to assist with code development, to polish the writing, and to produce some of the figures. We also used generative AI models as part of the data annotation pipeline to generate pseudo-labels for model training. We did not use generative AI tools to develop the research ideas or methodology, to design or interpret the experiments, or to outline the paper. We have reviewed all AI-assisted work. We take responsibility for the final content of this work, including text, claims, data annotations, or artifacts produced with the aid of generative AI.

\subsection*{Ethics Statement}
This work does not involve human-subject studies. All datasets and evaluation benchmarks used in this work were obtained from publicly available or appropriately licensed sources in accordance with their respective terms and licenses. The experiments focus on visual grounding and structured visual understanding and do not involve real-world robotic control, autonomous driving, or interaction with people. GroundAnything is developed as a general-purpose visual grounding model for research rather than for safety-critical real-world deployment. Any future integration into embodied or autonomous systems should be subject to appropriate safety evaluation and human oversight. The authors declare no conflicts of interest.

\subsection*{Reproducibility Statement}

We will publicly release the code, model weights of GroundAnything and GroundAnything-VLM, and evaluation data to support systematic and reproducible evaluation. The release will include training and decoding configurations, inference implementations, and evaluation scripts with standardized task prompts, output parsers, and metric implementations. The shared architecture, tokenizer, visual processing, and input/output protocol are specified in \Cref{app:groundanything_architecture,app:groundanything_protocol}; the data engine and validation pipeline in \Cref{app:groundanything_data_engine}; and base VLM training, AR-to-diffusion conversion, and continued supervised fine-tuning in \Cref{app:groundanything_vlm_training,app:groundanything_conversion,app:groundanything_training}. Variant-specific GRPO configurations and task rewards are detailed in \Cref{app:groundanything_rl,app:groundanything_vlm_rl,app:groundanything_rewards}. Direct diffusion decoding, optional self-speculative decoding, and execution optimizations are described in \Cref{app:groundanything_decoding,app:groundanything_speculative,app:groundanything_execution}. Controlled DLM--MTP comparisons are documented in \Cref{app:dlm_mtp}, while speed metrics, decoding sensitivity, and infrastructure comparisons appear in \Cref{app:speed_reporting,app:infrastructure_analysis}. Architecture and training ablations and output-token efficiency analyses are provided in \Cref{app:architecture_ablations,app:output_efficiency}. The benchmark suite, evaluation metrics, reporting conventions, and complete task-level results are given in \Cref{app:grounding_results}.

\subsection*{Research Scope, Data Use, and Institutional Disclaimer}

This work originated from exploratory academic research undertaken by the project leader Qize Yu during their internship at Xpeng Inc. The project was conducted exclusively for scientific investigation and academic publication and does not involve commercial applications, product development, or commercial deployment. All data used in this project were used solely for academic research and maintained under strict segregation from the company’s commercial model development and deployment activities. No project data were used to train, fine-tune, evaluate, or otherwise support commercial models, products, or services. Internal legal review of the dataset materials was completed on September 21, 2026. This work neither uses nor discloses business data containing users’ private or personally identifiable information. The research-only scope described here does not modify or supersede the applicable terms and licenses of the source datasets.

The views, methods, findings, and conclusions presented in this paper are those of the authors and do not represent the official positions, technical direction, product roadmap, or commercial commitments of Xpeng Inc. The company’s support for this research should not be construed as endorsement of any commercial application. Neither the research findings nor their publication constitute a claim of readiness, safety, or suitability for commercial deployment.

\subsection*{Acknowledgments}

We thank Xpeng Inc. for providing computational and data resources in support of this academic research, and the data team for their assistance with data preparation and research support. We are particularly grateful to Professor Ping Luo for his guidance on the research ideas and manuscript writing. We also thank Xinghang Li, Qing Li, Baiqiao Yin, Xinyu Wei, Hang Ye, Weiyang Jin, Ruowen Zhao, Mengfan Dong, Xirui Liang, Tianshu Wu, Daiheng Gao, Jiadi You, Linhao Zhou, Qiman Wu, Ziteng Cui, Yi Zou, Wei Wei, Hanzhen Zhang and Zhuo Li for their valuable suggestions and constructive feedback.
\FinishArxivWrap
\FloatBarrier
\bibliography{paper}
\bibliographystyle{arxiv-numbered}

\clearpage
\etocdepthtag.toc{appendix}
\beginappendix
\suppressfloats[t]
\setlength{\parskip}{3pt plus 0.5pt minus 0.5pt}
\pretocmd{\section}{\FinishPlacedArxivWrap}{}{}
\pretocmd{\subsection}{\FinishPlacedArxivWrap}{}{}
\pretocmd{\subsubsection}{\FinishPlacedArxivWrap}{}{}
\makeatletter
\let\ArxivOuterFloathand\WF@floathand
\def\WF@floathand{\ifinner\else\ArxivOuterFloathand\fi}
\makeatother
\RenewDocumentEnvironment{wraptable}{O{} m O{\wrapoverhang} m}
  {\wrapfloat{table}[#1]{r}[#3]{#4}}{\endwrapfloat}
\label{app:appendix}

\ifdefined\BenchmarkTableFont\else
\newcommand{\BenchmarkTableFont}{\normalfont}
\newcommand{\BenchHead}[1]{\begin{tabular}{@{}c@{}}#1\end{tabular}}
\colorlet{benchmarktype}{black!8}
\definecolor{benchmarkpurple}{HTML}{F5F1FA}
\definecolor{benchmarkgreen}{HTML}{F0F3E8}
\fi

\ifdefined\GPIBoxHelpersLoaded\else
\def\GPIBoxHelpersLoaded{1}
\ifdefined\tcolorbox
  \providecommand{\finding}[2]{%
    \begin{tcolorbox}[colback=blue!4,colframe=blue!55!black,
      boxrule=0.6pt,arc=2pt,left=5pt,right=5pt,top=3pt,bottom=3pt,
      before skip=5pt,after skip=5pt]\small
      \textbf{\textit{Finding #1:}} #2
    \end{tcolorbox}}
  \providecommand{\takeaway}[2]{%
    \begin{tcolorbox}[colback=blue!4,colframe=blue!55!black,
      boxrule=0.6pt,arc=2pt,left=5pt,right=5pt,top=3pt,bottom=3pt,
      before skip=5pt,after skip=5pt]\small
      \textbf{\textit{Takeaway #1:}} #2
    \end{tcolorbox}}
\else
  \providecommand{\finding}[2]{\begin{gpiquestions}\textbf{\textit{Finding #1:}} #2\end{gpiquestions}}
  \providecommand{\takeaway}[2]{\begin{gpiquestions}\textbf{\textit{Takeaway #1:}} #2\end{gpiquestions}}
\fi
\fi

\section{Model and Training Details}
\label{app:additional_details}

\providecommand{\GAcode}[1]{\texttt{\detokenize{#1}}}

\subsection{Input/Output Protocol}
\label{app:groundanything_protocol}

GroundAnything and GroundAnything-VLM share a structured interface: the query specifies what evidence to extract, and each response entry associates a semantic identifier with spatial coordinates. The geometry is task-dependent, while the enclosing grammar remains unchanged, following the unified grounding formulation of \citet{rexomni}.

\paragraph{Query templates.}
\Cref{tab:groundanything_prompts} lists the task instructions. Bracketed terms are placeholders. A category list joins the requested names with \GAcode{</c>}, without intervening spaces; for example, \GAcode{person</c>car}. Visual reference boxes use the same coordinate tokens as outputs. The image and instruction form the user turn; the structured response forms the assistant turn.

\begin{table}[hb]
\centering\small
\setlength{\tabcolsep}{5pt}
\renewcommand{\arraystretch}{1.18}
\caption{Task-specific input instructions. Category names and referring expressions are preserved in the response identifiers.}
\label{tab:groundanything_prompts}
\begin{tabular}{p{0.20\linewidth}p{0.73\linewidth}}
\hline
Task & Instruction \\
\hline
Category grounding & Locate all the instances that match the following categories: [CATEGORIES]. \\
Referring grounding & Locate the target referred to by the following description: [EXPRESSION]. \\
Category pointing & Point to: [CATEGORIES]. \\
Referring pointing & Point to the target referred to by the following description: [EXPRESSION]. \\
GUI grounding & Point to the UI element to click for the following instruction: [INSTRUCTION]. \\
OCR & OCR task detect all the text in box format. \\
Layout grounding & Detect all document layout elements that match the following categories: [CATEGORIES]. \\
Visual prompting & Given reference boxes \GAcode{<|box_start|>}[BOXES]\GAcode{<|box_end|>} indicating one or more objects, find all similar objects in the image and output their bounding boxes. \\
\hline
\end{tabular}
\end{table}

\paragraph{Response grammar.}
The entry template and illustrative payloads are shown below. Displayed line breaks are for readability and are omitted in the serialized response.
\begin{center}
\begin{tcolorbox}[enhanced,breakable,width=\dimexpr0.94\linewidth+2\fboxsep+2\fboxrule\relax,colback=white,colframe=black,boxrule=\fboxrule,arc=0pt,boxsep=\fboxsep,left=0pt,right=0pt,top=0pt,bottom=0pt]
\small
\textbf{Entry template}\\[2pt]
\GAcode{<|object_ref_start|>}[IDENTIFIER]\GAcode{<|object_ref_end|>}\\
\GAcode{<|box_start|>}[PAYLOAD]\GAcode{<|box_end|>}

\medskip
\textbf{Boxes: two instances of one category}\\[2pt]
\GAcode{<|object_ref_start|>person<|object_ref_end|>}\\
\GAcode{<|box_start|><10><20><30><40>,<50><60><70><80><|box_end|>}

\medskip
\textbf{Point: one target location}\\[2pt]
\GAcode{<|object_ref_start|>button<|object_ref_end|>}\\
\GAcode{<|box_start|><512><384><|box_end|>}

\medskip
\textbf{OCR: transcription as identifier}\\[2pt]
\GAcode{<|object_ref_start|>OPEN<|object_ref_end|>}\\
\GAcode{<|box_start|><100><200><300><260><|box_end|>}

\medskip
\textbf{Explicitly queried but absent category}\\[2pt]
\GAcode{<|object_ref_start|>car<|object_ref_end|>}\\
\GAcode{<|box_start|>None<|box_end|>}
\end{tcolorbox}
\end{center}

Boxes use four atomic coordinate tokens in $(x_1,y_1,x_2,y_2)$ order; points use two in $(x,y)$ order. Instances sharing an identifier occupy one wrapper, with tuples separated by commas without spaces. Distinct entries are separated by a comma and a space. Ordered point sequences retain their temporal order. The ordinary text \GAcode{None} denotes an absent queried target, not a padding element. OCR transcriptions occupy the identifier field rather than a separate text channel.

\paragraph{Coordinates and canonicalization.}
Coordinates are normalized by image width and height and quantized to \GAcode{<0>}--\GAcode{<999>}. For annotations on a $[0,1000]$ grid, the conversion is
\begin{equation}
q(v)=\left\lfloor\frac{999v+500}{1000}\right\rfloor.
\end{equation}
Input \texttt{xywh} boxes are converted to \texttt{xyxy}; polygons used for box supervision become axis-aligned enclosing boxes. Non-finite or out-of-range coordinates and inverted or degenerate boxes fail validation. Box instances are stably sorted by $x_1$, and unordered points by $x$; semantic sequence order is never replaced by spatial sorting.

\paragraph{Structural tokens and termination.}
\Cref{tab:groundanything_tokens} distinguishes entry boundaries from response termination. Structural and coordinate tokens are retained when decoding text for the parser. The diffusion mask is an internal input token and is excluded from the generatable vocabulary.

\begin{table}[hb]
\centering\small
\setlength{\tabcolsep}{5pt}
\renewcommand{\arraystretch}{1.1}
\caption{Grounding and multimodal token conventions.}
\label{tab:groundanything_tokens}
\begin{tabular}{p{0.47\linewidth}p{0.46\linewidth}}
\hline
Token(s) & Function \\
\hline
\GAcode{<|object_ref_start|>}, \GAcode{<|object_ref_end|>} & Delimit a semantic identifier \\
\GAcode{<|box_start|>}, \GAcode{<|box_end|>} & Delimit a geometric payload \\
\GAcode{<0>}--\GAcode{<999>} & Atomic quantized coordinates \\
\GAcode{</c>} & Separate requested categories \\
\GAcode{<|vision_start|>}, \GAcode{<|vision_end|>} & Delimit visual input \\
\GAcode{<|image_pad|>} & Image placeholder in the input template \\
\GAcode{<|im_start|>}, \GAcode{<|im_end|>} & Turn boundaries; the latter terminates the response \\
\GAcode{|<MASK>|} & Internal diffusion mask \\
\hline
\end{tabular}
\end{table}

\subsection{Architecture, Tokenizer, and Visual Processing}
\label{app:groundanything_architecture}

The variants share a MoonViT-V2 (Kimi K3) visual encoder, a multimodal projector, and a Qwen3-4B language architecture. All language layers use full attention and one-dimensional rotary position indices. Dynamic image processing supplies at most 1024 projected visual tokens, corresponding to 4096 patches before $2\times2$ aggregation; channel means and standard deviations are both $(0.5,0.5,0.5)$. \Cref{tab:groundanything_architecture_details} gives the module specifications.

\begin{table}[htbp]
\centering\small
\setlength{\tabcolsep}{5pt}
\renewcommand{\arraystretch}{1.1}
\caption{Shared backbone architecture. Positional capacity is independent of the training sequence lengths.}
\label{tab:groundanything_architecture_details}
\begin{tabular}{p{0.25\linewidth}p{0.68\linewidth}}
\hline
Component & Configuration \\
\hline
Vision encoder & 27 layers; width 1024; FFN width 4096; 12 heads; QKV width 1536; patch size 14 \\
Projector & LayerNorm(1024), $2\times2$ aggregation, bias-free linear layers $4096\to4096\to2560$ with GELU, then RMSNorm(2560); normalization $\epsilon=10^{-5}$ \\
Language backbone & 36 layers; width 2560; FFN width 9728; 32 query heads and 8 KV heads; head dimension 128 \\
Language numerics & SiLU; attention dropout 0; RMSNorm $\epsilon=10^{-6}$; RoPE base $5{,}000{,}000$ \\
Positional capacity & 262,144 positions \\
\hline
\end{tabular}
\end{table}

\sbox{\ArxivNaturalTable}{%
\small
\setlength{\tabcolsep}{5pt}
\begin{tabular}{lrr}
\hline
Module & GroundAnything-VLM & GroundAnything \\
\hline
Vision encoder & 401,214,464 & 401,214,464 \\
Projector & 27,267,584 & 27,267,584 \\
Decoder excluding vocabulary & 3,633,511,936 & 3,633,511,936 \\
Input embedding & 390,835,200 & 390,837,760 \\
Output head & 390,835,200 & 390,837,760 \\
\hline
Total & 4,843,664,384 & 4,843,669,504 \\
\hline
\end{tabular}%
}
\FinishArxivWrap
\begin{wraptable}{R}{\wd\ArxivNaturalTable}
\centering\small
\setlength{\tabcolsep}{5pt}
\caption{Parameter counts. The 4B designation refers to the language backbone.}
\label{tab:groundanything_parameter_counts}
\usebox{\ArxivNaturalTable}
\end{wraptable}

\paragraph{Vocabulary and parameter sharing.}
GroundAnything-VLM extends the 151,669-entry base vocabulary with 1000 coordinate tokens and \GAcode{</c>}, giving 152,670 entries. Coordinate IDs are 151669--152668; the category delimiter has ID 152669. GroundAnything adds \GAcode{|<MASK>|} at ID 152670, giving 152,671 entries. EOS and padding have distinct IDs, 151645 and 151643; the image, vision-start, and vision-end IDs are 151655, 151652, and 151653.

Input embeddings and output heads are untied. Within GroundAnything, causal and diffusion modes share all model weights. The new mask row is initialized separately in each vocabulary matrix using the FP32 mean of the rows for \GAcode{<|im_start|>}, \GAcode{<|im_end|>}, \GAcode{<|vision_start|>}, and \GAcode{<|vision_end|>}, then cast to the model dtype. This adds 5120 parameters. The complete multimodal model has approximately 4.844B parameters (\Cref{tab:groundanything_parameter_counts}).

\subsection{Expert-Driven Data Engine}
\label{app:groundanything_data_engine}

\paragraph{Field-level fusion.}
Complementary teachers provide semantic, localization, segmentation, text, interface, and layout evidence. Category discovery precedes category-conditioned localization; category names are normalized while attributes needed for referring expressions are retained. Predictions are mapped back to the original image and aligned only when their query scope and annotation granularity agree. Fusion operates on individual fields, allowing reliable text or geometry to be retained without accepting an entire inconsistent annotation.

\paragraph{Validation and coverage.}
Task-specific checks assess geometric validity, semantic consistency, text--region agreement, and cross-view support. Each dependency has a validation state $a(f)\in\{\mathrm{pass},\mathrm{fail},\mathrm{pending}\}$; query-level coverage $c(I,Q)$ uses the same state space. Coverage checks whether the required query scope has been examined and relevant unresolved proposals and matching conflicts have been addressed. Unresolved dependencies trigger complementary teachers or targeted local observations.

\paragraph{Required dependencies and acceptance.}
Let $\mathcal{F}(I,Q,Y)$ contain the fields and semantic facts required to validate candidate $Y$, including query--target correspondence. The task protocol and query semantics determine these requirements, which are then instantiated for the candidate. Missing required fields remain in $\mathcal{F}$ with status $\mathrm{pending}$: omitting a field cannot remove its validation requirement. Let $h(Q)\in\{0,1\}$ indicate whether the query requires exhaustive instance coverage. Acceptance requires
\begin{equation}
\mathcal{A}(I,Q,Y)=
\mathbf{1}\!\left[\forall f\in\mathcal{F}(I,Q,Y),\ a(f)=\mathrm{pass}\right]
\mathbf{1}\!\left[h(Q)=0\ \lor\ c(I,Q)=\mathrm{pass}\right],
\end{equation}
so every required dependency must pass, and exhaustive queries additionally require validated coverage. For example, validating the returned boxes does not establish that every queried instance has been found; an all-instance query remains unaccepted while coverage is pending.

\paragraph{Empty-target claims.}
An answer with no instances still requires verification of its query--answer relation. Its dependency set explicitly includes a query-scoped absence fact $f_{\mathrm{abs}}\in\mathcal{F}(I,Q,Y)$, so an empty prediction never yields an empty dependency set. An unsupported absence claim remains pending; contrary evidence makes it fail. It passes only when absence is verified and $c(I,Q)=\mathrm{pass}$, even when $h(Q)=0$. Missing outputs, failed annotation runs, and unresolved empty predictions do not constitute evidence of absence. These requirements prevent empty predictions from being accepted through a vacuously true check over an empty set.

Derived box, point, referring, and visual-prompt supervision preserves its dependencies on the validated source evidence; crops and coordinate transforms remain consistent with visible content.

\paragraph{Expert iteration.}
Accepted annotations train a unified grounding expert, which then proposes annotations for further validation. Difficult cases receive targeted crops or complementary teacher evidence. New evidence can correct or retire previous labels, with versioned annotation snapshots keeping each training input fixed. This loop improves both coverage and field consistency rather than merely accumulating predictions.

\subsection{Training Pipeline}
\label{app:groundanything_training}

GroundAnything follows four training phases: Base VLM Training (Pretrain~1), Coordinate Alignment (Pretrain~2, also termed SFT), AR-to-Diffusion conversion, and reinforcement learning (RL). GroundAnything-VLM follows three phases: Base VLM Training, Coordinate Alignment, and RL. The Stage~1--4 numbering below refers specifically to pretraining and alignment: Stages~1--3 belong to Pretrain~1, while Stage~4 is Pretrain~2. \Cref{tab:groundanything_training_stages} summarizes the two pipelines and their relation to \Cref{fig:groundanything-sft-rl}.

\begin{table}[htbp]
\centering\small
\setlength{\tabcolsep}{5pt}
\renewcommand{\arraystretch}{1.12}
\caption{Training pipelines. GroundAnything has four phases; GroundAnything-VLM has three and proceeds directly from coordinate alignment to RL.}
\label{tab:groundanything_training_stages}
\begin{tabular}{@{}p{0.30\linewidth}p{0.31\linewidth}p{0.31\linewidth}@{}}
\hline
Training phase & GroundAnything-VLM & GroundAnything \\
\hline
Base VLM Training (Pretrain~1) & Stages~1--3; causal cross-entropy & Stages~1--3; causal cross-entropy \\
Coordinate Alignment (Pretrain~2 / SFT) & Stage~4; assistant-only spatial cross-entropy & Stage~4; assistant-only spatial cross-entropy \\
AR-to-Diffusion conversion & --- & $0.5\mathcal{L}_{\mathrm{MDM}}+0.5\mathcal{L}_{\mathrm{AR}}$; shared weights \\
Reinforcement learning (RL) & Causal GRPO & Causal GRPO; shared AR/diffusion weights \\
\hline
\end{tabular}
\end{table}

\subsection{Base VLM Training (Pretrain 1)}
\label{app:groundanything_vlm_training}

\paragraph{Stage 1: vision--language connector alignment.}
The vision encoder and language backbone are frozen, and only the projector is trained on image-caption data. Assistant-only causal cross-entropy aligns the visual features with the language input space.

\paragraph{Stage 2: joint multimodal pretraining.}
All modules are unfrozen and jointly trained on pure text and general image--text data. Pure-text examples supervise all valid causal text targets; multimodal examples supervise assistant responses, excluding visual inputs, prompts, and masked template positions. The text and multimodal losses are normalized separately over their valid target tokens and then summed, $\mathcal{L}_{\mathrm{mix}}=\mathcal{L}_{\mathrm{text}}+\mathcal{L}_{\mathrm{vlm}}$. Counts are aggregated across data-parallel workers so that each domain contributes its global token mean. Packing preserves independent subsequence boundaries and supervision masks.

\paragraph{Stage 3: general visual and video understanding.}
All modules continue training on general visual question answering and instruction-following data, image captions, and video data with assistant-only causal cross-entropy. The configured context and packing lengths are both 32K. This stage has no separate pure-text quota or dedicated spatial-specialization mixture; spatial supervision is concentrated in Pretrain~2. Aggregate base-pretraining exposure exceeds 700B tokens.

\subsection{Coordinate Alignment (Pretrain 2 / SFT)}
\label{app:groundanything_coordinate_alignment}

\paragraph{Stage 4: spatial specialization.}
Initialized from the Stage~3 checkpoint, coordinate alignment updates the vision encoder, projector, and language parameters on eight spatial task groups: detection, GUI grounding, referring grounding, referring pointing, OCR, document layout, dense pointing and counting, and visual prompting. Boxes, points, semantic identifiers, and protocol markers are supervised jointly as assistant-response tokens. The coordinate tokens \GAcode{<0>} through \GAcode{<999>} use the same causal cross-entropy objective as the other valid response tokens. This single alignment phase is the supervised fine-tuning (SFT) phase referred to in the main text. Its context and packing lengths are 8K.

\Cref{tab:groundanything_base_training} gives the configured recipe for both pretraining phases. After coordinate alignment, GroundAnything-VLM proceeds to RL, while GroundAnything first undergoes the AR-to-Diffusion conversion in \Cref{app:groundanything_conversion}.

\begin{table}[htbp]
\centering\small
\setlength{\tabcolsep}{3pt}
\renewcommand{\arraystretch}{1.12}
\caption{Configured pretraining and alignment recipe. Base VLM Training (Pretrain~1) comprises Stages~1--3; Coordinate Alignment (Pretrain~2 / SFT) is Stage~4. Context and packing lengths count complete multimodal sequences.}
\label{tab:groundanything_base_training}
\label{tab:groundanything_sft_recipe}
\label{tab:groundanything_vlm_sft}
\begin{tabular}{@{}p{0.27\linewidth}p{0.165\linewidth}p{0.165\linewidth}p{0.165\linewidth}p{0.165\linewidth}@{}}
\hline
 & \multicolumn{3}{c}{Base VLM Training (Pretrain~1)} & Pretrain~2 \\
Setting & Stage~1 & Stage~2 & Stage~3 & Stage~4 \\
\hline
Trainable modules & Projector & All & All & All \\
Language peak LR & Frozen & $10^{-5}$ & $10^{-5}$ & $10^{-5}$ \\
Vision peak LR & Frozen & $10^{-6}$ & $10^{-6}$ & $10^{-6}$ \\
Projector peak LR & $5\times10^{-5}$ & $10^{-6}$ & $10^{-6}$ & $10^{-6}$ \\
Per-device batch / accumulation & 6 / 1 & 2 / 1 & 2 / 1 & 2 / 1 \\
GPUs / global batch & 128 / 768 & 128 / 256 & 128 / 256 & 128 / 256 \\
Context / packing & 8K / 8K & 8K / 8K & 32K / 32K & 8K / 8K \\
Weight decay / gradient clipping & 0 / 1.0 & 0.1 / 1.0 & 0.1 / 1.0 & 0.1 / 1.0 \\
LR schedule / minimum ratio & Cosine / 2\% & Cosine / 10\% & Cosine / 10\% & Cosine / 10\% \\
Warmup / configured budget & 3\% / 1 epoch & 3\% / 1 epoch & 3\% / 1 epoch & 3\% / 1 epoch \\
\hline
\end{tabular}
\end{table}

All four pretraining/alignment stages use BF16, ZeRO-1, AdamW with $(\beta_1,\beta_2)=(0.9,0.95)$, FlashAttention, and activation checkpointing. Global batches count packed sequences and equal the number of GPUs times the per-device batch times gradient accumulation. Here 8K and 32K denote 8192 and 32768 tokens, respectively, and one epoch denotes the configured training budget. Language parameters include the untied input embedding and output head.

\subsection{AR-to-Diffusion Conversion (GroundAnything Only)}
\label{app:groundanything_conversion}

GroundAnything is initialized from the autoregressive checkpoint after coordinate alignment. Direct multimodal conversion~\citep{FastdVLM} retains its grounding vocabulary and output grammar. Let $x=[V,P,Y]$ contain visual embeddings $V$, the text prompt $P$ encoding query $Q$, and response $Y$. Supervised response spans are partitioned into consecutive blocks of $B=32$ tokens. Blocks may cross geometric-tuple boundaries, stop at response-span boundaries, and have a shorter final block; they require no semantic padding.

\paragraph{Complementary corruption.}
For each response block $b$, sample $t_b\sim\mathcal{U}(0,1)$. For every supervised non-EOS position $i$ in that block, draw $m_i^{(1)}\sim\operatorname{Bernoulli}(t_b)$ and set $m_i^{(2)}=1-m_i^{(1)}$. View $a$ replaces the target by \GAcode{|<MASK>|} if $m_i^{(a)}=1$, giving $w^{(a)}=[P,\widetilde{Y}^{(a)}]$. EOS is masked in both views. Images, prompts, headers, unsupervised turns, and empty thinking prefixes are not corrupted.

\paragraph{Attention visibility.}
The joint input $[w^{(1)};w^{(2)};x]$ shares one clean multimodal stream and one visual encoding; each noisy view retains an uncorrupted prompt copy. Within a view, a noisy response block attends bidirectionally to itself and reads the clean visual/prompt context and strictly preceding clean response blocks. It cannot read current or future clean response targets or the other noisy view. Clean tokens use token-level causal attention and cannot attend to either noisy view. \Cref{fig:groundanything-training-attention} illustrates one view with two-token blocks for clarity.

\paragraph{Joint objective.}
For masked target sets $\mathcal{M}^{(1)}$ and $\mathcal{M}^{(2)}$, and valid clean response targets $\mathcal{T}$, the two losses are
\begin{align}
\mathcal{L}_{\mathrm{MDM}}
&=-\frac{\sum_{a=1}^{2}\sum_{i\in\mathcal{M}^{(a)}}
\log p_\theta(y_i\mid x,w^{(a)};\mathcal{A}_a)}
{|\mathcal{M}^{(1)}|+|\mathcal{M}^{(2)}|},\\
\mathcal{L}_{\mathrm{AR}}
&=-\frac{1}{|\mathcal{T}|}\sum_{i\in\mathcal{T}}
\log p_\theta(y_i\mid V,P,y_{<i}).
\end{align}
Here $\mathcal{A}_a$ enforces the visibility rules above, so writing $x$ as an input does not expose the clean target. The MDM mean includes both masked EOS targets and has no inverse-noise weighting. Both losses retain the next-token shift: hidden position $i-1$ predicts target $i$. Prompts, visual positions, headers, padding, unsupervised turns, and empty thinking prefixes are excluded. Valid target losses are averaged across the training batch and combined as in \Cref{eq:groundanything-conversion-loss}.

\paragraph{Conversion optimization.}
Within this conversion phase, optimization has two substages: Phase I freezes the vision encoder and trains the projector and language parameters; Phase II updates all modules. Both retain the fixed block size and joint objective. The optimizer and scheduler restart between phases. \Cref{tab:groundanything_conversion_recipe} reports the settings.

\begin{table}[htbp]
\centering\small
\setlength{\tabcolsep}{5pt}
\renewcommand{\arraystretch}{1.1}
\caption{GroundAnything conversion configuration.}
\label{tab:groundanything_conversion_recipe}
\begin{tabular}{p{0.36\linewidth}p{0.27\linewidth}p{0.27\linewidth}}
\hline
Setting & Phase I & Phase II \\
\hline
GPUs & 256 & 256 \\
Per-device / global batch & 32 / 8192 & 32 / 8192 \\
Clean sequence length & 4096 & 4096 \\
Trainable modules & Projector and language & All \\
Language / projector peak LR & $2\times10^{-5}$ / $2\times10^{-5}$ & $2\times10^{-5}$ / $2\times10^{-5}$ \\
Vision peak LR & Frozen & $2\times10^{-6}$ \\
Gradient checkpointing & Language & Language and vision \\
Seed / data seed & 32 / 32 & 42 / 42 \\
\hline
\end{tabular}
\end{table}

Conversion uses BF16, ZeRO-1, one gradient-accumulation step, and AdamW with betas $(0.9,0.95)$, epsilon $10^{-8}$, weight decay 0.1, and gradient clipping at 1.0. Both phases use cosine decay with 3\% warmup.

\subsection{Reinforcement Learning (RL): Causal GRPO}
\label{app:groundanything_rl}

RL is the final training phase for both variants, following coordinate alignment for GroundAnything-VLM and AR-to-Diffusion conversion for GroundAnything. GroundAnything uses causal attention for both rollout generation and likelihood evaluation. For each prompt, $G=8$ responses are sampled at temperature $T=0.7$. Before updating, a no-gradient teacher-forcing pass records detached old-policy log-probabilities; a second pass computes current-policy gradients. Both use FP32 log-softmax at the rollout temperature. These are exact likelihoods of the causal policy, not likelihoods of a diffusion denoising trajectory.

\paragraph{Advantage normalization.}
For a scalar component $u$ within one prompt group, define
\begin{equation}
Z_G(u_i)=\frac{u_i-\bar{u}}{\sqrt{G^{-1}\sum_{j=1}^{G}(u_j-\bar{u})^2}+10^{-6}},
\qquad \bar{u}=G^{-1}\sum_{j=1}^{G}u_j.
\label{eq:groundanything-reward-normalization}
\end{equation}
Each active reward component is normalized before weighting, and their sum is normalized again within the same group:
\begin{equation}
A_i=Z_G\!\left(\sum_{c\in\mathcal{C}_{\mathrm{task}}}w_c Z_G(R_{c,i})\right).
\label{eq:groundanything-grounding-advantage}
\end{equation}
Box rewards have weights $(0.7,0.3)$; OCR and pointing each supply one complete scalar reward with unit weight. Internal OCR subterms are not normalized separately. Constant groups produce zero advantage.

\paragraph{Policy objective.}
Let $\pi_{\theta,T}$ denote the causal policy at temperature $T$, and let $\mathcal{T}_i$ contain response tokens including EOS, excluding prompts and post-EOS padding. Expanding the response-wise averaging in \Cref{eq:groundanything-grpo-main}, the update minimizes
\begin{align}
r_{i,t}(\theta)&=\frac{\pi_{\theta,T}(y_{i,t}\mid I,Q,y_{i,<t})}
{\pi_{\mathrm{old},T}(y_{i,t}\mid I,Q,y_{i,<t})},\\
\mathcal{L}_{\mathrm{GRPO}}&=-\frac{1}{G}\sum_{i=1}^{G}\frac{1}{|\mathcal{T}_i|}
\sum_{t\in\mathcal{T}_i}\min\!\left(r_{i,t}A_i,
\operatorname{clip}(r_{i,t},1-\epsilon,1+\epsilon)A_i\right),
\label{eq:groundanything-grpo}
\end{align}
with $\epsilon=0.2$. There is one update per rollout and no reference-policy KL term. The vision encoder is frozen; the projector and language parameters are updated. Causal and diffusion attention modes subsequently reuse these parameters without changing the output protocol.

\subsection{GroundAnything-VLM Reinforcement Settings}
\label{app:groundanything_vlm_rl}

GroundAnything-VLM uses the same task-reward definitions but a distinct optimization configuration. It freezes both the vision encoder and projector and retains a frozen supervised reference with KL coefficient 0.02. Active reward components are standardized within each prompt group using the sample standard deviation and epsilon $10^{-8}$; after weighted summation, advantages are standardized across the complete response batch. GroundAnything instead uses population standard deviations and a second within-group normalization. \Cref{tab:groundanything_rl_recipe} keeps these recipes separate.

\begin{table}[htbp]
\centering\small
\setlength{\tabcolsep}{4pt}
\renewcommand{\arraystretch}{1.12}
\caption{Variant-specific GRPO settings. Global batches count generated responses, not distinct prompts.}
\label{tab:groundanything_rl_recipe}
\label{tab:groundanything_vlm_rl}
\begin{tabular}{p{0.35\linewidth}p{0.27\linewidth}p{0.27\linewidth}}
\hline
Setting & GroundAnything & GroundAnything-VLM \\
\hline
GPUs & 256 & 256 \\
Per-device response batch / accumulation & 1 / 1 & 1 / 1 \\
Responses per prompt & 8 & 8 \\
Prompt / response batch & 32 / 256 & 32 / 256 \\
Updates per rollout & 1 & 1 \\
Learning rate & $2\times10^{-6}$ & $5\times10^{-7}$ \\
Schedule / warmup & Constant / none & Constant / none \\
AdamW betas / epsilon & $(0.9,0.999)$ / $10^{-8}$ & $(0.9,0.999)$ / $10^{-8}$ \\
Weight decay / gradient clipping & 0 / 1.0 & 0.01 / 1.0 \\
Policy clipping & 0.2 & 0.2 \\
Reference-policy KL & No reference & 0.02 \\
Trainable modules & Projector and language & Language \\
Precision / sharding & BF16 / ZeRO-1 & BF16 / ZeRO-2 \\
Seed & 20260902 & 20260812 \\
\hline
\end{tabular}
\end{table}

GroundAnything maintains FP32 optimizer states. Its rollout and gradient passes use the same causal policy; the inference implementations described below do not alter this training factorization.

\subsection{Task Rewards}
\label{app:groundanything_rewards}

The rewards below evaluate complete structured responses. We write $n$ and $m$ for the numbers of predictions and reference instances, respectively, and $\operatorname{clip}_{[0,1]}$ for clipping to the unit interval. Content scores below assume nonempty references; missing predictions receive zero content agreement.

\subsubsection{Box Grounding}
\label{app:groundanything_box_reward}

\paragraph{Set completeness.}
For each reference box $g_j$, select the predicted box with maximum IoU before checking its label:
\begin{equation}
i^*(j)=\arg\max_i\operatorname{IoU}(b_i,g_j),\qquad
s_j=\operatorname{IoU}(b_{i^*(j)},g_j)\,
\mathbf{1}[\ell_{i^*(j)}=\ell_j].
\end{equation}
With $S=\sum_j s_j$, soft precision and recall are $P=S/n$ and $R=S/m$, and
\begin{equation}
R_{\mathrm{set}}=\frac{2PR}{P+R+10^{-8}}.
\end{equation}
This GT-wise matching permits a prediction to support multiple references. Consequently, $R_{\mathrm{set}}$ is a completeness proxy rather than a bounded one-to-one F1 metric. It is combined with a stricter reward that penalizes duplicate and oversized predictions.

\paragraph{Strict localization.}
Let $\bar{F}$ average localization F1 at IoU thresholds $0.50$, $0.75$, and $0.95$. The strict reward is
\begin{equation}
\begin{split}
R_{\mathrm{strict}}=\operatorname{clip}_{[0,1]}\bigl(&0.10R_{\mathrm{fmt}}+0.10R_{\mathrm{count}}
+0.50\bar{F}+0.25R_{\mathrm{IoU}}+0.05R_{\mathrm{sort}}\\
&-0.07P_{\mathrm{big}}-0.03P_{\mathrm{dup}}\bigr).
\end{split}
\end{equation}
The auxiliary terms score protocol validity, instance-count agreement, matched IoU, and canonical ordering. Penalties discourage unsupported large boxes and duplicate predictions. The set and strict rewards are normalized separately before combination, rather than treating their raw weighted sum as a single reward.

\subsubsection{OCR}
\label{app:groundanything_ocr_reward}

\paragraph{Text--geometry matching.}
OCR uses one-to-one Hungarian matching. The main text key applies Unicode NFKC normalization and case folding and retains alphanumeric characters; symbol-only strings retain codepoint-based keys. Let $E_{ij}$ be one minus normalized Levenshtein distance between the resulting keys, and $I_{ij}$ the box IoU. For an affinity matrix $A$, define
\begin{equation}
F(A)=\frac{2}{n+m}\max_{\mathcal{M}}\sum_{(i,j)\in\mathcal{M}}A_{ij},
\end{equation}
where $\mathcal{M}$ is a one-to-one assignment. Hard affinity requires equal text keys and $I_{ij}\geq\tau$. Write its score as $H_\tau$ and its average over $\tau\in\{0.50,0.55,\ldots,0.95\}$ as $\bar{H}$. The soft affinity and single-reference score are
\begin{align}
A^{\mathrm{soft}}_{ij}
&=\sqrt{I_{ij}}\,E_{ij}\,
\mathbf{1}[I_{ij}\geq0.10,\ E_{ij}\geq0.20],\\
V(P,T)&=\operatorname{clip}_{[0,1]}
\left(0.45\bar{H}+0.35F(A^{\mathrm{soft}})+0.10H_{0.50}+0.10C\right),
\end{align}
with count agreement $C=\min(n,m)/\max(n,m)$.

\paragraph{Word/line consistency.}
Let $V_{\mathrm{hi}}$ and $V_{\mathrm{lo}}$ be the higher and lower scores against complete word- and line-level references. For a region with alternative granularities, local agreement selects the better \emph{complete} representation, not independently chosen fragments. Singleton consensus uses the maximum text and box similarities among the reference alternatives, followed by one-to-one assignment, with score $0.60\bar{H}^{*}+0.40S^{*}$. Aggregating local agreements gives $\Gamma$, with lower weights for text-conflict regions. The global and local terms are combined as
\begin{equation}
D=\alpha V_{\mathrm{hi}}+(1-\alpha)V_{\mathrm{lo}},\qquad
Q_g=\beta D+(0.90-\beta)\Gamma,
\end{equation}
with $(\alpha,\beta)=(0.90,0.35)$ for conflicting granularities and $(0.75,0.50)$ otherwise. Define
\[
m_g=\max(|T_{\mathrm{word}}|,|T_{\mathrm{line}}|,1),\qquad
C_g=\frac{\min(n,m_g)}{m_g}.
\]
The reward is
\begin{equation}
R_g=\operatorname{clip}_{[0,1]}
\left((Q_g+0.10f_gC_g)f_g^2-0.05P_{\mathrm{dup}}-0.10P_{\mathrm{over}}-0.05P_{\mathrm{invalid}}\right).
\end{equation}

\paragraph{Complementary-reference consistency.}
For complex text, each reference score additionally preserves surface-form accuracy:
$W=0.85V+0.15U$, where $U$ is exact-string F1 after NFKC and case folding, retaining punctuation and requiring IoU at least 0.50. Two primary references and a complementary reference yield
\begin{equation}
W_{\mathrm{text}}=0.82\!\left(0.70\max(W_a,W_b)+0.30\min(W_a,W_b)\right)+0.18W_c.
\end{equation}
When auxiliary geometry is available, $Q_c=0.95W_{\mathrm{text}}+0.05G_{\mathrm{aux}}$; otherwise $Q_c=W_{\mathrm{text}}$. The geometry-only score is $G_{\mathrm{aux}}=0.60\bar{H}_{\mathrm{geo}}+0.30S_{\mathrm{geo}}+0.10C_{\mathrm{geo}}$, excluding transcription agreement. Let $m_c$ be the rounded median reference count and $C_c=\min(n,m_c)/m_c$. Then
\begin{equation}
\begin{split}
R_c=\operatorname{clip}_{[0,1]}\bigl(&(0.90Q_c+0.10f_cC_c)f_c^2
-0.05P_{\mathrm{dup}}-0.10P_{\mathrm{over}}\\
&-0.05P_{\mathrm{invalid}}-0.08P_{\mathrm{large}}\bigr).
\end{split}
\end{equation}

\paragraph{Format gating and penalties.}
The parse grades in \Cref{tab:groundanything_ocr_format} gate the full content score quadratically. Penalties measure duplicated, excessive, or invalid predictions; $P_{\mathrm{large}}$ additionally penalizes boxes covering over 80\% of the image without geometric reference support. The annotation structure selects $R_g$ or $R_c$, which enters GRPO as one scalar reward.

\begin{table}[htbp]
\centering\small
\setlength{\tabcolsep}{5pt}
\renewcommand{\arraystretch}{1.1}
\caption{OCR format grades used for squared reward gating.}
\label{tab:groundanything_ocr_format}
\begin{tabular}{p{0.18\linewidth}p{0.61\linewidth}r}
\hline
Branch & Parse status & Grade \\
\hline
Word/line & Complete, valid structure with no invalid instances & 1.00 \\
& Incomplete structure with reliably parsed instances & 0.60 \\
& Only some valid instances can be recovered & 0.30 \\
& No reliable parse & 0.00 \\
\hline
Complementary & Strictly valid structure with no invalid content & 1.00 \\
references & Complete main structure with extraneous content & 0.45 \\
& Incomplete structure with parseable instances & 0.35 \\
& Recoverable instances without a compliant overall structure & 0.15 \\
& No reliable parse & 0.00 \\
\hline
\end{tabular}
\end{table}

\subsubsection{Pointing}
\label{app:groundanything_point_reward}

Point predictions are matched one-to-one within each label. For distance $d$ on the normalized $0$--$999$ coordinate grid, similarity is
\begin{equation}
s(d)=\begin{cases}\exp[-\tfrac12(d/75)^2],&d\leq200,\\0,&d>200.\end{cases}
\end{equation}
The matched similarities define soft precision $P$ and recall $R$. With $F_2=5PR/(4P+R)$, defined as zero when the denominator vanishes, the reward is
\begin{equation}
R_{\mathrm{point}}=0.55F_2+0.25R_{\mathrm{count}}+0.10F_{1,\mathrm{label}}+0.10R_{\mathrm{fmt}}.
\end{equation}
Count agreement is the smaller-to-larger count ratio when the larger count is positive. Format validity is 1 for a valid response, 0.5 when valid points remain recoverable from a malformed response, and 0 otherwise.

\section{Parallel Inference}
\label{app:groundanything_inference}

\subsection{Direct Diffusion Decoding}
\label{app:groundanything_decoding}

\paragraph{Block state and token alignment.}
Image/query prefill establishes a causal prefix cache and the first anchor token. A physical block contains this known anchor followed by $B-1$ mask positions. Predictions retain the training token shift. Sub-blocks are completed from left to right; every denoising forward still evaluates the full physical block. Completed tokens remain visible to subsequent masked positions, and the prefix cache is reused throughout.

\paragraph{Entropy-guided commitment.}
Following the use of entropy as a reliability signal in diffusion decoding~\citep{WeDLM}, we compute
\begin{equation}
p_j=\operatorname{softmax}(z_j),\qquad
H_j=-\sum_{v\in\mathcal{V}_{\mathrm{gen}}}p_j(v)\log p_j(v),
\end{equation}
over the generatable vocabulary, excluding the mask token. For the still-masked positions $\mathcal{U}$ of the active sub-block, commit all positions in $\mathcal{S}=\{j\in\mathcal{U}:H_j\leq\tau\}$. If $\mathcal{S}$ is empty, commit only $\arg\min_{j\in\mathcal{U}}H_j$. Candidate sampling and reliability assessment are separate operations. The procedure repeats until the sub-block is complete; committed tokens are neither remasked nor revised.

\paragraph{Causal cache construction.}
Bidirectional hidden states depend on positions to their right and cannot serve directly as causal history. After completing a block, one causal forward recomputes its authoritative KV entries and predicts the next anchor. This pass does not verify, reject, or replace the completed block. Thus, a block requiring $D$ denoising passes uses $D+1$ NFEs, excluding initial prefill. The number of passes depends on commitment decisions rather than a fixed per-response denoising count.

\paragraph{Other commitment policies.}
Dynamic Decoding uses a confidence threshold with a forced best-position fallback. Static Decoding commits the highest-confidence remaining positions under a prescribed per-pass quota, distributing the remaining positions across the remaining passes. These policies change token commitment, while preserving blockwise generation. They are distinct from AR verification; their confidence scores should not be identified with the unmodified-logit entropy used above.

\subsection{Optional Self-Speculative Decoding}
\label{app:groundanything_speculative}

Self-speculative decoding~\citep{FastdVLM} uses GroundAnything's shared weights for bidirectional drafting and causal verification, as illustrated in \Cref{fig:self-speculative-decoding}. Verification accepts only the longest consecutive draft prefix agreeing with the causal predictions. The first disagreement ends acceptance; later coincidental matches are discarded. Rejected suffix states are removed from the cache.

\paragraph{Linear schedule.}
Given a committed seed, a bidirectional pass predicts a candidate block, and a causal pass verifies it in parallel. Accepted draft tokens are committed with the verifier correction at the first mismatch. Each round uses two NFEs and $2B$ query tokens. The seed is counted once and is excluded from draft-acceptance statistics.

\paragraph{Quadratic schedule.}
The fused query contains $B$ groups of $B+1$ tokens, with group $i$ arranged as $[d_i,M,\ldots,M]$ at overlapping positions $[t+i,\ldots,t+i+B]$. Row leaders support causal verification; each row's mask positions form a bidirectional proposal group. All queries read the cached prefix. Compare row prediction $a_i$ with $d_{i+1}$, starting the accepted count at one for the known leading token $d_0$. At the first mismatch, select that row's proposal as the next draft. This selection reuses computation but does not verify the selected proposal. After initialization, each round uses one NFE and $B(B+1)$ query tokens. The linear/quadratic terminology concerns query-token counts, not total Transformer FLOPs or a fixed number of accepted tokens.

These schedules use greedy causal comparisons. Their verification semantics concern the converted model's causal branch and do not establish equivalence to sampling an arbitrary autoregressive distribution or to an earlier model's outputs.

\subsection{Execution Optimizations}
\label{app:groundanything_execution}

The implementation progresses from Native PyTorch (Eager) to SGLang (Eager), CUDA Graph replay, and selective FP8 execution. Denoising and causal passes retain their respective attention and cache semantics. FP8 applies to eligible language-model linear operations, with the remaining components kept in BF16. These changes reduce execution overhead or arithmetic cost; quantization can still alter logits and decoding decisions. The corresponding latency results are analyzed in \Cref{app:infrastructure_analysis}.

\section{DLM--MTP Comparison: Experimental Details}
\label{app:dlm_mtp}
\label{app:dlm-mtp-protocol}

\subsection{Controlled Model Comparisons}

The drafters share visual preprocessing, the projector, coordinate vocabulary, serialization, and autoregressive backbone initialization. A single frozen causal branch of GroundAnything verifies all proposals. Adaptation matches ordered examples, supervised-token exposure, optimizer updates, effective batch size, and candidate-length sampling. These controls do not imply equal trainable parameter counts or training FLOPs.

The primary in-house MTP baseline uses a causal recurrent future-token module with weights shared across prediction horizons. It is distinct from LocateAnything, which uses bidirectional attention within box-aligned MTP blocks and autoregressive fallback in Hybrid Mode~\citep{locateanything}. An additional in-house block-MTP control predicts a fully masked candidate block in one bidirectional pass; it shares our data and token format but is not a reproduction of LocateAnything.

The attention ablation compares separately trained causal and bidirectional DLMs with the same architecture, initialization, denoising objective, corruption distribution, training exposure, and refinement schedule. Only candidate-block attention differs during training and inference. All candidate positions can read the committed prefix; the prefix cannot read candidates. This pair isolates attention, whereas comparing block MTP with DLM also changes the objective and refinement procedure.

\subsection{Draft Acceptance and Token Cost}

All configurations receive the same bank of verifier-generated prefixes from held-out COCO prompts, with validation and test prefixes separated by image. The deadline comparison uses long-output prefixes and therefore characterizes decode progress on that subset rather than full-dataset request throughput. Candidate length $K$ counts newly proposed tokens, excluding the committed anchor, padding, and verifier correction or bonus. Greedy verification accepts the longest matching prefix, then emits a correction at the first mismatch or a bonus after complete acceptance, subject to response termination.

For round $r$, let $K_r$, $A_r$, and $C_r$ denote proposed, accepted-draft, and committed-output token counts, and let $T_r$ be full-round latency in milliseconds. We report
\begin{align}
R_{\mathrm{draft}}&=\frac{\sum_r A_r}{\sum_r K_r},
&\bar{A}&=\frac{1}{N_r}\sum_r A_r,
&\bar{C}&=\frac{1}{N_r}\sum_r C_r,\\
c_A&=\frac{\sum_r T_r}{\sum_r A_r},
&c_C&=\frac{\sum_r T_r}{\sum_r C_r},
&\bar{T}&=\frac{1}{N_r}\sum_r T_r,
\end{align}
where $N_r>0$ is the number of rounds; each ratio requires a positive denominator. Ratios of totals retain the cost of zero-acceptance and malformed-proposal rounds. For a full nonterminal round, $C_r=A_r+1$; terminal rounds use actual emitted counts. Output throughput is $1000/c_C$, while $1000/c_A$ measures accepted draft tokens per second.

Round timing begins after the shared prefix cache is ready and ends after output and cache commitment. It includes drafting, verification, rejected work, candidate selection, scheduling, and cache maintenance; shared visual encoding and prefix prefill are outside this decode-only cost. All methods use the same execution conditions.

\paragraph{Acceptance and latency results.}
At $K=16$, one-step DLM accepts 21.76\% of candidates (3.48 tokens per round), compared with 16.81\% (2.69 tokens) for causal MTP.
This reduces $c_A$ from 7.71 to 5.38 ms, a 30.2\% reduction at matched candidate length.
Across the tested lengths, DLM's lowest cost is 5.38 ms at $K=16$, while MTP's is 6.58 ms at $K=8$.
However, MTP is preferable at $K=4$: its cost is 7.15 ms versus 9.31 ms for DLM.
At $K=32$, DLM's accepted prefix decreases to 3.30 tokens despite the longer proposal.
These results favor selecting candidate length by accepted work and elapsed time rather than nominal parallelism.

\subsection{Refinement and Fixed-Deadline Decoding}

The refinement comparison fills all $K$ candidate positions on the first pass. Before pass $j\in\{2,\ldots,S\}$, it remasks the lowest-confidence
\begin{equation}
m_j=\left\lceil\frac{K(S-j+1)}{S}\right\rceil
\end{equation}
positions and predicts them jointly. Both DLM attention variants use this schedule, followed by causal verification. This controlled draft-refinement experiment is distinct from monotone entropy-guided commitment in direct generation.

For two refinement depths $S$ and $S'$ with positive mean latency and accepted-token counts, extra refinement lowers accepted-token cost precisely when
\begin{equation}
\frac{\bar{T}_{S'}}{\bar{T}_S}<\frac{\bar{A}_{S'}}{\bar{A}_S}.
\end{equation}
The condition for committed-token cost replaces $\bar{A}$ by $\bar{C}$. Thus, improved draft acceptance alone does not establish a wall-clock benefit.

For a deadline $D$, candidate length and denoising depth are selected on validation prefixes and fixed for testing. Methods may execute repeated rounds, and only rounds completed by the deadline contribute:
\begin{equation}
\bar{N}(D)=\frac{1}{N}\sum_{i=1}^{N}\sum_r C_{i,r}
\mathbf{1}[t^{\mathrm{finish}}_{i,r}\leq D].
\end{equation}
This measures progress at equal elapsed time, counting verifier corrections and bonuses.

\paragraph{Refinement and deadline results.}
At $K=16$, two DLM passes reduce $c_A$ from 5.38 to 4.70 ms, whereas four and eight passes increase it to 6.84 and 11.83 ms.
Cost per committed output token decreases by only 3.8\%, because the verifier contributes a correction or bonus token in each full round.
At 400 ms, DLM commits 97.1 tokens versus 84.6 for MTP; at 50 ms, MTP leads with 10.6 versus 8.6.
Thus, refinement is beneficial over part of the latency range, but additional passes and very short budgets can reverse the advantage.

\subsection{Structural and Geometric Diagnostics}

Each drafter receives the same diagnostic windows: category transitions from COCO, crowded objects from Dense200, document regions from DocLayNet, and GUI targets from ScreenSpot-Pro. Candidate length is matched, both DLM variants use two refinement steps, and the bidirectional block-MTP control uses one pass. Diagnostic windows are selected from the common reference serialization before inspecting proposals. Raw format error detects illegal structural transitions, token types, or coordinate arity, using the parser state inherited from the prefix. Ending partway through an otherwise valid tuple at the candidate boundary is not itself a format error. Diagnostics precede grammar filtering, fallback, and verification.

The valid-box miss rate is
\begin{equation}
M_{\mathrm{box}}=
\frac{\#\text{ unmatched complete, syntactically valid proposed boxes}}
{\#\text{ complete, syntactically valid proposed boxes}}.
\end{equation}
Matching is one-to-one against relevant remaining references, constrained by category or referring target and IoU at least 0.5. Unmatched duplicates, reversed corners, and zero-area boxes count as geometric misses. The numerator counts unmatched complete, syntactically valid proposed boxes; the denominator counts all complete, syntactically valid proposed boxes. The rate is undefined when this denominator is zero. It measures the fraction of eligible proposals that remain unmatched and is interpreted alongside format errors; it is not a ground-truth miss rate.

Verifier agreement, structural validity, and geometric accuracy measure different properties. Exact greedy verification produces the same sequence as the shared causal verifier; a finite deadline can yield different-length prefixes of that sequence. These diagnostics therefore assess draft reliability and rejected computation, rather than improved final localization under a fixed exact verifier.

\paragraph{Where bidirectional refinement helps.}
Relative to causal MTP, bidirectional DLM reduces COCO format errors from 5.01\% to 1.80\%.
Its valid-box miss rate decreases by 10.10 pp on Dense200 and 7.52 pp on DocLayNet, compared with only 0.46 pp on GUI.
These differences are consistent with greater benefits for proposals containing several spatially related elements.
The retrained DLM pair provides the attention control: bidirectional attention lowers the Dense200 rate from 21.74\% to 14.79\% at the same refinement depth.
Bidirectional block-MTP also improves over causal MTP, so bidirectionality itself is not a uniquely diffusion-based advantage.
Its one-pass comparison with two-step DLM changes objective and refinement cost; it does not isolate diffusion training at equal elapsed time.
The separate deadline analysis provides the matched-time comparison.

\paragraph{Relation to the grounding formulation.}
Together, the acceptance, cost, and deadline results support \hyperref[finding:2.2]{Finding~2.2}: bidirectional diffusion with limited refinement makes draft generation more efficient than the tested causal MTP baseline over the reported operating range.
The attention control supports joint access to visual evidence, while the refinement sweep identifies when an additional denoising pass repays its cost; neither establishes superiority over every MTP design.
The geometric diagnostic measures proposal reliability before verification, separately from ScreenSpot accuracy, F1mIoU, and ground-truth recall.
Exact verification preserves the shared causal model's completed output.

\section{Additional Experimental Analysis}
\label{app:experimental_analysis}

\subsection{Speed Reporting and Decoding Sensitivity}
\label{app:speed_reporting}

\paragraph{Metrics and scope.}
TPF denotes committed output tokens per model forward; TPS denotes output tokens per second.
Forward counts must include the causal cache-construction or verification passes required by the relevant schedule.
The physical block size $B$ in the decoding sweeps includes the known anchor, whereas candidate length $K$ in the DLM--MTP comparison counts newly proposed tokens.
The two quantities should not be identified.
All quality values in the decoding sweeps are COCO F1mIoU, and their speed--accuracy trade-offs should not be extrapolated to a 30-benchmark quality average.

\paragraph{Reference operating points.}
The AR reference is 49.6 TPS with 63.70 F1mIoU.
At $B=32$, throughput is 223.8 TPS for linear self-speculation, 49.9 TPS for quadratic self-speculation, and 127.0 TPS for entropy guidance at $\tau=0.8$.
The corresponding COCO quality is 62.96 for linear self-speculation and 60.72 for entropy guidance.
Thus, the two main speedups are $223.8/49.6=4.51\times$ and $127.0/49.6=2.56\times$, with quality gaps of 0.74 and 2.98 pp.

\subsubsection*{Interpreting the Decoding Sweeps}
\label{app:decoding_sensitivity}

\paragraph{Linear versus quadratic self-speculation.}
Linear TPF rises from 1.41 at $B=4$ to 2.21 at $B=16$, then changes little at $B=32$ (2.18).
Quadratic TPF reaches 3.17 at both $B=16$ and $B=32$.
At $B=32$, its 45.4\% higher TPF nevertheless accompanies much lower TPS than the linear schedule.
Fusing verification and proposal reduces forward-call count but increases the number of query tokens, illustrating why TPF alone is not a latency metric.
From $B=16$ to $B=32$, linear throughput changes from 226.6 to 223.8 TPS, whereas quadratic throughput falls from 88.0 to 49.9 TPS.

\paragraph{Entropy threshold.}
As $\tau$ increases from 0.2 to 1.2, TPF increases from 1.02 to 1.60 and TPS from 107.3 to 148.7.
COCO F1mIoU remains near 60 for $\tau\leq0.8$, peaks at 60.72 at $\tau=0.8$, falls to 54.06 at $\tau=1.0$, and partly recovers to 56.14 at $\tau=1.2$.
The non-monotonic quality curve supports a speed--accuracy trade-off rather than a universal monotonic accuracy law.
Increasing $\tau$ permits more uncertain tokens to be committed together; the observed quality drop is consistent with the risk of premature commitment.

\paragraph{Block size and decoding mode.}
Entropy-guided TPF changes only from 1.18 to 1.24 over the tested block sizes, while its best COCO score is 61.58 at $B=16$.
Linear self-speculation attains its best tested COCO score at $B=32$.
Consequently, the $B=32$ operating point used for the headline comparison is not the maximum-quality choice for every decoding strategy.
Block-size selection therefore depends on task quality and output length; the COCO sweep alone does not establish a universal optimum.

\paragraph{What exact verification preserves.}
Self-speculation verifies against the converted model's causal branch.
Exact greedy verification preserves that branch's outputs, not necessarily the outputs of the separately trained GroundAnything-VLM.
The 0.74 pp gap to GroundAnything-VLM therefore does not imply that the verifier accepts mismatching tokens.
Direct entropy-guided decoding has no AR acceptance test and remains the appropriate main-table setting for assessing diffusion generation itself.

\subsection{Infrastructure Analysis}
\label{app:infrastructure_analysis}

\Cref{fig:decode-infrastructure} adds SGLang (Eager), CUDA Graph, and FP8 sequentially to each decoding mode.
The final time per output token is 6.59 ms for Dynamic Decoding, 6.76 ms for Static Decoding, 7.15 ms for Entropy-Guided Decoding, and 3.73 ms for Self-Speculative Decoding.
Relative to each mode's Native PyTorch (Eager) implementation, the cumulative speedups are $3.48\times$, $3.45\times$, $3.51\times$, and $4.06\times$, respectively.
The consistent reductions support the claim that parallel decoding requires corresponding execution support to realize practical acceleration.

These ratios compare implementations within a decoding mode.
They are not multiplied by the separate $4.51\times$ algorithmic comparison, whose denominator is GroundAnything-VLM at a different reported operating point.
The figure reports latency rather than a matched quality ablation for every infrastructure stage.
In particular, it does not establish that FP8 preserves accuracy; selective quantization can change logits and thus commitment or verification decisions.
The implementation-level distinction is described in \Cref{app:groundanything_execution}.

\subsection{Architecture and Training Ablations}
\label{app:architecture_ablations}

\paragraph{Coordinate representation.}
In \Cref{fig:grounding-model-ablations}, replacing quantized coordinates with textual coordinates reduces the AR score from 73.03 to 71.60 and the DLM score from 61.66 to 55.98.
The respective 1.43 and 5.68 pp drops support the compact coordinate vocabulary as a useful part of both variants, with a larger effect in the diffusion setting.
The figure's $0.25\times$ annotation concerns the textual-representation comparison; it should not be identified with the external SEED1.5-VL token-count ratios.

\paragraph{Visual encoder.}
With MoonViT (Kimi-VL), the AR and DLM scores are 72.81 and 61.03; with Qwen3-ViT, they are 71.12 and 60.12.
MoonViT-V2 is consistently strongest among these tested choices.
These comparisons support the selected encoder within this training recipe; they do not establish a universal ranking across vision architectures.
\Cref{app:deepstack_discussion} analyzes the task-dependent value of multi-level visual injection at the coordinate-token readout; this is a separate design question from the encoder comparison.

\paragraph{Training-phase ablation.}
The ablation labeled ``Stage~III'' in \Cref{fig:grounding-model-ablations} gives 71.74 for GroundAnything-VLM and 45.04 for GroundAnything, drops of 1.29 and 16.62 pp.
We retain the figure's ablation label here; it is distinct from the Stage~1--4 pretraining/alignment numbering above.
The larger DLM sensitivity indicates that the ablated training phase contributes substantially to useful grounding behavior after conversion.
The experiment changes a complete phase and does not isolate individual reward components or optimization choices within it.

\paragraph{Comparison scope.}
The full ablation scores (73.03/61.66) differ from the final 30-benchmark summary (72.42/61.75).
We retain the ablation values and interpret only within-comparison changes, without replacing either set of scores or assuming identical aggregation.

\subsubsection{Multi-level Visual Injection and Coordinate Readout}
\label{app:deepstack_discussion}

\paragraph{Connection to the encoder choice.}
GroundAnything uses MoonViT-V2 with an input-level projector and no intermediate visual injection.
DeepStack supplies additional visual evidence at intermediate language layers~\citep{DeepStack}; Qwen3-VL adds three projected ViT feature levels to its first three language layers and reports fine-grained understanding gains~\citep{qwen3vl}.
The relevant question here is whether grounding adaptation makes this evidence useful to the coordinate-token readout under both causal and masked contexts.
Our encoder ablation supports MoonViT-V2 under the current recipe but does not isolate the injection scheme.

\paragraph{Readout sensitivity.}
Fix model parameters, image, query, and one decoding context: an AR prefix or a DLM block with fixed masks and visible tokens.
For vectorized hidden states, write $h_{\ell+1}=F_\ell(h_\ell)+S_\ell u_\ell$ and $q=g(h_L)$, where $u_\ell$ contains projected visual features, $S_\ell$ inserts them at visual-token positions, and $q\in\mathbb R^V$ contains logits over a fixed vocabulary at one coordinate position.
The readout $g$ includes final normalization and the output head; $\mathcal I$ indexes injection layers.
With $h_0$ fixed, changing injected features by $\delta u_\ell$ gives
\begin{equation}
\begin{aligned}
d&=\sum_{\ell\in\mathcal I}B_\ell\delta u_\ell
   +O(\lVert\delta u\rVert_2^2),\\
B_\ell&=Dg(h_L)J_{L-1}\cdots J_{\ell+1}S_\ell,
\qquad J_j=DF_j(h_j).
\end{aligned}
\label{eq:deepstack_sensitivity}
\end{equation}
Here $d$ is the exact logit change, $\delta u$ concatenates the feature changes, and empty products are identities; all derivatives are evaluated at the unchanged interface.
The chain rule gives each path contribution; locally bounded second derivatives give the remainder.
This is sensitivity relative to the specified interface, not prediction error relative to an ideal alignment.

\paragraph{Task alignment.}
Let $y$ denote the reference coordinate token and $\mathcal L_y(q)=-\log p_y$ its cross-entropy loss.
Write $p=\operatorname{softmax}(q)$ and $p'=\operatorname{softmax}(q+d)$ for the original and updated token distributions.
For the exact change $d$,
\begin{equation}
\begin{aligned}
\mathcal L_y(q+d)-\mathcal L_y(q)
 &=\log\!\left(\sum_{v=1}^{V}p_v e^{d_v}\right)-d_y\\
 &=(p-e_y)^\top d+D_{\mathrm{KL}}(p\Vert p'),
\end{aligned}
\label{eq:deepstack_task_loss}
\end{equation}
where $e_y$ is the one-hot target and $D_{\mathrm{KL}}(p\Vert p')=\sum_v p_v\log(p_v/p'_v)$.
The first equality follows from log-softmax; expanding the KL divergence gives the second.
Additional evidence lowers coordinate loss precisely when its contribution opposing the current loss gradient exceeds the nonnegative KL remainder.
Averaging over matched reference coordinate positions and decoding contexts gives the corresponding token-loss comparison.
The loss identity is exact; substituting the linearized change from \Cref{eq:deepstack_sensitivity} requires controlling its remainder.

If additional branches can be zeroed with all other components unchanged, the multi-injection model retains the single-interface model as a special case; path count alone cannot raise optimal task loss.
The testable issue is whether finite grounding data and optimization learn useful contributions across causal and masked contexts.
A controlled comparison should fix encoder and language-backbone initialization, coordinate vocabulary, data, and training budget, then compare input-only and multi-level interfaces with comparable capacity.
Coordinate loss across adaptation budgets and dense/tiny-object localization would test this hypothesis; entropy-based commitment and final box metrics require separate evaluation beyond the fixed-context analysis.
Disabling branches after training measures sensitivity only.
Thus, the current ablation supports our encoder choice, while attributing its advantage to the absence of DeepStack requires an injection-specific experiment.

\subsection{Output Token Efficiency and Generation Cost}
\label{app:output_efficiency}

\paragraph{Compact serialization.}
\Cref{tab:grounding-token-efficiency} follows the output-length analysis of \citet{rexomni}.
Its SEED1.5-VL row is taken from that work.
The GroundAnything variants use the same structured vocabulary and have identical reported output statistics: 7.6 tokens/box on COCO and 5.1 on Dense200.
Compared with 148.8 and 74.5 tokens/box for SEED1.5-VL, these are approximately $19.6\times$ and $14.6\times$ lower token counts per box.
The ratios reflect serialization length, not matched-hardware latency; differences in detected instance counts and model execution preclude reading them as speedups.

\paragraph{Generation cost across object counts.}
\Cref{fig:efficiency-report} groups outputs by predicted box count and reports average generation time alongside output-token count.
Longer coordinate lists increase decoding work for both variants, while GroundAnything reduces generation time across the displayed ranges, with larger absolute savings for longer outputs.
The token-efficiency table characterizes serialization length; this figure characterizes generation cost as the number of predicted instances grows.
The speed--quality comparisons in \Cref{app:speed_reporting} complement this output-length analysis.

\section{Comprehensive Grounding Benchmark Results}
\label{app:grounding_results}
\label{app:detailed_results}

We expand the selected main-text comparisons with complete baseline tables, threshold-specific localization metrics, and task-level analysis. GroundAnything uses Entropy-Guided Decoding throughout these tables; GroundAnything-VLM uses AR decoding. External results and unresolved evaluations retain their original qualifications.

\subsection*{Benchmark Suite and Metrics}
\label{app:benchmark_suite}

\paragraph{Tasks and datasets.}
The suite covers common and long-tailed detection on COCO~\citep{COCO} and LVIS~\citep{LVIS}; dense and tiny-object detection on Dense200~\citep{rexomni} and VisDrone~\citep{VisDrone}; referring grounding on RefCOCO, RefCOCO+, and RefCOCOg~\citep{RefCOCO,RefCOCOg,RefCOCOgUMD}; and object pointing on COCO, LVIS, Dense200, VisDrone, and RefCOCOg val/test.
Spatial pointing uses RefSpatial~\citep{RoboRefer2B} and RoboSpatial~\citep{RoboSpatial}, while GUI grounding uses ScreenSpot-V2~\citep{ScreenSpotV2}, ScreenSpot-Pro~\citep{ScreenSpotPro}, and OSWorld-G~\citep{JEDI3B}.
OCR covers HierText~\citep{HierText}, ICDAR2015~\citep{ICDAR2015}, TotalText~\citep{TotalText}, and SROIE~\citep{SROIE}; layout grounding uses DocLayNet~\citep{DocLayNet} and M6Doc~\citep{M6Doc}; visual prompting uses FSC147~\citep{FSC147} and Dense200.

\paragraph{Metrics and aggregation.}
Following the grounding evaluation convention of \citet{rexomni}, box tasks report recall, precision, and F1 at IoU 0.50 and 0.95, together with their reported aggregates over IoU thresholds from 0.50 to 0.95.
F1mIoU denotes the threshold-aggregated F1 score, not mean matched-box IoU.
OCR additionally requires transcription agreement under the loose-match evaluation protocol and reports parse-error rates separately.
Object pointing uses point-in-mask F1; spatial pointing uses point-in-mask accuracy, ScreenSpot uses action accuracy, and OSWorld-G uses exact accuracy.
RefCOCO avg is the unweighted mean of the three family entries; RefSpatial avg averages Location and Placement.
Reported benchmark entries are preserved, and missing metrics are not reconstructed from other scores.

\paragraph{Reporting conventions and external scores.}
The notation in \Cref{sec:benchmark_reporting} applies throughout. R and P denote recall and precision; lower is better only for parse error.
Model-name stars identify external-only rows, while entry-level stars identify individual substitutions or task-interface exceptions.
The relevant captions specify their sources. In particular, selected detector, SEED1.5-VL, Molmo, spatial, and GUI references are taken from \citet{rexomni}; GUI-Owl references follow \citet{locateanything}.
Size groups refer to language-backbone configurations; the 7B understanding branch is used for MoT models and total language parameters for MoE models. Qwen3.7-Max, Kimi-K2.6, Kimi-K3, and GPT-6 Astra are grouped separately in the $>1$T category.

\paragraph{Interpreting unsupported or uncertain evaluations.}
N/A means that the required task interface or a reliable prompt/parser combination was unavailable; it does not establish that a model intrinsically lacks the capability.
The \textsuperscript{\textdagger} marker identifies uncertain prompt/protocol alignment, including the affected Kimi, DeepSeek, and SenseNova-Vision evaluations.
Repeated DeepSeek GUI attempts did not yield a suitable prompt/parser combination.
Starred BAGEL and MiMo observations under suspected protocol incompatibility are retained descriptively.
Where external scores replace unreproduced local results, only the matching model and available metrics are used: DeepSeek-VL2-Small scores are never substituted for the full model, and missing parse-error rates remain unreported.
The following comparisons use the reported compatible entries.

\subsection{Common and Long-tailed Object Detection}
\label{app:common_longtailed_detection}

\paragraph{COCO.}
GroundAnything-VLM achieves 63.70 F1mIoU, while direct diffusion retains 60.72, exceeding Rex-Omni by 4.44 pp.
The threshold breakdown qualifies this result: GroundAnything's F1 at IoU 0.95 is 22.39, below LocateAnything Hybrid's 27.61, despite its higher aggregate.
Thus, the overall gain does not imply uniformly tighter boxes at the strictest threshold.
\begin{table}[htbp]
  \centering
  \BenchmarkTableFont
  \caption{\textbf{COCO.} Complete box-grounding metrics.  Model-name stars denote external scores from Table~2 of \citet{rexomni}.}
  \label{tab:app-coco}
  \begingroup
  \fontsize{8}{9.7}\selectfont
  \setlength{\tabcolsep}{3pt}
  \renewcommand{\arraystretch}{1.15}
  \resizebox{\linewidth}{!}{%
  \begin{NiceTabular}{@{}>{\raggedright\arraybackslash}p{177pt}cccccccccc@{}}
  \CodeBefore
    \rowcolor{benchmarktype}{3}
    \rowcolor{benchmarktype}{7}
    \rowcolor{benchmarktype}{9}
    \rowcolor{benchmarkpurple}{26}
    \rowcolor{benchmarkgreen}{27}
    \rowcolor{benchmarktype}{28}
    \rowcolor{benchmarktype}{36}
  \Body
  \toprule
  \textbf{Model} & \BenchHead{Zero-shot} & \multicolumn{3}{c}{\textbf{IoU 0.50}} & \multicolumn{3}{c}{\textbf{IoU 0.95}} & \multicolumn{3}{c}{\textbf{mIoU}} \\
  \cmidrule(lr){3-5}\cmidrule(lr){6-8}\cmidrule(lr){9-11}
   &  & \BenchHead{R} & \BenchHead{P} & \BenchHead{F1} & \BenchHead{R} & \BenchHead{P} & \BenchHead{F1} & \BenchHead{R} & \BenchHead{P} & \BenchHead{F1} \\
  \midrule
  \multicolumn{11}{@{}l}{\hspace{0.4em}\strut\textbf{Closed-set Specialized Detectors}} \\
  DINO-R50\textsuperscript{*}\hspace{0.35em}\citep{DINOR50} & NO & 62.60 & 76.50 & 68.80 & 17.80 & 25.80 & 21.10 & 50.00 & 62.40 & 55.60 \\
  DETR-R50\textsuperscript{*}\hspace{0.35em}\citep{DETRR50} & NO & 59.60 & 73.90 & 65.90 & 10.60 & 19.00 & 13.60 & 42.90 & 55.30 & 48.30 \\
  DyHead-R50\textsuperscript{*}\hspace{0.35em}\citep{DyHeadR50} & NO & 58.10 & 76.60 & 66.10 & 11.90 & 20.60 & 15.00 & 44.80 & 60.10 & 51.30 \\
  \addlinespace[2pt]
  \multicolumn{11}{@{}l}{\hspace{0.4em}\strut\textbf{Open-set Specialized Detectors}} \\
  GroundingDINO\hspace{0.35em}\citep{groundingdino} & YES & 79.60 & 83.23 & 81.37 & 24.66 & 28.07 & 26.25 & 61.13 & 63.65 & 60.56 \\
  \addlinespace[2pt]
  \multicolumn{11}{@{}l}{\hspace{0.4em}\strut\textbf{Vision-Language Models (<10B)}} \\
  Rex-Omni\hspace{0.35em}\citep{rexomni} & YES & 69.90 & 80.99 & 75.04 & 17.93 & 21.73 & 19.65 & 53.92 & 59.36 & 56.28 \\
  LocateAnything Fast\hspace{0.35em}\citep{locateanything} & YES & 66.15 & 73.34 & 69.56 & 24.73 & 26.90 & 25.77 & 50.89 & 59.26 & 53.06 \\
  LocateAnything Hybrid\hspace{0.35em}\citep{locateanything} & YES & 74.67 & 80.80 & 77.61 & 26.64 & 28.66 & 27.61 & 53.43 & 66.19 & 59.12 \\
  LocateAnything Slow NTP\hspace{0.35em}\citep{locateanything} & YES & 75.02 & 81.87 & 78.30 & 24.97 & 27.19 & 26.03 & 54.70 & 65.09 & 59.37 \\
  Qwen2.5-VL-7B\hspace{0.35em}\citep{qwen25vl} & UNK & 57.18 & 62.87 & 59.89 & 12.16 & 13.23 & 12.67 & 44.67 & 49.05 & 46.27 \\
  Qwen3-VL-2B\hspace{0.35em}\citep{qwen3vl} & UNK & 63.50 & 67.14 & 65.27 & 21.77 & 22.57 & 22.16 & 42.64 & 44.86 & 42.48 \\
  Qwen3-VL-4B\hspace{0.35em}\citep{qwen3vl} & UNK & 66.05 & 70.64 & 68.27 & 23.69 & 24.87 & 24.27 & 45.13 & 48.02 & 46.53 \\
  Qwen3-VL-8B\hspace{0.35em}\citep{qwen3vl} & UNK & 65.70 & 69.48 & 67.54 & 23.91 & 25.11 & 24.49 & 45.09 & 47.58 & 46.30 \\
  Qwen3.5-4B\hspace{0.35em}\citep{qwen35} & UNK & 65.86 & 67.75 & 66.79 & 22.94 & 23.67 & 23.30 & 48.97 & 46.28 & 47.58 \\
  Qwen3.5-9B\hspace{0.35em}\citep{qwen35} & UNK & 73.17 & 68.13 & 70.56 & 23.83 & 24.48 & 24.15 & 58.80 & 46.60 & 51.99 \\
  RynnBrain1.1\hspace{0.35em}\citep{RynnBrain112B} & UNK & 36.49 & 60.51 & 45.53 & 11.56 & 16.53 & 13.60 & 28.72 & 45.31 & 35.15 \\
  SenseNova-Vision\hspace{0.35em}\citep{SenseNovaVision7BMoT} & UNK & 79.07 & 80.54 & 79.80 & \textbf{27.19} & \textbf{30.19} & \textbf{28.61} & 63.13 & 55.37 & 57.49 \\
  MiMo-VL-7B-SFT\hspace{0.35em}\citep{MimoVL} & UNK & 65.48 & 72.18 & 68.67 & 10.02 & 11.11 & 10.54 & 45.72 & 50.47 & 47.98 \\
  MiMo-VL-7B-RL\hspace{0.35em}\citep{MimoVL} & UNK & 65.64 & 74.52 & 69.79 & 9.17 & 10.29 & 9.69 & 44.48 & 50.24 & 47.18 \\
  BAGEL\hspace{0.35em}\citep{BAGEL7BMoT} & UNK & 65.57 & 75.20 & 70.06 & 11.04 & 13.01 & 11.94 & 46.30 & 47.15 & 45.98 \\
  RynnBrain\hspace{0.35em}\citep{RynnBrain2B} & UNK & 21.27 & 49.89 & 29.82 & 4.89 & 10.87 & 6.74 & 15.46 & 35.37 & 21.51 \\
  GroundAnything-VLM & YES & 81.53 & 81.43 & \textbf{81.48} & 23.91 & 23.83 & 23.87 & 63.75 & 63.64 & \textbf{63.70} \\
  GroundAnything & YES & 81.02 & 70.38 & 75.33 & 26.38 & 19.45 & 22.39 & \textbf{65.44} & 56.65 & 60.72 \\
  \addlinespace[2pt]
  \multicolumn{11}{@{}l}{\hspace{0.4em}\strut\textbf{Vision-Language Models (10B--1T)}} \\
  Qwen3.6-27B\hspace{0.35em}\citep{qwen3627b} & UNK & 78.19 & 79.00 & 78.59 & 25.02 & 25.68 & 25.35 & 61.26 & 62.19 & 61.72 \\
  Qwen3-VL-32B\hspace{0.35em}\citep{qwen3vl} & UNK & 76.47 & 77.79 & 77.12 & 23.05 & 23.84 & 23.44 & 59.50 & 60.84 & 60.16 \\
  Qwen3.8-27B\hspace{0.35em}\citep{qwen38} & UNK & 77.48 & 79.51 & 78.48 & 23.60 & 24.44 & 24.01 & 59.97 & 61.72 & 60.84 \\
  Qwen3.5-35B-A3B\hspace{0.35em}\citep{qwen35} & UNK & 78.37 & 79.65 & 79.00 & 25.20 & 25.92 & 25.56 & 61.45 & 62.71 & 62.07 \\
  DeepSeek-VL2-Small-16B\hspace{0.35em}\citep{Deepseekvl2} & UNK & 20.86 & 45.22 & 28.55 & 0.85 & 1.73 & 1.14 & 10.12 & 21.19 & 13.70 \\
  DeepSeek-VL2-27B\hspace{0.35em}\citep{Deepseekvl2} & UNK & 34.58 & \textbf{84.65} & 49.10 & 12.89 & 28.12 & 17.67 & 28.60 & \textbf{67.97} & 40.25 \\
  SEED1.5-VL\textsuperscript{*}\hspace{0.35em}\citep{SEED15VL} & YES & 65.30 & 78.60 & 71.30 & 12.70 & 16.40 & 14.30 & 46.80 & 56.90 & 51.40 \\
  \addlinespace[2pt]
  \multicolumn{11}{@{}l}{\hspace{0.4em}\strut\textbf{Vision-Language Models (>1T)}} \\
  Qwen3.7-Max\hspace{0.35em}\citep{qwen37} & UNK & 78.98 & 82.62 & 80.76 & 24.19 & 25.15 & 24.66 & 60.97 & 64.72 & 62.79 \\
  Kimi-K2.6 & UNK & 72.27 & 80.10 & 75.99 & 25.04 & 27.19 & 26.07 & 58.28 & 64.34 & 61.16 \\
  Kimi-K3\hspace{0.35em}\citep{KimiK3} & UNK & 72.69 & 84.33 & 78.08 & 23.40 & 26.13 & 24.69 & 56.99 & 65.37 & 60.89 \\
  GPT-6 Astra & UNK & \textbf{82.65} & 77.22 & 79.81 & 26.50 & 25.74 & 26.11 & 64.61 & 61.03 & 62.75 \\
  \addlinespace[2pt]
  \bottomrule
  
  \end{NiceTabular}%
  }
  \endgroup
\end{table}

\paragraph{LVIS.}
GroundAnything reaches 52.59 F1mIoU, above Rex-Omni (46.74) and LocateAnything Fast (42.90), while GroundAnything-VLM reaches 56.63.
This extends competitive grounding beyond common categories, although SenseNova-Vision remains stronger than the DLM variant at 56.12.
The AR variant's 75.46 F1 at IoU 0.50 versus 23.23 at IoU 0.95 also shows that rare-category coverage and very tight localization remain distinct challenges.
\begin{table}[htbp]
  \centering
  \BenchmarkTableFont
  \caption{\textbf{LVIS.} Complete box-grounding metrics.  The starred SEED1.5-VL scores are from Table~3 of \citet{rexomni}.}
  \label{tab:app-lvis}
  \begingroup
  \fontsize{8}{9.7}\selectfont
  \setlength{\tabcolsep}{3pt}
  \renewcommand{\arraystretch}{1.15}
  \resizebox{\linewidth}{!}{%
  \begin{NiceTabular}{@{}>{\raggedright\arraybackslash}p{177pt}cccccccccc@{}}
  \CodeBefore
    \rowcolor{benchmarktype}{3}
    \rowcolor{benchmarktype}{5}
    \rowcolor{benchmarkpurple}{22}
    \rowcolor{benchmarkgreen}{23}
    \rowcolor{benchmarktype}{24}
    \rowcolor{benchmarktype}{32}
  \Body
  \toprule
  \textbf{Model} & \BenchHead{Zero-shot} & \multicolumn{3}{c}{\textbf{IoU 0.50}} & \multicolumn{3}{c}{\textbf{IoU 0.95}} & \multicolumn{3}{c}{\textbf{mIoU}} \\
  \cmidrule(lr){3-5}\cmidrule(lr){6-8}\cmidrule(lr){9-11}
   &  & \BenchHead{R} & \BenchHead{P} & \BenchHead{F1} & \BenchHead{R} & \BenchHead{P} & \BenchHead{F1} & \BenchHead{R} & \BenchHead{P} & \BenchHead{F1} \\
  \midrule
  \multicolumn{11}{@{}l}{\hspace{0.4em}\strut\textbf{Open-set Specialized Detectors}} \\
  GroundingDINO\hspace{0.35em}\citep{groundingdino} & YES & 52.64 & \textbf{82.61} & 64.30 & 23.34 & 26.55 & 24.84 & 46.06 & \textbf{61.34} & 52.61 \\
  \addlinespace[2pt]
  \multicolumn{11}{@{}l}{\hspace{0.4em}\strut\textbf{Vision-Language Models (<10B)}} \\
  Rex-Omni\hspace{0.35em}\citep{rexomni} & YES & 58.50 & 73.11 & 64.99 & 17.76 & 20.04 & 18.83 & 45.50 & 48.06 & 46.74 \\
  LocateAnything Fast\hspace{0.35em}\citep{locateanything} & YES & 48.06 & 61.58 & 53.98 & 20.37 & 25.12 & 22.50 & 38.43 & 48.55 & 42.90 \\
  LocateAnything Hybrid\hspace{0.35em}\citep{locateanything} & YES & 55.98 & 71.98 & 62.98 & 22.43 & 27.72 & 24.79 & 44.30 & 56.25 & 49.56 \\
  LocateAnything Slow NTP\hspace{0.35em}\citep{locateanything} & YES & 57.98 & 75.03 & 65.41 & 20.96 & 26.03 & 23.22 & 44.95 & 57.46 & 50.44 \\
  Qwen2.5-VL-7B\hspace{0.35em}\citep{qwen25vl} & UNK & 49.52 & 66.52 & 56.78 & 8.16 & 9.87 & 8.93 & 33.37 & 43.60 & 37.80 \\
  Qwen3-VL-2B\hspace{0.35em}\citep{qwen3vl} & UNK & 55.20 & 68.93 & 61.30 & 16.50 & 19.12 & 17.71 & 40.66 & 49.71 & 44.73 \\
  Qwen3-VL-4B\hspace{0.35em}\citep{qwen3vl} & UNK & 59.60 & 76.16 & 66.87 & 18.71 & 22.25 & 20.32 & 44.84 & 56.15 & 49.86 \\
  Qwen3-VL-8B\hspace{0.35em}\citep{qwen3vl} & UNK & 59.47 & 74.39 & 66.10 & 18.00 & 21.19 & 19.46 & 44.25 & 54.49 & 48.84 \\
  Qwen3.5-4B\hspace{0.35em}\citep{qwen35} & UNK & 58.19 & 70.41 & 63.72 & 17.87 & 20.52 & 19.10 & 42.65 & 50.95 & 46.43 \\
  Qwen3.5-9B\hspace{0.35em}\citep{qwen35} & UNK & 60.86 & 72.03 & 65.98 & 19.33 & 22.00 & 20.58 & 45.28 & 53.07 & 48.87 \\
  RynnBrain1.1\hspace{0.35em}\citep{RynnBrain112B} & UNK & 26.86 & 52.27 & 35.48 & 8.76 & 13.56 & 10.64 & 20.18 & 36.64 & 26.01 \\
  SenseNova-Vision\hspace{0.35em}\citep{SenseNovaVision7BMoT} & UNK & 64.50 & 75.97 & 69.76 & \textbf{27.09} & \textbf{30.94} & \textbf{28.88} & 52.08 & 60.84 & 56.12 \\
  MiMo-VL-7B-SFT\hspace{0.35em}\citep{MimoVL} & UNK & 42.88 & 60.78 & 50.28 & 6.26 & 8.50 & 7.21 & 28.39 & 39.89 & 33.17 \\
  MiMo-VL-7B-RL\hspace{0.35em}\citep{MimoVL} & UNK & 43.50 & 61.87 & 51.09 & 5.55 & 7.52 & 6.38 & 27.53 & 38.74 & 32.18 \\
  BAGEL\hspace{0.35em}\citep{BAGEL7BMoT} & UNK & 34.68 & 51.47 & 41.44 & 8.74 & 11.58 & 9.96 & 24.88 & 35.76 & 29.34 \\
  RynnBrain\hspace{0.35em}\citep{RynnBrain2B} & UNK & 15.06 & 43.29 & 22.35 & 3.24 & 8.90 & 4.75 & 10.34 & 28.98 & 15.25 \\
  GroundAnything-VLM & YES & \textbf{72.91} & 78.18 & \textbf{75.46} & 22.75 & 23.72 & 23.23 & \textbf{54.94} & 58.42 & \textbf{56.63} \\
  GroundAnything & YES & 69.98 & 67.68 & 68.81 & 27.01 & 19.78 & 22.84 & 53.50 & 54.73 & 52.59 \\
  \addlinespace[2pt]
  \multicolumn{11}{@{}l}{\hspace{0.4em}\strut\textbf{Vision-Language Models (10B--1T)}} \\
  Qwen3.6-27B\hspace{0.35em}\citep{qwen3627b} & UNK & 62.89 & 75.25 & 68.52 & 19.87 & 22.54 & 21.12 & 47.15 & 55.67 & 51.05 \\
  Qwen3-VL-32B\hspace{0.35em}\citep{qwen3vl} & UNK & 61.21 & 72.44 & 66.36 & 18.01 & 20.44 & 19.15 & 45.41 & 53.20 & 48.99 \\
  Qwen3.8-27B\hspace{0.35em}\citep{qwen38} & UNK & 62.86 & 74.17 & 68.05 & 18.79 & 21.14 & 19.90 & 46.43 & 54.15 & 49.99 \\
  Qwen3.5-35B-A3B\hspace{0.35em}\citep{qwen35} & UNK & 62.87 & 74.82 & 68.33 & 20.04 & 22.75 & 21.31 & 47.14 & 55.41 & 50.94 \\
  DeepSeek-VL2-Small-16B\hspace{0.35em}\citep{Deepseekvl2} & UNK & 12.80 & 28.41 & 17.65 & 0.50 & 1.20 & 0.71 & 6.10 & 13.05 & 8.31 \\
  DeepSeek-VL2-27B\hspace{0.35em}\citep{Deepseekvl2} & UNK & 26.13 & 70.96 & 38.19 & 10.40 & 24.05 & 14.52 & 21.06 & 54.62 & 30.39 \\
  SEED1.5-VL\textsuperscript{*}\hspace{0.35em}\citep{SEED15VL} & YES & 54.70 & 82.00 & 65.60 & 15.00 & 28.10 & 19.50 & 38.50 & 59.30 & 46.70 \\
  \addlinespace[2pt]
  \multicolumn{11}{@{}l}{\hspace{0.4em}\strut\textbf{Vision-Language Models (>1T)}} \\
  Qwen3.7-Max\hspace{0.35em}\citep{qwen37} & UNK & 63.86 & 79.06 & 70.65 & 20.27 & 23.49 & 21.76 & 48.54 & 59.04 & 53.28 \\
  Kimi-K2.6 & UNK & 57.79 & 76.44 & 65.82 & 21.58 & 26.91 & 23.95 & 45.23 & 58.95 & 51.19 \\
  Kimi-K3\hspace{0.35em}\citep{KimiK3} & UNK & 57.17 & 76.66 & 65.50 & 17.46 & 21.54 & 19.29 & 42.08 & 55.14 & 47.73 \\
  GPT-6 Astra & UNK & 71.35 & 73.85 & 72.58 & 23.47 & 24.60 & 24.02 & 54.23 & 55.73 & 54.97 \\
  \addlinespace[2pt]
  \bottomrule
  
  \end{NiceTabular}%
  }
  \endgroup
\end{table}

\subsection{Dense and Tiny Object Detection}
\label{app:dense_tiny_detection}

\paragraph{Dense200.}
GroundAnything-VLM and GroundAnything reach 75.55 and 70.04 F1mIoU, respectively, both above GPT-6 Astra (65.04), Rex-Omni (53.29), and LocateAnything Hybrid (50.07).
GroundAnything also scores 26.86 at IoU 0.95 versus 12.64 for GPT-6 Astra, so its advantage is not restricted to coarse overlap.
These results directly support precise visual evidence extraction when many neighboring instances must be represented in one response.
\begin{table}[htbp]
  \centering
  \BenchmarkTableFont
  \caption{\textbf{Dense200.} Complete box-grounding metrics. Local BAGEL and DeepSeek runs did not reproduce the reference results, so only available external scores are reported for these models. The starred BAGEL scores are taken from Table~1 of \citet{SenseNovaVision7BMoT}. No matching external result is available for full DeepSeek-VL2; Small and Tiny checkpoint results are not substituted. The starred DeepSeek-VL2-Small and SEED1.5-VL scores are from Table~4 of \citet{rexomni}, also reported in Table~2 of \citet{locateanything}.}
  \label{tab:app-dense200}
  \begingroup
  \fontsize{8}{9.7}\selectfont
  \setlength{\tabcolsep}{3pt}
  \renewcommand{\arraystretch}{1.15}
  \resizebox{\linewidth}{!}{%
  \begin{NiceTabular}{@{}>{\raggedright\arraybackslash}p{177pt}ccccccccc@{}}
  \CodeBefore
    \rowcolor{benchmarktype}{3}
    \rowcolor{benchmarktype}{5}
    \rowcolor{benchmarkpurple}{22}
    \rowcolor{benchmarkgreen}{23}
    \rowcolor{benchmarktype}{24}
    \rowcolor{benchmarktype}{32}
  \Body
  \toprule
  \textbf{Model} & \multicolumn{3}{c}{\textbf{IoU 0.50}} & \multicolumn{3}{c}{\textbf{IoU 0.95}} & \multicolumn{3}{c}{\textbf{mIoU}} \\
  \cmidrule(lr){2-4}\cmidrule(lr){5-7}\cmidrule(lr){8-10}
   & \BenchHead{R} & \BenchHead{P} & \BenchHead{F1} & \BenchHead{R} & \BenchHead{P} & \BenchHead{F1} & \BenchHead{R} & \BenchHead{P} & \BenchHead{F1} \\
  \midrule
  \multicolumn{10}{@{}l}{\hspace{0.4em}\strut\textbf{Open-set Specialized Detectors}} \\
  GroundingDINO\hspace{0.35em}\citep{groundingdino} & 22.43 & 36.30 & 27.73 & 10.92 & 18.38 & 13.70 & 20.16 & 32.62 & 24.92 \\
  \addlinespace[2pt]
  \multicolumn{10}{@{}l}{\hspace{0.4em}\strut\textbf{Vision-Language Models (<10B)}} \\
  Rex-Omni\hspace{0.35em}\citep{rexomni} & 70.61 & 75.13 & 72.80 & 8.66 & 9.17 & 8.91 & 51.79 & 54.88 & 53.29 \\
  LocateAnything Fast\hspace{0.35em}\citep{locateanything} & 22.45 & 26.12 & 24.15 & 9.16 & 10.50 & 9.78 & 19.30 & 22.31 & 20.70 \\
  LocateAnything Hybrid\hspace{0.35em}\citep{locateanything} & 58.06 & 63.32 & 60.57 & 21.52 & 22.90 & 22.19 & 48.13 & 52.18 & 50.07 \\
  LocateAnything Slow NTP\hspace{0.35em}\citep{locateanything} & 74.89 & 81.46 & 78.03 & 20.64 & 22.11 & 21.35 & 59.65 & 64.62 & 62.04 \\
  Qwen2.5-VL-7B\hspace{0.35em}\citep{qwen25vl} & 0.50 & 0.57 & 0.53 & 0.00 & 0.00 & 0.00 & 0.22 & 0.24 & 0.23 \\
  Qwen3-VL-2B\hspace{0.35em}\citep{qwen3vl} & 9.93 & 13.28 & 11.36 & 1.27 & 1.93 & 1.53 & 7.18 & 9.65 & 8.23 \\
  Qwen3-VL-4B\hspace{0.35em}\citep{qwen3vl} & 17.61 & 22.94 & 19.92 & 2.68 & 3.76 & 3.13 & 12.33 & 16.23 & 14.02 \\
  Qwen3-VL-8B\hspace{0.35em}\citep{qwen3vl} & 15.42 & 16.40 & 15.90 & 2.26 & 2.58 & 2.41 & 11.13 & 11.77 & 11.44 \\
  Qwen3.5-4B\hspace{0.35em}\citep{qwen35} & 43.44 & 48.47 & 45.82 & 8.20 & 9.01 & 8.58 & 32.20 & 36.12 & 34.04 \\
  Qwen3.5-9B\hspace{0.35em}\citep{qwen35} & 39.02 & 42.00 & 40.46 & 7.37 & 8.19 & 7.76 & 28.82 & 31.33 & 30.02 \\
  RynnBrain1.1\hspace{0.35em}\citep{RynnBrain112B} & 0.10 & 4.00 & 0.19 & 0.02 & 0.50 & 0.04 & 0.06 & 2.40 & 0.12 \\
  SenseNova-Vision\hspace{0.35em}\citep{SenseNovaVision7BMoT} & 79.18 & 86.84 & 82.83 & 24.52 & 25.72 & 25.10 & 65.31 & 71.20 & 68.13 \\
  MiMo-VL-7B-SFT\hspace{0.35em}\citep{MimoVL} & 11.75 & 12.38 & 12.06 & 0.14 & 0.14 & 0.14 & 5.62 & 5.95 & 5.78 \\
  MiMo-VL-7B-RL\hspace{0.35em}\citep{MimoVL} & 13.10 & 14.14 & 13.60 & 0.08 & 0.07 & 0.07 & 6.31 & 6.88 & 6.58 \\
  BAGEL\textsuperscript{*}\hspace{0.35em}\citep{BAGEL7BMoT} & -- & -- & -- & -- & -- & -- & -- & -- & 42.40 \\
  RynnBrain\hspace{0.35em}\citep{RynnBrain2B} & 0.03 & 1.50 & 0.06 & 0.00 & 0.00 & 0.00 & 0.02 & 0.80 & 0.03 \\
  GroundAnything-VLM & \textbf{90.38} & \textbf{92.93} & \textbf{91.64} & \textbf{27.75} & \textbf{28.45} & \textbf{28.09} & \textbf{74.58} & \textbf{76.54} & \textbf{75.55} \\
  GroundAnything & 83.16 & 89.25 & 86.10 & 26.30 & 27.45 & 26.86 & 69.73 & 70.35 & 70.04 \\
  \addlinespace[2pt]
  \multicolumn{10}{@{}l}{\hspace{0.4em}\strut\textbf{Vision-Language Models (10B--1T)}} \\
  Qwen3.6-27B\hspace{0.35em}\citep{qwen3627b} & 51.74 & 56.09 & 53.83 & 9.91 & 10.76 & 10.32 & 39.12 & 42.56 & 40.77 \\
  Qwen3-VL-32B\hspace{0.35em}\citep{qwen3vl} & 24.97 & 28.35 & 26.56 & 2.89 & 4.04 & 3.37 & 17.78 & 20.69 & 19.12 \\
  Qwen3.8-27B\hspace{0.35em}\citep{qwen38} & 44.31 & 47.44 & 45.82 & 8.92 & 9.56 & 9.23 & 33.28 & 35.40 & 34.30 \\
  Qwen3.5-35B-A3B\hspace{0.35em}\citep{qwen35} & 46.43 & 49.58 & 47.95 & 9.22 & 10.00 & 9.60 & 35.18 & 37.36 & 36.24 \\
  DeepSeek-VL2-Small-16B\textsuperscript{*}\hspace{0.35em}\citep{Deepseekvl2} & -- & -- & 16.00 & -- & -- & 3.90 & -- & -- & 12.70 \\
  DeepSeek-VL2-27B\textsuperscript{*}\hspace{0.35em}\citep{Deepseekvl2} & -- & -- & -- & -- & -- & -- & -- & -- & -- \\
  SEED1.5-VL\textsuperscript{*}\hspace{0.35em}\citep{SEED15VL} & -- & -- & 76.90 & -- & -- & 5.30 & -- & -- & 53.20 \\
  \addlinespace[2pt]
  \multicolumn{10}{@{}l}{\hspace{0.4em}\strut\textbf{Vision-Language Models (>1T)}} \\
  Qwen3.7-Max\hspace{0.35em}\citep{qwen37} & 37.56 & 43.69 & 40.40 & 7.20 & 8.65 & 7.86 & 28.94 & 33.84 & 31.20 \\
  Kimi-K2.6 & 55.64 & 59.00 & 57.27 & 9.21 & 9.91 & 9.55 & 41.38 & 44.04 & 42.67 \\
  Kimi-K3\hspace{0.35em}\citep{KimiK3} & 65.95 & 78.99 & 71.88 & 10.40 & 11.88 & 11.09 & 47.61 & 56.40 & 51.64 \\
  GPT-6 Astra & 89.01 & 84.63 & 86.76 & 12.90 & 12.39 & 12.64 & 66.54 & 63.61 & 65.04 \\
  \addlinespace[2pt]
  \bottomrule
  
  \end{NiceTabular}%
  }
  \endgroup
\end{table}

\paragraph{VisDrone.}
GroundAnything improves over Rex-Omni from 27.19 to 33.76 F1mIoU, but remains below GroundingDINO (34.47) and SenseNova-Vision (42.35).
GroundAnything-VLM scores 40.74.
Both variants have low F1 at IoU 0.95 (3.42 and 2.24), identifying precise tiny-object boundaries as a remaining limitation despite the stronger Dense200 results.
\begin{table}[htbp]
  \centering
  \BenchmarkTableFont
  \caption{\textbf{VisDrone.} Complete box-grounding metrics. Local BAGEL and DeepSeek runs did not reproduce the reference results, so only available external scores are reported for these models. The starred BAGEL scores are taken from Table~1 of \citet{SenseNovaVision7BMoT}. No matching external result is available for full DeepSeek-VL2; Small and Tiny checkpoint results are not substituted. The starred DeepSeek-VL2-Small and SEED1.5-VL scores are from Table~4 of \citet{rexomni}, also reported in Table~2 of \citet{locateanything}.}
  \label{tab:app-visdrone}
  \begingroup
  \fontsize{8}{9.7}\selectfont
  \setlength{\tabcolsep}{3pt}
  \renewcommand{\arraystretch}{1.15}
  \resizebox{\linewidth}{!}{%
  \begin{NiceTabular}{@{}>{\raggedright\arraybackslash}p{177pt}ccccccccc@{}}
  \CodeBefore
    \rowcolor{benchmarktype}{3}
    \rowcolor{benchmarktype}{5}
    \rowcolor{benchmarkpurple}{22}
    \rowcolor{benchmarkgreen}{23}
    \rowcolor{benchmarktype}{24}
    \rowcolor{benchmarktype}{32}
  \Body
  \toprule
  \textbf{Model} & \multicolumn{3}{c}{\textbf{IoU 0.50}} & \multicolumn{3}{c}{\textbf{IoU 0.95}} & \multicolumn{3}{c}{\textbf{mIoU}} \\
  \cmidrule(lr){2-4}\cmidrule(lr){5-7}\cmidrule(lr){8-10}
   & \BenchHead{R} & \BenchHead{P} & \BenchHead{F1} & \BenchHead{R} & \BenchHead{P} & \BenchHead{F1} & \BenchHead{R} & \BenchHead{P} & \BenchHead{F1} \\
  \midrule
  \multicolumn{10}{@{}l}{\hspace{0.4em}\strut\textbf{Open-set Specialized Detectors}} \\
  GroundingDINO\hspace{0.35em}\citep{groundingdino} & 32.38 & \textbf{87.09} & 47.21 & 2.81 & \textbf{7.42} & \textbf{4.08} & 23.65 & \textbf{63.54} & 34.47 \\
  \addlinespace[2pt]
  \multicolumn{10}{@{}l}{\hspace{0.4em}\strut\textbf{Vision-Language Models (<10B)}} \\
  Rex-Omni\hspace{0.35em}\citep{rexomni} & 42.47 & 52.33 & 46.89 & 1.07 & 1.27 & 1.16 & 24.78 & 30.12 & 27.19 \\
  LocateAnything Fast\hspace{0.35em}\citep{locateanything} & 13.58 & 15.44 & 14.45 & 0.85 & 0.97 & 0.90 & 9.27 & 10.44 & 9.82 \\
  LocateAnything Hybrid\hspace{0.35em}\citep{locateanything} & 40.66 & 46.99 & 43.59 & 2.34 & 2.64 & 2.48 & 26.78 & 30.62 & 28.57 \\
  LocateAnything Slow NTP\hspace{0.35em}\citep{locateanything} & 59.28 & 70.06 & 64.22 & 2.73 & 3.13 & 2.91 & 37.32 & 43.81 & 40.31 \\
  Qwen2.5-VL-7B\hspace{0.35em}\citep{qwen25vl} & 27.63 & 42.83 & 33.59 & 0.44 & 0.63 & 0.52 & 15.48 & 23.55 & 18.67 \\
  Qwen3-VL-2B\hspace{0.35em}\citep{qwen3vl} & 37.16 & 46.15 & 41.17 & 1.16 & 1.46 & 1.29 & 23.01 & 28.23 & 25.36 \\
  Qwen3-VL-4B\hspace{0.35em}\citep{qwen3vl} & 47.85 & 52.36 & 50.00 & 1.78 & 1.86 & 1.82 & 29.92 & 32.47 & 31.14 \\
  Qwen3-VL-8B\hspace{0.35em}\citep{qwen3vl} & 46.61 & 44.84 & 45.71 & 1.74 & 1.70 & 1.72 & 28.80 & 27.81 & 28.29 \\
  Qwen3.5-4B\hspace{0.35em}\citep{qwen35} & 54.93 & 50.78 & 52.77 & 2.60 & 2.53 & 2.57 & 34.91 & 32.71 & 33.77 \\
  Qwen3.5-9B\hspace{0.35em}\citep{qwen35} & 57.52 & 45.04 & 50.52 & 2.87 & 2.48 & 2.66 & 36.75 & 29.69 & 32.84 \\
  RynnBrain1.1\hspace{0.35em}\citep{RynnBrain112B} & 7.42 & 46.41 & 12.80 & 0.38 & 2.05 & 0.64 & 4.83 & 29.23 & 8.29 \\
  SenseNova-Vision\hspace{0.35em}\citep{SenseNovaVision7BMoT} & \textbf{65.10} & 69.99 & \textbf{67.45} & 2.86 & 3.07 & 2.96 & \textbf{40.91} & 43.88 & \textbf{42.35} \\
  MiMo-VL-7B-SFT\hspace{0.35em}\citep{MimoVL} & 26.42 & 28.73 & 27.53 & 0.14 & 0.15 & 0.15 & 12.60 & 13.88 & 13.21 \\
  MiMo-VL-7B-RL\hspace{0.35em}\citep{MimoVL} & 27.94 & 32.86 & 30.20 & 0.28 & 0.34 & 0.31 & 13.79 & 16.62 & 15.07 \\
  BAGEL\textsuperscript{*}\hspace{0.35em}\citep{BAGEL7BMoT} & -- & -- & -- & -- & -- & -- & -- & -- & 23.00 \\
  RynnBrain\hspace{0.35em}\citep{RynnBrain2B} & 8.78 & 32.85 & 13.86 & 0.20 & 0.75 & 0.32 & 4.97 & 18.52 & 7.84 \\
  GroundAnything-VLM & 57.81 & 69.16 & 62.98 & \textbf{3.17} & 3.71 & 3.42 & 37.55 & 44.52 & 40.74 \\
  GroundAnything & 46.94 & 54.60 & 50.48 & 2.09 & 2.42 & 2.24 & 32.52 & 35.10 & 33.76 \\
  \addlinespace[2pt]
  \multicolumn{10}{@{}l}{\hspace{0.4em}\strut\textbf{Vision-Language Models (10B--1T)}} \\
  Qwen3.6-27B\hspace{0.35em}\citep{qwen3627b} & 57.36 & 56.32 & 56.83 & 2.61 & 2.73 & 2.67 & 36.44 & 36.43 & 36.44 \\
  Qwen3-VL-32B\hspace{0.35em}\citep{qwen3vl} & 48.94 & 46.14 & 47.50 & 1.48 & 1.46 & 1.47 & 29.51 & 28.24 & 28.86 \\
  Qwen3.8-27B\hspace{0.35em}\citep{qwen38} & 56.44 & 49.01 & 52.46 & 2.76 & 2.62 & 2.69 & 35.48 & 31.42 & 33.33 \\
  Qwen3.5-35B-A3B\hspace{0.35em}\citep{qwen35} & 59.78 & 50.81 & 54.93 & 2.92 & 2.76 & 2.84 & 37.97 & 32.86 & 35.23 \\
  DeepSeek-VL2-Small-16B\textsuperscript{*}\hspace{0.35em}\citep{Deepseekvl2} & -- & -- & 35.80 & -- & -- & 1.70 & -- & -- & 23.30 \\
  DeepSeek-VL2-27B\textsuperscript{*}\hspace{0.35em}\citep{Deepseekvl2} & -- & -- & -- & -- & -- & -- & -- & -- & -- \\
  SEED1.5-VL\textsuperscript{*}\hspace{0.35em}\citep{SEED15VL} & -- & -- & 55.90 & -- & -- & 0.60 & -- & -- & 27.40 \\
  \addlinespace[2pt]
  \multicolumn{10}{@{}l}{\hspace{0.4em}\strut\textbf{Vision-Language Models (>1T)}} \\
  Qwen3.7-Max\hspace{0.35em}\citep{qwen37} & 60.27 & 71.40 & 65.36 & 2.84 & 3.45 & 3.11 & 38.28 & 45.52 & 41.59 \\
  Kimi-K2.6 & 41.36 & 65.79 & 50.79 & 1.20 & 1.73 & 1.42 & 24.42 & 38.73 & 29.95 \\
  Kimi-K3\hspace{0.35em}\citep{KimiK3} & 41.51 & 76.49 & 53.81 & 0.98 & 1.59 & 1.21 & 23.57 & 42.47 & 30.31 \\
  GPT-6 Astra & 64.60 & 65.83 & 65.18 & 2.69 & 2.80 & 2.74 & 36.70 & 37.58 & 37.12 \\
  \addlinespace[2pt]
  \bottomrule
  
  \end{NiceTabular}%
  }
  \endgroup
\end{table}

\paragraph{Why strict overlap is difficult for small objects.}
For equal axis-aligned boxes of width $w>0$ and height $h>0$ separated only horizontally by $\delta$, with $|\delta|<w$,
\begin{equation}
\operatorname{IoU}(\delta)=\frac{w-|\delta|}{w+|\delta|},
\qquad
\operatorname{IoU}\ge\eta
\ \Longleftrightarrow\ 
|\delta|\le w\frac{1-\eta}{1+\eta},\quad 0<\eta<1.
\label{eq:deepstack_iou_sensitivity}
\end{equation}
The intersection and union areas are $(w-|\delta|)h$ and $(w+|\delta|)h$; IoU is zero for $|\delta|\ge w$.
At $\eta=0.75$, the permissible shift is only $w/7$.
Under this translation-only model, the failure rate at a given width equals the probability of exceeding this displacement threshold, a prediction testable from measured offsets.
The calculation explains size sensitivity without identifying its architectural cause; size errors and missed instances require separate analysis.

\subsection{Referring Object Detection}
\label{app:referring_detection}

\paragraph{RefCOCOg val/test.}
On RefCOCOg, GroundAnything reaches 91.61 F1mIoU on validation and 91.10 on test, versus 84.37 and 83.56 for its AR counterpart. LocateAnything Hybrid scores 76.43 and 77.67.
The reported gains persist at IoU 0.95, where GroundAnything scores 88.63/87.15.
Within this evaluation, the improvement therefore reflects precise language-conditioned localization rather than only successful target identification.
\begin{table}[htbp]
  \centering
  \BenchmarkTableFont
  \caption{\textbf{RefCOCOg validation and test.} Complete recorded F1 metrics at IoU 0.50, 0.95, and mIoU. The starred BAGEL scores are taken from Table~1 of \citet{SenseNovaVision7BMoT}, because local runs did not reproduce the reported performance. The starred SEED1.5-VL scores are from Table~5 of \citet{rexomni}.}
  \label{tab:app-refcocog-splits}
  \begingroup
  \fontsize{8}{9.7}\selectfont
  \setlength{\tabcolsep}{3pt}
  \renewcommand{\arraystretch}{1.15}
  \resizebox{\linewidth}{!}{%
  \begin{NiceTabular}{@{}>{\raggedright\arraybackslash}p{177pt}cccccc@{}}
  \CodeBefore
    \rowcolor{benchmarktype}{3}
    \rowcolor{benchmarktype}{5}
    \rowcolor{benchmarkpurple}{22}
    \rowcolor{benchmarkgreen}{23}
    \rowcolor{benchmarktype}{24}
    \rowcolor{benchmarktype}{32}
  \Body
  \toprule
  \textbf{Model} & \multicolumn{3}{c}{\textbf{RefCOCOg val}} & \multicolumn{3}{c}{\textbf{RefCOCOg test}} \\
  \cmidrule(lr){2-4}\cmidrule(lr){5-7}
   & \BenchHead{F1@.50} & \BenchHead{F1@.95} & \BenchHead{F1mIoU} & \BenchHead{F1@.50} & \BenchHead{F1@.95} & \BenchHead{F1mIoU} \\
  \midrule
  \multicolumn{7}{@{}l}{\hspace{0.4em}\strut\textbf{Open-set Specialized Detectors}} \\
  GroundingDINO\hspace{0.35em}\citep{groundingdino} & 58.37 & 22.47 & 49.77 & 58.52 & 24.38 & 50.43 \\
  \addlinespace[2pt]
  \multicolumn{7}{@{}l}{\hspace{0.4em}\strut\textbf{Vision-Language Models (<10B)}} \\
  Rex-Omni\hspace{0.35em}\citep{rexomni} & 87.01 & 35.23 & 73.90 & 87.36 & 36.51 & 74.76 \\
  LocateAnything Fast\hspace{0.35em}\citep{locateanything} & 87.99 & 39.39 & 75.30 & 88.34 & 42.08 & 76.50 \\
  LocateAnything Hybrid\hspace{0.35em}\citep{locateanything} & 88.50 & 40.40 & 76.43 & 88.91 & 42.63 & 77.67 \\
  LocateAnything Slow NTP\hspace{0.35em}\citep{locateanything} & 88.18 & 34.81 & 74.93 & 88.64 & 36.99 & 76.46 \\
  Qwen2.5-VL-7B\hspace{0.35em}\citep{qwen25vl} & 78.19 & 15.10 & 61.86 & 78.49 & 16.31 & 62.93 \\
  Qwen3-VL-2B\hspace{0.35em}\citep{qwen3vl} & 85.25 & 32.13 & 71.80 & 85.87 & 33.88 & 72.64 \\
  Qwen3-VL-4B\hspace{0.35em}\citep{qwen3vl} & 88.53 & 36.61 & 75.27 & 88.69 & 36.61 & 75.88 \\
  Qwen3-VL-8B\hspace{0.35em}\citep{qwen3vl} & 88.92 & 35.61 & 75.79 & 89.26 & 37.03 & 76.26 \\
  Qwen3.5-4B\hspace{0.35em}\citep{qwen35} & 89.24 & 34.89 & 75.03 & 89.00 & 35.14 & 75.54 \\
  Qwen3.5-9B\hspace{0.35em}\citep{qwen35} & 89.89 & 36.63 & 76.20 & 89.42 & 36.89 & 76.28 \\
  RynnBrain1.1\hspace{0.35em}\citep{RynnBrain112B} & 83.21 & 24.36 & 67.66 & 83.75 & 25.14 & 68.24 \\
  SenseNova-Vision\hspace{0.35em}\citep{SenseNovaVision7BMoT} & 89.94 & 43.58 & 78.69 & 89.85 & 44.91 & 79.48 \\
  MiMo-VL-7B-SFT\hspace{0.35em}\citep{MimoVL} & 86.51 & 14.72 & 66.44 & 86.59 & 15.26 & 66.78 \\
  MiMo-VL-7B-RL\hspace{0.35em}\citep{MimoVL} & 88.28 & 12.56 & 64.89 & 87.69 & 12.92 & 64.19 \\
  BAGEL\textsuperscript{*}\hspace{0.35em}\citep{BAGEL7BMoT} & -- & -- & 76.40 & -- & -- & 77.80 \\
  RynnBrain\hspace{0.35em}\citep{RynnBrain2B} & 73.72 & 17.32 & 57.99 & 73.77 & 18.29 & 58.38 \\
  GroundAnything-VLM & \textbf{94.11} & 51.46 & 84.37 & \textbf{93.75} & 48.32 & 83.56 \\
  GroundAnything & 92.66 & \textbf{88.63} & \textbf{91.61} & 92.78 & \textbf{87.15} & \textbf{91.10} \\
  \addlinespace[2pt]
  \multicolumn{7}{@{}l}{\hspace{0.4em}\strut\textbf{Vision-Language Models (10B--1T)}} \\
  Qwen3.6-27B\hspace{0.35em}\citep{qwen3627b} & 91.66 & 38.19 & 77.76 & 91.36 & 38.80 & 78.43 \\
  Qwen3-VL-32B\hspace{0.35em}\citep{qwen3vl} & 86.27 & 34.17 & 73.79 & 86.03 & 34.83 & 74.02 \\
  Qwen3.8-27B\hspace{0.35em}\citep{qwen38} & 90.14 & 35.69 & 76.03 & 90.86 & 36.98 & 77.37 \\
  Qwen3.5-35B-A3B\hspace{0.35em}\citep{qwen35} & 90.92 & 38.46 & 77.32 & 90.88 & 39.09 & 78.38 \\
  DeepSeek-VL2-Small-16B\hspace{0.35em}\citep{Deepseekvl2} & 59.86 & 0.90 & 26.18 & 61.06 & 0.89 & 27.49 \\
  DeepSeek-VL2-27B\hspace{0.35em}\citep{Deepseekvl2} & 92.46 & 55.36 & 82.73 & 92.19 & 58.33 & 84.07 \\
  SEED1.5-VL\textsuperscript{*}\hspace{0.35em}\citep{SEED15VL} & 84.70 & 30.90 & 71.90 & 85.20 & 32.10 & 73.20 \\
  \addlinespace[2pt]
  \multicolumn{7}{@{}l}{\hspace{0.4em}\strut\textbf{Vision-Language Models (>1T)}} \\
  Qwen3.7-Max\hspace{0.35em}\citep{qwen37} & 91.56 & 48.45 & 80.54 & 92.36 & 48.75 & 81.71 \\
  Kimi-K2.6 & 80.15 & 41.17 & 70.33 & 81.19 & 42.36 & 71.87 \\
  Kimi-K3\hspace{0.35em}\citep{KimiK3} & 88.02 & 34.39 & 73.39 & 87.92 & 35.06 & 73.99 \\
  GPT-6 Astra & 92.13 & 35.89 & 74.98 & 89.93 & 40.25 & 78.91 \\
  \addlinespace[2pt]
  \bottomrule
  
  \end{NiceTabular}%
  }
  \endgroup
\end{table}

\paragraph{RefCOCO family.}
GroundAnything scores 88.49, 83.06, and 84.43 on RefCOCO, RefCOCOg, and RefCOCO+, giving an 85.33 mean versus 83.43 for GroundAnything-VLM.
All three exceed LocateAnything Fast, although the larger DeepSeek-VL2-27B remains stronger on these family entries.
These results support broad referring competence without claiming a universal lead over every model size or evaluation split.
\begin{table}[htbp]
  \centering
  \BenchmarkTableFont
  \caption{\textbf{RefCOCO family.} The three datasets are reported separately; their F1mIoU arithmetic mean is RefCOCO avg in the main text.}
  \label{tab:app-refcoco-family}
  \begingroup
  \fontsize{8}{9.7}\selectfont
  \setlength{\tabcolsep}{3pt}
  \renewcommand{\arraystretch}{1.15}
  \resizebox{\linewidth}{!}{%
  \begin{NiceTabular}{@{}>{\raggedright\arraybackslash}p{177pt}ccccccccc@{}}
  \CodeBefore
    \rowcolor{benchmarktype}{3}
    \rowcolor{benchmarktype}{5}
    \rowcolor{benchmarkpurple}{22}
    \rowcolor{benchmarkgreen}{23}
    \rowcolor{benchmarktype}{24}
    \rowcolor{benchmarktype}{31}
  \Body
  \toprule
  \textbf{Model} & \multicolumn{3}{c}{\textbf{RefCOCO}} & \multicolumn{3}{c}{\textbf{RefCOCOg}} & \multicolumn{3}{c}{\textbf{RefCOCOplus}} \\
  \cmidrule(lr){2-4}\cmidrule(lr){5-7}\cmidrule(lr){8-10}
   & \BenchHead{F1@.50} & \BenchHead{F1@.95} & \BenchHead{F1mIoU} & \BenchHead{F1@.50} & \BenchHead{F1@.95} & \BenchHead{F1mIoU} & \BenchHead{F1@.50} & \BenchHead{F1@.95} & \BenchHead{F1mIoU} \\
  \midrule
  \multicolumn{10}{@{}l}{\hspace{0.4em}\strut\textbf{Open-set Specialized Detectors}} \\
  GroundingDINO\hspace{0.35em}\citep{groundingdino} & 51.19 & 23.28 & 44.33 & 57.99 & 23.17 & 49.69 & 49.20 & 20.89 & 41.42 \\
  \addlinespace[2pt]
  \multicolumn{10}{@{}l}{\hspace{0.4em}\strut\textbf{Vision-Language Models (<10B)}} \\
  Rex-Omni\hspace{0.35em}\citep{rexomni} & 84.37 & 32.85 & 71.43 & 84.74 & 34.74 & 72.34 & 76.92 & 29.53 & 64.10 \\
  LocateAnything Fast\hspace{0.35em}\citep{locateanything} & 92.48 & 43.95 & 80.61 & 88.47 & 40.69 & 76.17 & 84.43 & 39.53 & 73.08 \\
  LocateAnything Hybrid\hspace{0.35em}\citep{locateanything} & 92.73 & 44.32 & 81.37 & 89.42 & 41.46 & 77.73 & 85.80 & 40.57 & 75.08 \\
  LocateAnything Slow NTP\hspace{0.35em}\citep{locateanything} & 92.43 & 38.43 & 80.21 & 88.65 & 35.63 & 76.08 & 85.22 & 35.69 & 73.90 \\
  Qwen2.5-VL-7B\hspace{0.35em}\citep{qwen25vl} & 82.42 & 17.71 & 66.93 & 75.28 & 15.56 & 60.28 & 73.04 & 16.39 & 59.27 \\
  Qwen3-VL-2B\hspace{0.35em}\citep{qwen3vl} & 88.33 & 26.65 & 72.39 & 85.89 & 25.73 & 69.81 & 81.21 & 24.34 & 66.47 \\
  Qwen3-VL-4B\hspace{0.35em}\citep{qwen3vl} & 92.09 & 32.75 & 77.63 & 88.79 & 32.25 & 74.73 & 86.76 & 30.94 & 73.36 \\
  Qwen3-VL-8B\hspace{0.35em}\citep{qwen3vl} & 91.11 & 29.87 & 75.83 & 88.74 & 28.82 & 73.33 & 86.43 & 29.02 & 71.98 \\
  Qwen3.5-4B\hspace{0.35em}\citep{qwen35} & 90.11 & 29.07 & 74.46 & 88.42 & 26.98 & 71.61 & 84.13 & 27.79 & 69.48 \\
  Qwen3.5-9B\hspace{0.35em}\citep{qwen35} & 91.97 & 35.09 & 78.31 & 89.53 & 33.47 & 75.27 & 87.45 & 33.90 & 74.64 \\
  RynnBrain1.1\hspace{0.35em}\citep{RynnBrain112B} & 82.61 & 20.88 & 65.80 & 84.28 & 21.48 & 67.57 & 72.45 & 19.18 & 57.52 \\
  SenseNova-Vision\hspace{0.35em}\citep{SenseNovaVision7BMoT} & 90.12 & 45.07 & 79.66 & 89.18 & 44.73 & 78.69 & 84.40 & 41.68 & 74.28 \\
  MiMo-VL-7B-SFT\hspace{0.35em}\citep{MimoVL} & 86.05 & 14.45 & 66.26 & 84.87 & 14.33 & 64.76 & 76.79 & 13.13 & 58.86 \\
  MiMo-VL-7B-RL\hspace{0.35em}\citep{MimoVL} & 90.44 & 14.15 & 68.45 & 87.69 & 13.11 & 64.80 & 84.83 & 13.55 & 64.42 \\
  BAGEL\hspace{0.35em}\citep{BAGEL7BMoT} & 79.39 & 25.09 & 65.23 & 77.69 & 22.20 & 62.30 & 68.34 & 20.45 & 54.45 \\
  RynnBrain\hspace{0.35em}\citep{RynnBrain2B} & 73.99 & 13.60 & 55.59 & 76.49 & 14.29 & 57.91 & 65.51 & 11.72 & 48.88 \\
  GroundAnything-VLM & \textbf{95.54} & 48.01 & 85.53 & 92.66 & 43.78 & 81.62 & \textbf{92.65} & 48.01 & 83.15 \\
  GroundAnything & 91.52 & \textbf{78.64} & 88.49 & 88.00 & \textbf{68.78} & 83.06 & 87.73 & \textbf{73.94} & 84.43 \\
  \addlinespace[2pt]
  \multicolumn{10}{@{}l}{\hspace{0.4em}\strut\textbf{Vision-Language Models (10B--1T)}} \\
  Qwen3.6-27B\hspace{0.35em}\citep{qwen3627b} & 93.46 & 39.83 & 80.78 & 91.33 & 38.69 & 78.37 & 89.57 & 38.36 & 77.44 \\
  Qwen3-VL-32B\hspace{0.35em}\citep{qwen3vl} & 90.33 & 34.50 & 77.21 & 84.49 & 31.54 & 71.75 & 85.34 & 33.08 & 73.17 \\
  Qwen3.8-27B\hspace{0.35em}\citep{qwen38} & 92.93 & 38.44 & 79.83 & 90.37 & 37.15 & 76.91 & 89.07 & 37.76 & 76.91 \\
  Qwen3.5-35B-A3B\hspace{0.35em}\citep{qwen35} & 92.57 & 36.44 & 79.30 & 90.62 & 34.77 & 76.49 & 87.61 & 34.38 & 75.06 \\
  DeepSeek-VL2-Small-16B\hspace{0.35em}\citep{Deepseekvl2} & 66.95 & 1.10 & 30.95 & 64.47 & 1.29 & 29.60 & 64.93 & 1.25 & 30.58 \\
  DeepSeek-VL2-27B\hspace{0.35em}\citep{Deepseekvl2} & 94.73 & 74.35 & \textbf{90.43} & \textbf{94.02} & 65.74 & \textbf{87.32} & 91.33 & 71.62 & \textbf{87.19} \\
  \addlinespace[2pt]
  \multicolumn{10}{@{}l}{\hspace{0.4em}\strut\textbf{Vision-Language Models (>1T)}} \\
  Qwen3.7-Max\hspace{0.35em}\citep{qwen37} & 95.11 & 45.55 & 82.99 & 92.78 & 44.87 & 80.38 & 91.78 & 32.41 & 77.23 \\
  Kimi-K2.6 & 82.55 & 39.91 & 72.27 & 82.06 & 40.18 & 71.83 & 76.00 & 33.87 & 64.89 \\
  Kimi-K3\hspace{0.35em}\citep{KimiK3} & 88.46 & 35.78 & 75.12 & 88.57 & 34.51 & 74.14 & 81.86 & 33.77 & 69.31 \\
  GPT-6 Astra & 93.61 & 37.14 & 79.45 & 87.21 & 44.27 & 76.15 & 90.45 & 36.30 & 77.84 \\
  \addlinespace[2pt]
  \bottomrule
  
  \end{NiceTabular}%
  }
  \endgroup
\end{table}

\subsection{Object Pointing}
\label{app:object_pointing}

Following \citet{rexomni}, SAM~\citep{SAM} converts ground-truth boxes to object masks. A predicted point is correct when it lies inside the corresponding mask; F1@Point balances point precision and recall.

\paragraph{Referring pointing.}
On RefCOCOg val/test, GroundAnything-VLM reaches 90.44 and 91.03 F1@Point.
GroundAnything scores 85.75/85.73, modestly exceeding Rex-Omni (84.96/85.32) while remaining below its AR counterpart.
Unlike referring boxes, referring points do not improve after conversion in these results, demonstrating that the diffusion--AR trade-off depends on the output primitive.
\begin{table}[htbp]
  \centering
  \BenchmarkTableFont
  \caption{\textbf{Referring object pointing.} Starred BAGEL scores are retained despite unreliable support for the unified pointing protocol. Kimi-K3, both MiMo variants, and both DeepSeek variants are N/A under that protocol. These outcomes do not establish intrinsic pointing capability. The starred Molmo and SEED1.5-VL scores are from Table~7 of \citet{rexomni}; its Molmo checkpoint is Molmo-7B-D.}
  \label{tab:app-point-referring}
  \begingroup
  \fontsize{8}{9.7}\selectfont
  \setlength{\tabcolsep}{4pt}
  \renewcommand{\arraystretch}{1.15}
  \resizebox{\linewidth}{!}{%
  \begin{NiceTabular}{@{}>{\raggedright\arraybackslash}p{177pt}>{\centering\arraybackslash}p{110pt}>{\centering\arraybackslash}p{110pt}@{}}
  \CodeBefore
    \rowcolor{benchmarktype}{3}
    \rowcolor{benchmarktype}{5}
    \rowcolor{benchmarkpurple}{23}
    \rowcolor{benchmarkgreen}{24}
    \rowcolor{benchmarktype}{25}
    \rowcolor{benchmarktype}{33}
  \Body
  \toprule
  \textbf{Model} & \multicolumn{1}{c}{\textbf{RefCOCOg val}} & \multicolumn{1}{c}{\textbf{RefCOCOg test}} \\
  \cmidrule(lr){2-2}\cmidrule(lr){3-3}
   & \BenchHead{F1@Point} & \BenchHead{F1@Point} \\
  \midrule
  \multicolumn{3}{@{}l}{\hspace{0.4em}\strut\textbf{Open-set Specialized Detectors}} \\
  GroundingDINO\hspace{0.35em}\citep{groundingdino} & 49.34 & 49.97 \\
  \addlinespace[2pt]
  \multicolumn{3}{@{}l}{\hspace{0.4em}\strut\textbf{Vision-Language Models (<10B)}} \\
  Rex-Omni\hspace{0.35em}\citep{rexomni} & 84.96 & 85.32 \\
  LocateAnything Fast\hspace{0.35em}\citep{locateanything} & 73.84 & 74.94 \\
  LocateAnything Hybrid\hspace{0.35em}\citep{locateanything} & 75.89 & 76.65 \\
  LocateAnything Slow NTP\hspace{0.35em}\citep{locateanything} & 77.17 & 77.59 \\
  Qwen2.5-VL-7B\hspace{0.35em}\citep{qwen25vl} & 81.65 & 82.48 \\
  Qwen3-VL-2B\hspace{0.35em}\citep{qwen3vl} & 76.17 & 76.01 \\
  Qwen3-VL-4B\hspace{0.35em}\citep{qwen3vl} & 76.43 & 77.64 \\
  Qwen3-VL-8B\hspace{0.35em}\citep{qwen3vl} & 81.97 & 82.09 \\
  Qwen3.5-4B\hspace{0.35em}\citep{qwen35} & 79.35 & 79.31 \\
  Qwen3.5-9B\hspace{0.35em}\citep{qwen35} & 77.59 & 77.85 \\
  RynnBrain1.1\hspace{0.35em}\citep{RynnBrain112B} & 74.42 & 74.17 \\
  SenseNova-Vision\hspace{0.35em}\citep{SenseNovaVision7BMoT} & 74.63 & 75.42 \\
  MiMo-VL-7B-SFT\hspace{0.35em}\citep{MimoVL} & N/A\textsuperscript{*} & N/A\textsuperscript{*} \\
  MiMo-VL-7B-RL\hspace{0.35em}\citep{MimoVL} & N/A\textsuperscript{*} & N/A\textsuperscript{*} \\
  BAGEL\hspace{0.35em}\citep{BAGEL7BMoT} & 55.92\textsuperscript{*} & 54.57\textsuperscript{*} \\
  RynnBrain\hspace{0.35em}\citep{RynnBrain2B} & 73.34 & 73.67 \\
  Molmo-7B\textsuperscript{*}\hspace{0.35em}\citep{Molmo} & 83.70 & 83.60 \\
  GroundAnything-VLM & \textbf{90.44} & \textbf{91.03} \\
  GroundAnything & 85.75 & 85.73 \\
  \addlinespace[2pt]
  \multicolumn{3}{@{}l}{\hspace{0.4em}\strut\textbf{Vision-Language Models (10B--1T)}} \\
  Qwen3.6-27B\hspace{0.35em}\citep{qwen3627b} & 82.73 & 82.88 \\
  Qwen3-VL-32B\hspace{0.35em}\citep{qwen3vl} & 76.52 & 76.23 \\
  Qwen3.8-27B\hspace{0.35em}\citep{qwen38} & 75.84 & 75.86 \\
  Qwen3.5-35B-A3B\hspace{0.35em}\citep{qwen35} & 75.35 & 76.09 \\
  DeepSeek-VL2-Small-16B\hspace{0.35em}\citep{Deepseekvl2} & N/A\textsuperscript{*} & N/A\textsuperscript{*} \\
  DeepSeek-VL2-27B\hspace{0.35em}\citep{Deepseekvl2} & N/A\textsuperscript{*} & N/A\textsuperscript{*} \\
  SEED1.5-VL\textsuperscript{*}\hspace{0.35em}\citep{SEED15VL} & 83.60 & 84.20 \\
  \addlinespace[2pt]
  \multicolumn{3}{@{}l}{\hspace{0.4em}\strut\textbf{Vision-Language Models (>1T)}} \\
  Qwen3.7-Max\hspace{0.35em}\citep{qwen37} & 71.21 & 72.40 \\
  Kimi-K2.6 & 39.59\textsuperscript{\textdagger} & 39.60\textsuperscript{\textdagger} \\
  Kimi-K3\hspace{0.35em}\citep{KimiK3} & N/A\textsuperscript{*} & N/A\textsuperscript{*} \\
  GPT-6 Astra & 87.80 & 84.90 \\
  \addlinespace[2pt]
  \bottomrule
  
  \end{NiceTabular}%
  }
  \endgroup
\end{table}

\paragraph{Common and long-tailed pointing.}
GroundAnything-VLM attains 84.92 on COCO and 79.81 on LVIS; GroundAnything decreases to 71.32 and 64.57.
For the DLM variant, precision is lower than recall on both datasets: 66.42 versus 77.00 on COCO and 58.67 versus 71.80 on LVIS.
This imbalance points to excess or incorrectly localized point predictions as an important source of the gap, rather than missed instances alone.
\begin{table}[htbp]
  \centering
  \BenchmarkTableFont
  \caption{\textbf{Object pointing on COCO and LVIS.} Starred BAGEL scores are retained despite unreliable support for the unified pointing protocol. Kimi-K3, both MiMo variants, and both DeepSeek variants are N/A under that protocol. These outcomes do not establish intrinsic pointing capability. The starred Molmo and SEED1.5-VL scores are from Table~7 of \citet{rexomni}; its Molmo checkpoint is Molmo-7B-D.}
  \label{tab:app-point-common}
  \begingroup
  \fontsize{8}{9.7}\selectfont
  \setlength{\tabcolsep}{3pt}
  \renewcommand{\arraystretch}{1.15}
  \resizebox{\linewidth}{!}{%
  \begin{NiceTabular}{@{}>{\raggedright\arraybackslash}p{177pt}cccccc@{}}
  \CodeBefore
    \rowcolor{benchmarktype}{3}
    \rowcolor{benchmarktype}{5}
    \rowcolor{benchmarkpurple}{23}
    \rowcolor{benchmarkgreen}{24}
    \rowcolor{benchmarktype}{25}
    \rowcolor{benchmarktype}{33}
  \Body
  \toprule
  \textbf{Model} & \multicolumn{3}{c}{\textbf{COCO}} & \multicolumn{3}{c}{\textbf{LVIS}} \\
  \cmidrule(lr){2-4}\cmidrule(lr){5-7}
   & \BenchHead{R@Point} & \BenchHead{P@Point} & \BenchHead{F1@Point} & \BenchHead{R@Point} & \BenchHead{P@Point} & \BenchHead{F1@Point} \\
  \midrule
  \multicolumn{7}{@{}l}{\hspace{0.4em}\strut\textbf{Open-set Specialized Detectors}} \\
  GroundingDINO\hspace{0.35em}\citep{groundingdino} & 68.92 & 71.97 & 70.41 & 44.91 & 71.15 & 55.07 \\
  \addlinespace[2pt]
  \multicolumn{7}{@{}l}{\hspace{0.4em}\strut\textbf{Vision-Language Models (<10B)}} \\
  Rex-Omni\hspace{0.35em}\citep{rexomni} & 77.81 & 81.77 & 79.74 & 63.46 & 78.14 & 70.04 \\
  LocateAnything Fast\hspace{0.35em}\citep{locateanything} & 68.99 & 74.26 & 71.53 & 55.81 & 70.05 & 62.13 \\
  LocateAnything Hybrid\hspace{0.35em}\citep{locateanything} & 73.80 & 73.77 & 73.78 & 60.96 & 69.36 & 64.89 \\
  LocateAnything Slow NTP\hspace{0.35em}\citep{locateanything} & 74.68 & 75.05 & 74.86 & 62.36 & 72.42 & 67.01 \\
  Qwen2.5-VL-7B\hspace{0.35em}\citep{qwen25vl} & 61.24 & 65.76 & 63.42 & 46.46 & 56.82 & 51.12 \\
  Qwen3-VL-2B\hspace{0.35em}\citep{qwen3vl} & 55.65 & 54.98 & 55.31 & 43.70 & 50.11 & 46.69 \\
  Qwen3-VL-4B\hspace{0.35em}\citep{qwen3vl} & 63.13 & 67.69 & 65.33 & 49.45 & 62.17 & 55.08 \\
  Qwen3-VL-8B\hspace{0.35em}\citep{qwen3vl} & 64.92 & 66.76 & 65.83 & 52.25 & 61.41 & 56.46 \\
  Qwen3.5-4B\hspace{0.35em}\citep{qwen35} & 70.15 & 68.85 & 69.50 & 55.61 & 64.72 & 59.82 \\
  Qwen3.5-9B\hspace{0.35em}\citep{qwen35} & 71.87 & 72.56 & 72.21 & 58.90 & 70.07 & 64.00 \\
  RynnBrain1.1\hspace{0.35em}\citep{RynnBrain112B} & 21.06 & 32.96 & 25.70 & 13.35 & 25.63 & 17.56 \\
  SenseNova-Vision\hspace{0.35em}\citep{SenseNovaVision7BMoT} & 70.85 & 75.21 & 72.96 & 57.21 & 69.26 & 62.66 \\
  MiMo-VL-7B-SFT\hspace{0.35em}\citep{MimoVL} & N/A\textsuperscript{*} & N/A\textsuperscript{*} & N/A\textsuperscript{*} & N/A\textsuperscript{*} & N/A\textsuperscript{*} & N/A\textsuperscript{*} \\
  MiMo-VL-7B-RL\hspace{0.35em}\citep{MimoVL} & N/A\textsuperscript{*} & N/A\textsuperscript{*} & N/A\textsuperscript{*} & N/A\textsuperscript{*} & N/A\textsuperscript{*} & N/A\textsuperscript{*} \\
  BAGEL\hspace{0.35em}\citep{BAGEL7BMoT} & 33.57\textsuperscript{*} & 39.16\textsuperscript{*} & 36.15\textsuperscript{*} & 22.73\textsuperscript{*} & 31.65\textsuperscript{*} & 26.46\textsuperscript{*} \\
  RynnBrain\hspace{0.35em}\citep{RynnBrain2B} & 9.46 & 17.73 & 12.33 & 5.38 & 13.73 & 7.73 \\
  Molmo-7B\textsuperscript{*}\hspace{0.35em}\citep{Molmo} & -- & -- & 77.30 & -- & -- & 40.30 \\
  GroundAnything-VLM & 84.77 & \textbf{85.08} & \textbf{84.92} & \textbf{76.56} & \textbf{83.35} & \textbf{79.81} \\
  GroundAnything & 77.00 & 66.42 & 71.32 & 71.80 & 58.67 & 64.57 \\
  \addlinespace[2pt]
  \multicolumn{7}{@{}l}{\hspace{0.4em}\strut\textbf{Vision-Language Models (10B--1T)}} \\
  Qwen3.6-27B\hspace{0.35em}\citep{qwen3627b} & 74.01 & 71.98 & 72.98 & 62.98 & 72.61 & 67.45 \\
  Qwen3-VL-32B\hspace{0.35em}\citep{qwen3vl} & 72.70 & 70.24 & 71.45 & 60.44 & 66.42 & 63.29 \\
  Qwen3.8-27B\hspace{0.35em}\citep{qwen38} & 72.69 & 75.38 & 74.01 & 61.98 & 74.47 & 67.65 \\
  Qwen3.5-35B-A3B\hspace{0.35em}\citep{qwen35} & 72.21 & 71.42 & 71.81 & 61.05 & 70.99 & 65.65 \\
  DeepSeek-VL2-Small-16B\hspace{0.35em}\citep{Deepseekvl2} & N/A\textsuperscript{*} & N/A\textsuperscript{*} & N/A\textsuperscript{*} & N/A\textsuperscript{*} & N/A\textsuperscript{*} & N/A\textsuperscript{*} \\
  DeepSeek-VL2-27B\hspace{0.35em}\citep{Deepseekvl2} & N/A\textsuperscript{*} & N/A\textsuperscript{*} & N/A\textsuperscript{*} & N/A\textsuperscript{*} & N/A\textsuperscript{*} & N/A\textsuperscript{*} \\
  SEED1.5-VL\textsuperscript{*}\hspace{0.35em}\citep{SEED15VL} & -- & -- & 78.20 & -- & -- & 70.70 \\
  \addlinespace[2pt]
  \multicolumn{7}{@{}l}{\hspace{0.4em}\strut\textbf{Vision-Language Models (>1T)}} \\
  Qwen3.7-Max\hspace{0.35em}\citep{qwen37} & 71.61 & 72.65 & 72.13 & 60.97 & 74.22 & 66.94 \\
  Kimi-K2.6 & 29.83\textsuperscript{\textdagger} & 31.52\textsuperscript{\textdagger} & 30.65\textsuperscript{\textdagger} & 23.20\textsuperscript{\textdagger} & 28.32\textsuperscript{\textdagger} & 25.51\textsuperscript{\textdagger} \\
  Kimi-K3\hspace{0.35em}\citep{KimiK3} & N/A\textsuperscript{*} & N/A\textsuperscript{*} & N/A\textsuperscript{*} & N/A\textsuperscript{*} & N/A\textsuperscript{*} & N/A\textsuperscript{*} \\
  GPT-6 Astra & \textbf{86.56} & 78.14 & 82.17 & 74.92 & 79.50 & 77.14 \\
  \addlinespace[2pt]
  \bottomrule
  
  \end{NiceTabular}%
  }
  \endgroup
\end{table}

\paragraph{Dense and tiny-object pointing.}
GroundAnything scores 77.14 on Dense200 and 57.94 on VisDrone, exceeding Rex-Omni on both.
GroundAnything-VLM reaches 84.16 and 68.03, while GPT-6 Astra remains stronger on Dense200 at 86.57.
The remaining gap to the AR variant confirms that strong dense box grounding does not automatically guarantee equally strong point-set prediction.
\begin{table}[htbp]
  \centering
  \BenchmarkTableFont
  \caption{\textbf{Object pointing on Dense200 and VisDrone.} Starred BAGEL scores are retained despite unreliable support for the unified pointing protocol. Kimi-K3, both MiMo variants, and both DeepSeek variants are N/A under that protocol. These outcomes do not establish intrinsic pointing capability. The starred Molmo and SEED1.5-VL scores are from Table~7 of \citet{rexomni}; its Molmo checkpoint is Molmo-7B-D.}
  \label{tab:app-point-dense}
  \begingroup
  \fontsize{8}{9.7}\selectfont
  \setlength{\tabcolsep}{3pt}
  \renewcommand{\arraystretch}{1.15}
  \resizebox{\linewidth}{!}{%
  \begin{NiceTabular}{@{}>{\raggedright\arraybackslash}p{177pt}cccccc@{}}
  \CodeBefore
    \rowcolor{benchmarktype}{3}
    \rowcolor{benchmarktype}{5}
    \rowcolor{benchmarkpurple}{23}
    \rowcolor{benchmarkgreen}{24}
    \rowcolor{benchmarktype}{25}
    \rowcolor{benchmarktype}{33}
  \Body
  \toprule
  \textbf{Model} & \multicolumn{3}{c}{\textbf{Dense200}} & \multicolumn{3}{c}{\textbf{VisDrone}} \\
  \cmidrule(lr){2-4}\cmidrule(lr){5-7}
   & \BenchHead{R@Point} & \BenchHead{P@Point} & \BenchHead{F1@Point} & \BenchHead{R@Point} & \BenchHead{P@Point} & \BenchHead{F1@Point} \\
  \midrule
  \multicolumn{7}{@{}l}{\hspace{0.4em}\strut\textbf{Open-set Specialized Detectors}} \\
  GroundingDINO\hspace{0.35em}\citep{groundingdino} & 22.00 & 65.45 & 32.93 & 27.20 & 71.76 & 39.45 \\
  \addlinespace[2pt]
  \multicolumn{7}{@{}l}{\hspace{0.4em}\strut\textbf{Vision-Language Models (<10B)}} \\
  Rex-Omni\hspace{0.35em}\citep{rexomni} & 75.59 & 77.76 & 76.66 & 47.81 & 56.92 & 51.97 \\
  LocateAnything Fast\hspace{0.35em}\citep{locateanything} & 64.63 & 66.65 & 65.63 & 55.22 & 57.55 & 56.36 \\
  LocateAnything Hybrid\hspace{0.35em}\citep{locateanything} & 77.43 & 78.73 & 78.07 & 59.18 & 55.54 & 57.30 \\
  LocateAnything Slow NTP\hspace{0.35em}\citep{locateanything} & 78.87 & 81.39 & 80.11 & 61.17 & 60.77 & 60.97 \\
  Qwen2.5-VL-7B\hspace{0.35em}\citep{qwen25vl} & 12.12 & 36.42 & 18.19 & 12.81 & 18.21 & 15.04 \\
  Qwen3-VL-2B\hspace{0.35em}\citep{qwen3vl} & 14.06 & 15.95 & 14.95 & 10.64 & 11.98 & 11.27 \\
  Qwen3-VL-4B\hspace{0.35em}\citep{qwen3vl} & 14.13 & 46.93 & 21.72 & 21.27 & 26.25 & 23.50 \\
  Qwen3-VL-8B\hspace{0.35em}\citep{qwen3vl} & 20.61 & 32.96 & 25.36 & 17.68 & 18.54 & 18.10 \\
  Qwen3.5-4B\hspace{0.35em}\citep{qwen35} & 56.82 & 60.30 & 58.51 & 32.72 & 31.57 & 32.13 \\
  Qwen3.5-9B\hspace{0.35em}\citep{qwen35} & 61.02 & 70.33 & 65.35 & 44.42 & 45.04 & 44.73 \\
  RynnBrain1.1\hspace{0.35em}\citep{RynnBrain112B} & 2.10 & 34.50 & 3.95 & 7.68 & 46.55 & 13.18 \\
  SenseNova-Vision\hspace{0.35em}\citep{SenseNovaVision7BMoT} & 74.86 & 82.98 & 78.71 & 59.36 & 64.48 & 61.81 \\
  MiMo-VL-7B-SFT\hspace{0.35em}\citep{MimoVL} & N/A\textsuperscript{*} & N/A\textsuperscript{*} & N/A\textsuperscript{*} & N/A\textsuperscript{*} & N/A\textsuperscript{*} & N/A\textsuperscript{*} \\
  MiMo-VL-7B-RL\hspace{0.35em}\citep{MimoVL} & N/A\textsuperscript{*} & N/A\textsuperscript{*} & N/A\textsuperscript{*} & N/A\textsuperscript{*} & N/A\textsuperscript{*} & N/A\textsuperscript{*} \\
  BAGEL\hspace{0.35em}\citep{BAGEL7BMoT} & 8.48\textsuperscript{*} & 16.84\textsuperscript{*} & 11.28\textsuperscript{*} & 5.09\textsuperscript{*} & 7.33\textsuperscript{*} & 6.01\textsuperscript{*} \\
  RynnBrain\hspace{0.35em}\citep{RynnBrain2B} & 0.51 & 1.00 & 0.68 & 5.64 & 34.46 & 9.69 \\
  Molmo-7B\textsuperscript{*}\hspace{0.35em}\citep{Molmo} & -- & -- & 33.10 & -- & -- & 29.20 \\
  GroundAnything-VLM & 81.93 & \textbf{86.51} & 84.16 & 61.77 & \textbf{75.70} & \textbf{68.03} \\
  GroundAnything & 76.50 & 77.79 & 77.14 & 56.27 & 59.72 & 57.94 \\
  \addlinespace[2pt]
  \multicolumn{7}{@{}l}{\hspace{0.4em}\strut\textbf{Vision-Language Models (10B--1T)}} \\
  Qwen3.6-27B\hspace{0.35em}\citep{qwen3627b} & 70.52 & 75.69 & 73.01 & 46.68 & 43.94 & 45.27 \\
  Qwen3-VL-32B\hspace{0.35em}\citep{qwen3vl} & 44.04 & 46.11 & 45.05 & 27.88 & 25.71 & 26.75 \\
  Qwen3.8-27B\hspace{0.35em}\citep{qwen38} & 73.97 & 75.15 & 74.55 & 51.07 & 51.25 & 51.16 \\
  Qwen3.5-35B-A3B\hspace{0.35em}\citep{qwen35} & 68.54 & 72.89 & 70.65 & 45.64 & 43.92 & 44.76 \\
  DeepSeek-VL2-Small-16B\hspace{0.35em}\citep{Deepseekvl2} & N/A\textsuperscript{*} & N/A\textsuperscript{*} & N/A\textsuperscript{*} & N/A\textsuperscript{*} & N/A\textsuperscript{*} & N/A\textsuperscript{*} \\
  DeepSeek-VL2-27B\hspace{0.35em}\citep{Deepseekvl2} & N/A\textsuperscript{*} & N/A\textsuperscript{*} & N/A\textsuperscript{*} & N/A\textsuperscript{*} & N/A\textsuperscript{*} & N/A\textsuperscript{*} \\
  SEED1.5-VL\textsuperscript{*}\hspace{0.35em}\citep{SEED15VL} & -- & -- & 72.10 & -- & -- & 56.70 \\
  \addlinespace[2pt]
  \multicolumn{7}{@{}l}{\hspace{0.4em}\strut\textbf{Vision-Language Models (>1T)}} \\
  Qwen3.7-Max\hspace{0.35em}\citep{qwen37} & 57.56 & 78.12 & 66.28 & 55.04 & 63.65 & 59.04 \\
  Kimi-K2.6 & 36.14\textsuperscript{\textdagger} & 34.27\textsuperscript{\textdagger} & 35.18\textsuperscript{\textdagger} & 11.22\textsuperscript{\textdagger} & 17.37\textsuperscript{\textdagger} & 13.63\textsuperscript{\textdagger} \\
  Kimi-K3\hspace{0.35em}\citep{KimiK3} & N/A\textsuperscript{*} & N/A\textsuperscript{*} & N/A\textsuperscript{*} & N/A\textsuperscript{*} & N/A\textsuperscript{*} & N/A\textsuperscript{*} \\
  GPT-6 Astra & \textbf{89.52} & 83.74 & \textbf{86.57} & \textbf{64.39} & 66.99 & 65.62 \\
  \addlinespace[2pt]
  \bottomrule
  
  \end{NiceTabular}%
  }
  \endgroup
\end{table}

\subsection{Robot and Spatial Pointing}
\label{app:robot_spatial_pointing}

\paragraph{Relational localization and placement.}
GroundAnything retains the AR variant's 62.34\% RefSpatial Unseen accuracy and reaches 69.67\% on RoboSpatial Context, above GPT-6 Astra's 65.69\%.
Its RefSpatial Location/Placement scores are 59.00/67.00, below the AR variant's 67.00/71.00 and GPT-6 Astra's 86.00/85.86.
Parallel grounding therefore transfers to relational and free-space targets, while more difficult spatial reasoning remains a clear source of headroom.
The external RoboRefer comparison uses the setting without a depth prior.
\begin{table}[p]
  \centering
  \BenchmarkTableFont
  \caption{\textbf{Robot and spatial pointing.} Point-in-mask accuracy is reported for each dataset. The starred RefSpatial baselines follow Table~11 of \citet{rexomni}; RoboRefer uses the setting without a depth prior. Values retain the precision recorded in our evaluation tables.}
  \label{tab:app-robot-spatial}
  \begingroup
  \fontsize{8}{9.7}\selectfont
  \setlength{\tabcolsep}{4pt}
  \renewcommand{\arraystretch}{1.10}
  \resizebox{\linewidth}{!}{%
  \begin{NiceTabular}{@{}>{\raggedright\arraybackslash}p{177pt}cccc@{}}
  \CodeBefore
    \rowcolor{benchmarktype}{3}
    \rowcolor{benchmarktype}{5}
    \rowcolor{benchmarkpurple}{24}
    \rowcolor{benchmarkgreen}{25}
    \rowcolor{benchmarktype}{26}
    \rowcolor{benchmarktype}{37}
  \Body
  \toprule
  \textbf{Model} & \multicolumn{4}{c}{\textbf{Point-in-mask accuracy}} \\
  \cmidrule(lr){2-5}
   & \BenchHead{RefSpatial\\Location} & \BenchHead{RefSpatial\\Placement} & \BenchHead{RefSpatial\\Unseen} & \BenchHead{RoboSpatial\\Context} \\
  \midrule
  \multicolumn{5}{@{}l}{\hspace{0.4em}\strut\textbf{Open-set Specialized Detectors}} \\
  GroundingDINO\hspace{0.35em}\citep{groundingdino} & 26.50 & 2.00 & 4.33 & 4.92 \\
  \addlinespace[2pt]
  \multicolumn{5}{@{}l}{\hspace{0.4em}\strut\textbf{Vision-Language Models (<10B)}} \\
  Rex-Omni\hspace{0.35em}\citep{rexomni} & 51.00 & 52.50 & 37.01 & 59.02 \\
  LocateAnything Fast\hspace{0.35em}\citep{locateanything} & 53.00 & 19.00 & 16.88 & 15.57 \\
  LocateAnything Hybrid\hspace{0.35em}\citep{locateanything} & 55.00 & 17.33 & 20.78 & 14.75 \\
  LocateAnything Slow NTP\hspace{0.35em}\citep{locateanything} & 54.00 & 27.20 & 16.88 & 16.39 \\
  Qwen2.5-VL-7B\hspace{0.35em}\citep{qwen25vl} & 43.00 & 15.50 & 18.18 & 26.23 \\
  Qwen3-VL-2B\hspace{0.35em}\citep{qwen3vl} & 42.00 & 24.00 & 11.69 & 32.79 \\
  Qwen3-VL-4B\hspace{0.35em}\citep{qwen3vl} & 48.00 & 50.00 & 27.27 & 64.75 \\
  Qwen3-VL-8B\hspace{0.35em}\citep{qwen3vl} & 51.00 & 45.00 & 28.57 & 59.02 \\
  Qwen3.5-4B\hspace{0.35em}\citep{qwen35} & 65.00 & 38.00 & 38.96 & 50.82 \\
  Qwen3.5-9B\hspace{0.35em}\citep{qwen35} & 65.00 & 46.83 & 37.01 & 60.66 \\
  RynnBrain1.1\hspace{0.35em}\citep{RynnBrain112B} & 49.70 & 51.50 & 36.90 & 54.10 \\
  SenseNova-Vision\hspace{0.35em}\citep{SenseNovaVision7BMoT} & 34.51 & 6.25 & 8.54 & 0.82 \\
  MiMo-VL-7B-SFT\hspace{0.35em}\citep{MimoVL} & 1.00\textsuperscript{\textdagger} & 8.03\textsuperscript{\textdagger} & 4.64\textsuperscript{\textdagger} & 4.10\textsuperscript{\textdagger} \\
  MiMo-VL-7B-RL\hspace{0.35em}\citep{MimoVL} & 1.00\textsuperscript{\textdagger} & 2.20\textsuperscript{\textdagger} & 2.61\textsuperscript{\textdagger} & 4.92\textsuperscript{\textdagger} \\
  BAGEL\hspace{0.35em}\citep{BAGEL7BMoT} & 51.79 & 12.01 & 23.38 & 13.93 \\
  RynnBrain\hspace{0.35em}\citep{RynnBrain2B} & 42.00 & 37.00 & 22.08 & 28.69 \\
  Molmo-7B\textsuperscript{*}\hspace{0.35em}\citep{Molmo} & 21.90 & 12.80 & 12.20 & -- \\
  RoboRefer\textsuperscript{*}\hspace{0.35em}\citep{RoboRefer2B} & 51.00 & 49.00 & 39.00 & -- \\
  GroundAnything-VLM & 67.00 & 71.00 & 62.34 & \textbf{72.13} \\
  GroundAnything & 59.00 & 67.00 & 62.34 & 69.67 \\
  \addlinespace[2pt]
  \multicolumn{5}{@{}l}{\hspace{0.4em}\strut\textbf{Vision-Language Models (10B--1T)}} \\
  Qwen3.6-27B\hspace{0.35em}\citep{qwen3627b} & 72.00 & 66.00 & 61.04 & 63.93 \\
  Qwen3-VL-32B\hspace{0.35em}\citep{qwen3vl} & 62.00 & 52.00 & 41.56 & 63.93 \\
  Qwen3.8-27B\hspace{0.35em}\citep{qwen38} & 64.00 & 56.00 & 46.75 & 63.93 \\
  Qwen3.5-35B-A3B\hspace{0.35em}\citep{qwen35} & 70.00 & 58.00 & 53.25 & 71.31 \\
  DeepSeek-VL2-Small-16B\hspace{0.35em}\citep{Deepseekvl2} & 2.11\textsuperscript{\textdagger} & 1.33\textsuperscript{\textdagger} & 0.32\textsuperscript{\textdagger} & 5.74\textsuperscript{\textdagger} \\
  DeepSeek-VL2-27B\hspace{0.35em}\citep{Deepseekvl2} & 3.75\textsuperscript{\textdagger} & 0.66\textsuperscript{\textdagger} & 0.72\textsuperscript{\textdagger} & 5.74\textsuperscript{\textdagger} \\
  SpaceLLaVA\textsuperscript{*} & 5.80 & 4.30 & 4.00 & -- \\
  RoboPoint\textsuperscript{*}\hspace{0.35em}\citep{RoboPoint13B} & 22.90 & 9.30 & 8.40 & -- \\
  Molmo-72B\textsuperscript{*}\hspace{0.35em}\citep{Molmo} & 45.80 & 14.70 & 21.20 & -- \\
  Gemini-2.5-Pro\textsuperscript{*}\hspace{0.35em}\citep{Gemini2.5Pro} & 47.00 & 24.20 & 27.10 & -- \\
  \addlinespace[2pt]
  \multicolumn{5}{@{}l}{\hspace{0.4em}\strut\textbf{Vision-Language Models (>1T)}} \\
  Qwen3.7-Max\hspace{0.35em}\citep{qwen37} & 71.50 & 66.00 & 57.14 & 69.67 \\
  Kimi-K2.6 & 54.09 & 28.57 & 41.56 & 30.33 \\
  Kimi-K3\hspace{0.35em}\citep{KimiK3} & 67.85 & 50.00 & 54.98 & 54.92 \\
  GPT-6 Astra & \textbf{86.00} & \textbf{85.86} & \textbf{81.93} & 65.69 \\
  \addlinespace[2pt]
  \bottomrule
  
  \end{NiceTabular}%
  }
  \endgroup
\end{table}

\subsection{OCR}
\label{app:ocr}

\paragraph{HierText and ICDAR2015.}
GroundAnything obtains 33.31/41.81 F1mIoU, compared with 41.33/42.50 for GroundAnything-VLM.
Both variants have zero reported parse errors on these two datasets, so their quality gap cannot be explained by malformed responses alone.
The DLM variant exceeds LocateAnything Fast and Hybrid on both datasets but trails Rex-Omni; the external SenseNova-Vision entries provide F1mIoU only and do not support conclusions about its parsing reliability.
\begin{table}[htbp]
  \centering
  \BenchmarkTableFont
  \caption{\textbf{OCR on HierText and ICDAR2015.} Each dataset reports four loose-match F1 measures and parse-error rate. GroundingDINO is N/A because OCR is unsupported. Kimi-K3 and both DeepSeek variants are N/A because their outputs do not satisfy the evaluation protocol; this does not establish a lack of OCR capability. SenseNova-Vision uses the HierText and ICDAR2015 F1mIoU scores of 31.20 and 49.50 reported in Table~1 of \citet{SenseNovaVision7BMoT}, because its local evaluation prompts could not be aligned. Other metrics for these datasets are unavailable; TotalText and SROIE use local results. The starred PaddleOCRv5 and SEED1.5-VL scores use the BBOX results in Table~10 of \citet{rexomni}.}
  \label{tab:app-ocr-hiertext-icdar}
  \begingroup
  \fontsize{8}{9.7}\selectfont
  \setlength{\tabcolsep}{2.5pt}
  \renewcommand{\arraystretch}{1.15}
  \resizebox{\linewidth}{!}{%
  \begin{NiceTabular}{@{}>{\raggedright\arraybackslash}p{177pt}cccccccccc@{}}
  \CodeBefore
    \rowcolor{benchmarktype}{3}
    \rowcolor{benchmarktype}{5}
    \rowcolor{benchmarktype}{7}
    \rowcolor{benchmarkpurple}{24}
    \rowcolor{benchmarkgreen}{25}
    \rowcolor{benchmarktype}{26}
    \rowcolor{benchmarktype}{34}
  \Body
  \toprule
  \textbf{Model} & \multicolumn{5}{c}{\textbf{HierText}} & \multicolumn{5}{c}{\textbf{ICDAR2015}} \\
  \cmidrule(lr){2-6}\cmidrule(lr){7-11}
   & \BenchHead{F1@.50} & \BenchHead{F1@.75} & \BenchHead{F1@.95} & \BenchHead{F1mIoU} & \BenchHead{Parse\\err.} & \BenchHead{F1@.50} & \BenchHead{F1@.75} & \BenchHead{F1@.95} & \BenchHead{F1mIoU} & \BenchHead{Parse\\err.} \\
  \midrule
  \multicolumn{11}{@{}l}{\hspace{0.4em}\strut\textbf{Closed-set Specialized Detectors}} \\
  PaddleOCRv5\textsuperscript{*}\hspace{0.35em}\citep{PaddleOCRv5} & 45.20 & -- & 3.40 & 30.50 & -- & 38.20 & -- & 1.20 & 25.60 & -- \\
  \addlinespace[2pt]
  \multicolumn{11}{@{}l}{\hspace{0.4em}\strut\textbf{Open-set Specialized Detectors}} \\
  GroundingDINO\hspace{0.35em}\citep{groundingdino} & N/A\textsuperscript{*} & N/A\textsuperscript{*} & N/A\textsuperscript{*} & N/A\textsuperscript{*} & N/A\textsuperscript{*} & N/A\textsuperscript{*} & N/A\textsuperscript{*} & N/A\textsuperscript{*} & N/A\textsuperscript{*} & N/A\textsuperscript{*} \\
  \addlinespace[2pt]
  \multicolumn{11}{@{}l}{\hspace{0.4em}\strut\textbf{Vision-Language Models (<10B)}} \\
  Rex-Omni\hspace{0.35em}\citep{rexomni} & 54.16 & 35.67 & 2.10 & 34.46 & 1.92 & 73.39 & 50.07 & 0.96 & 45.65 & \textbf{0.00} \\
  LocateAnything Fast\hspace{0.35em}\citep{locateanything} & 29.91 & 25.83 & 3.13 & 22.59 & 31.46 & 44.51 & 28.32 & 0.40 & 26.73 & 3.43 \\
  LocateAnything Hybrid\hspace{0.35em}\citep{locateanything} & 35.50 & 30.43 & 3.61 & 26.65 & 19.04 & 45.61 & 29.41 & 0.40 & 27.48 & 1.21 \\
  LocateAnything Slow NTP\hspace{0.35em}\citep{locateanything} & 58.34 & \textbf{48.65} & \textbf{5.28} & 42.94 & 1.10 & 53.07 & 33.29 & 0.65 & 31.81 & \textbf{0.00} \\
  Qwen2.5-VL-7B\hspace{0.35em}\citep{qwen25vl} & 29.61 & 14.28 & 0.51 & 15.49 & \textbf{0.00} & 55.50 & 23.37 & 1.19 & 27.72 & \textbf{0.00} \\
  Qwen3-VL-2B\hspace{0.35em}\citep{qwen3vl} & 25.08 & 11.38 & 0.33 & 12.65 & 1.74 & 49.47 & 24.52 & 1.93 & 27.43 & 0.40 \\
  Qwen3-VL-4B\hspace{0.35em}\citep{qwen3vl} & 41.06 & 23.42 & 0.97 & 23.48 & 0.93 & 51.52 & 26.37 & 1.67 & 28.41 & 0.20 \\
  Qwen3-VL-8B\hspace{0.35em}\citep{qwen3vl} & 42.32 & 23.47 & 0.99 & 23.89 & 0.29 & 54.20 & 29.71 & 1.40 & 30.82 & \textbf{0.00} \\
  Qwen3.5-4B\hspace{0.35em}\citep{qwen35} & 30.47 & 16.50 & 0.76 & 17.00 & 12.65 & 41.16 & 16.20 & 0.37 & 19.61 & \textbf{0.00} \\
  Qwen3.5-9B\hspace{0.35em}\citep{qwen35} & 48.60 & 30.10 & 1.90 & 29.63 & 13.93 & 56.02 & 26.65 & 0.87 & 29.90 & \textbf{0.00} \\
  RynnBrain1.1\hspace{0.35em}\citep{RynnBrain112B} & 3.55 & 2.04 & 0.11 & 2.07 & 30.24 & 36.78 & 15.63 & 0.20 & 17.81 & 6.05 \\
  SenseNova-Vision\textsuperscript{*}\hspace{0.35em}\citep{SenseNovaVision7BMoT} & -- & -- & -- & 31.20 & -- & -- & -- & -- & \textbf{49.50} & -- \\
  MiMo-VL-7B-SFT\hspace{0.35em}\citep{MimoVL} & 24.77 & 12.82 & 0.36 & 13.44 & 1.33 & 64.31 & 30.58 & 0.46 & 34.33 & 1.01 \\
  MiMo-VL-7B-RL\hspace{0.35em}\citep{MimoVL} & 26.13 & 13.01 & 0.32 & 13.88 & 0.29 & 56.84 & 24.32 & 0.37 & 28.79 & 7.86 \\
  BAGEL\hspace{0.35em}\citep{BAGEL7BMoT} & 11.91 & 3.30 & 0.05 & 4.87 & 0.75 & 36.29 & 10.41 & 0.17 & 15.48 & 0.20 \\
  RynnBrain\hspace{0.35em}\citep{RynnBrain2B} & 4.08 & 1.92 & 0.06 & 2.13 & 0.64 & 36.05 & 15.15 & 0.47 & 17.81 & 0.60 \\
  GroundAnything-VLM & 60.18 & 45.45 & 3.82 & 41.33 & \textbf{0.00} & 64.52 & 46.35 & 2.02 & 42.50 & \textbf{0.00} \\
  GroundAnything & 46.97 & 39.47 & 4.97 & 33.31 & \textbf{0.00} & 58.53 & 50.80 & \textbf{2.09} & 41.81 & \textbf{0.00} \\
  \addlinespace[2pt]
  \multicolumn{11}{@{}l}{\hspace{0.4em}\strut\textbf{Vision-Language Models (10B--1T)}} \\
  Qwen3.6-27B\hspace{0.35em}\citep{qwen3627b} & 36.27 & 25.16 & 1.94 & 23.49 & 4.12 & 53.83 & 32.09 & 1.75 & 31.64 & 0.20 \\
  Qwen3-VL-32B\hspace{0.35em}\citep{qwen3vl} & 15.55 & 9.08 & 0.49 & 9.00 & 54.32 & 42.26 & 26.11 & 1.69 & 25.42 & 25.40 \\
  Qwen3.8-27B\hspace{0.35em}\citep{qwen38} & 51.50 & 35.06 & 2.46 & 32.95 & 0.06 & 68.67 & 39.98 & 0.95 & 39.72 & \textbf{0.00} \\
  Qwen3.5-35B-A3B\hspace{0.35em}\citep{qwen35} & 49.83 & 33.22 & 2.16 & 31.38 & 3.71 & 51.85 & 25.41 & 1.22 & 28.04 & 3.02 \\
  DeepSeek-VL2-Small-16B\hspace{0.35em}\citep{Deepseekvl2} & N/A\textsuperscript{*} & N/A\textsuperscript{*} & N/A\textsuperscript{*} & N/A\textsuperscript{*} & N/A\textsuperscript{*} & N/A\textsuperscript{*} & N/A\textsuperscript{*} & N/A\textsuperscript{*} & N/A\textsuperscript{*} & N/A\textsuperscript{*} \\
  DeepSeek-VL2-27B\hspace{0.35em}\citep{Deepseekvl2} & N/A\textsuperscript{*} & N/A\textsuperscript{*} & N/A\textsuperscript{*} & N/A\textsuperscript{*} & N/A\textsuperscript{*} & N/A\textsuperscript{*} & N/A\textsuperscript{*} & N/A\textsuperscript{*} & N/A\textsuperscript{*} & N/A\textsuperscript{*} \\
  SEED1.5-VL\textsuperscript{*}\hspace{0.35em}\citep{SEED15VL} & 27.10 & -- & 0.20 & 12.00 & -- & 38.60 & -- & 0.00 & 18.70 & -- \\
  \addlinespace[2pt]
  \multicolumn{11}{@{}l}{\hspace{0.4em}\strut\textbf{Vision-Language Models (>1T)}} \\
  Qwen3.7-Max\hspace{0.35em}\citep{qwen37} & \textbf{62.98} & 47.45 & 3.93 & \textbf{42.95} & \textbf{0.00} & 69.09 & 45.15 & 1.97 & 43.07 & \textbf{0.00} \\
  Kimi-K2.6 & 45.89\textsuperscript{\textdagger} & 24.25\textsuperscript{\textdagger} & 1.25\textsuperscript{\textdagger} & 25.26\textsuperscript{\textdagger} & 0.12\textsuperscript{\textdagger} & 56.81\textsuperscript{\textdagger} & 31.14\textsuperscript{\textdagger} & 1.95\textsuperscript{\textdagger} & 32.25\textsuperscript{\textdagger} & \textbf{0.00}\textsuperscript{\textdagger} \\
  Kimi-K3\hspace{0.35em}\citep{KimiK3} & N/A\textsuperscript{*} & N/A\textsuperscript{*} & N/A\textsuperscript{*} & N/A\textsuperscript{*} & N/A\textsuperscript{*} & N/A\textsuperscript{*} & N/A\textsuperscript{*} & N/A\textsuperscript{*} & N/A\textsuperscript{*} & N/A\textsuperscript{*} \\
  GPT-6 Astra & 61.35 & 41.89 & 3.38 & 39.58 & 0.23 & \textbf{79.02} & \textbf{55.72} & 0.87 & 48.87 & \textbf{0.00} \\
  \addlinespace[2pt]
  \bottomrule
  
  \end{NiceTabular}%
  }
  \endgroup
\end{table}

\paragraph{TotalText and SROIE.}
GroundAnything scores 43.08 on TotalText and 43.64 on SROIE, versus 49.03 and 71.15 for its AR counterpart.
The 27.51 pp SROIE gap is substantially larger than the 5.95 pp TotalText gap, showing that diffusion's quality cost varies markedly across OCR settings.
Although GroundAnything exceeds LocateAnything Hybrid on SROIE, it remains below Rex-Omni and the dedicated PaddleOCRv5 reference.
Its parse-error rates are unreported on these datasets and must not be treated as zero.
\begin{table}[htbp]
  \centering
  \BenchmarkTableFont
  \caption{\textbf{OCR on TotalText and SROIE.} Each dataset reports four loose-match F1 measures and parse-error rate. GroundingDINO is N/A because OCR is unsupported. Kimi-K3 and both DeepSeek variants are N/A because their outputs do not satisfy the evaluation protocol; this does not establish a lack of OCR capability. The starred PaddleOCRv5 and SEED1.5-VL scores use the BBOX results in Table~10 of \citet{rexomni}.}
  \label{tab:app-ocr-totaltext-sroie}
  \begingroup
  \fontsize{8}{9.7}\selectfont
  \setlength{\tabcolsep}{2.5pt}
  \renewcommand{\arraystretch}{1.15}
  \resizebox{\linewidth}{!}{%
  \begin{NiceTabular}{@{}>{\raggedright\arraybackslash}p{177pt}cccccccccc@{}}
  \CodeBefore
    \rowcolor{benchmarktype}{3}
    \rowcolor{benchmarktype}{5}
    \rowcolor{benchmarktype}{7}
    \rowcolor{benchmarkpurple}{24}
    \rowcolor{benchmarkgreen}{25}
    \rowcolor{benchmarktype}{26}
    \rowcolor{benchmarktype}{34}
  \Body
  \toprule
  \textbf{Model} & \multicolumn{5}{c}{\textbf{TotalText}} & \multicolumn{5}{c}{\textbf{SROIE}} \\
  \cmidrule(lr){2-6}\cmidrule(lr){7-11}
   & \BenchHead{F1@.50} & \BenchHead{F1@.75} & \BenchHead{F1@.95} & \BenchHead{F1mIoU} & \BenchHead{Parse\\err.} & \BenchHead{F1@.50} & \BenchHead{F1@.75} & \BenchHead{F1@.95} & \BenchHead{F1mIoU} & \BenchHead{Parse\\err.} \\
  \midrule
  \multicolumn{11}{@{}l}{\hspace{0.4em}\strut\textbf{Closed-set Specialized Detectors}} \\
  PaddleOCRv5\textsuperscript{*}\hspace{0.35em}\citep{PaddleOCRv5} & 40.20 & -- & 0.70 & 25.70 & -- & 77.70 & -- & 5.60 & 58.60 & -- \\
  \addlinespace[2pt]
  \multicolumn{11}{@{}l}{\hspace{0.4em}\strut\textbf{Open-set Specialized Detectors}} \\
  GroundingDINO\hspace{0.35em}\citep{groundingdino} & N/A\textsuperscript{*} & N/A\textsuperscript{*} & N/A\textsuperscript{*} & N/A\textsuperscript{*} & N/A\textsuperscript{*} & N/A\textsuperscript{*} & N/A\textsuperscript{*} & N/A\textsuperscript{*} & N/A\textsuperscript{*} & N/A\textsuperscript{*} \\
  \addlinespace[2pt]
  \multicolumn{11}{@{}l}{\hspace{0.4em}\strut\textbf{Vision-Language Models (<10B)}} \\
  Rex-Omni\hspace{0.35em}\citep{rexomni} & 74.04 & 59.21 & 4.47 & 52.35 & \textbf{0.00} & 72.28 & 47.65 & 2.49 & 48.35 & 3.61 \\
  LocateAnything Fast\hspace{0.35em}\citep{locateanything} & 60.96 & 49.64 & 5.48 & 44.40 & 3.00 & 33.18 & 29.73 & 1.86 & 24.89 & 45.83 \\
  LocateAnything Hybrid\hspace{0.35em}\citep{locateanything} & 62.40 & 51.07 & 5.66 & 45.49 & 1.33 & 40.53 & 35.84 & 2.12 & 30.05 & 34.17 \\
  LocateAnything Slow NTP\hspace{0.35em}\citep{locateanything} & 70.79 & 53.40 & 4.88 & 49.13 & 0.33 & 88.23 & 77.85 & 4.49 & 65.49 & 1.11 \\
  Qwen2.5-VL-7B\hspace{0.35em}\citep{qwen25vl} & 56.62 & 28.47 & 2.54 & 31.13 & \textbf{0.00} & 30.76 & 14.04 & 0.35 & 15.58 & \textbf{0.00} \\
  Qwen3-VL-2B\hspace{0.35em}\citep{qwen3vl} & 60.76 & 29.25 & 6.45 & 33.55 & \textbf{0.00} & 34.86 & 12.51 & 0.15 & 15.93 & \textbf{0.00} \\
  Qwen3-VL-4B\hspace{0.35em}\citep{qwen3vl} & 65.02 & 36.49 & 7.75 & 38.35 & \textbf{0.00} & 68.73 & 41.96 & 0.98 & 40.41 & \textbf{0.00} \\
  Qwen3-VL-8B\hspace{0.35em}\citep{qwen3vl} & 61.40 & 35.44 & 7.40 & 36.70 & \textbf{0.00} & 49.41 & 24.33 & 0.46 & 25.88 & \textbf{0.00} \\
  Qwen3.5-4B\hspace{0.35em}\citep{qwen35} & 54.40 & 28.83 & 4.75 & 31.56 & \textbf{0.00} & 45.96 & 20.70 & 0.30 & 23.21 & 1.11 \\
  Qwen3.5-9B\hspace{0.35em}\citep{qwen35} & 63.78 & 35.22 & 4.51 & 37.26 & 1.00 & 51.52 & 20.59 & 0.81 & 26.74 & 1.94 \\
  RynnBrain1.1\hspace{0.35em}\citep{RynnBrain112B} & 30.38 & 13.04 & 4.29 & 16.90 & 10.67 & 7.23 & 3.24 & 0.06 & 3.59 & 0.83 \\
  SenseNova-Vision\hspace{0.35em}\citep{SenseNovaVision7BMoT} & 18.30 & 11.21 & 1.02 & 11.40 & \textbf{0.00} & 46.97 & 41.90 & 4.23 & 36.26 & \textbf{0.00} \\
  MiMo-VL-7B-SFT\hspace{0.35em}\citep{MimoVL} & 64.24 & 42.31 & 1.79 & 39.76 & \textbf{0.00} & 30.13 & 13.02 & 0.36 & 15.01 & 5.28 \\
  MiMo-VL-7B-RL\hspace{0.35em}\citep{MimoVL} & 65.24 & 40.22 & 2.19 & 39.55 & 0.67 & 39.77 & 15.71 & 0.47 & 19.00 & 0.28 \\
  BAGEL\hspace{0.35em}\citep{BAGEL7BMoT} & 51.19 & 19.02 & 1.02 & 24.25 & \textbf{0.00} & 18.63 & 4.94 & 0.07 & 7.74 & 1.11 \\
  RynnBrain\hspace{0.35em}\citep{RynnBrain2B} & 35.58 & 17.41 & 1.57 & 18.90 & \textbf{0.00} & 7.08 & 2.37 & 0.02 & 3.19 & 4.72 \\
  GroundAnything-VLM & 72.66 & 53.35 & 4.95 & 49.03 & \textbf{0.00} & \textbf{91.85} & \textbf{83.60} & \textbf{7.06} & \textbf{71.15} & \textbf{0.00} \\
  GroundAnything & 60.46 & 46.01 & 5.41 & 43.08 & -- & 49.31 & 44.85 & 3.02 & 43.64 & -- \\
  \addlinespace[2pt]
  \multicolumn{11}{@{}l}{\hspace{0.4em}\strut\textbf{Vision-Language Models (10B--1T)}} \\
  Qwen3.6-27B\hspace{0.35em}\citep{qwen3627b} & 63.52 & 42.49 & 5.38 & 41.21 & 2.33 & 52.05 & 29.37 & 0.63 & 29.37 & \textbf{0.00} \\
  Qwen3-VL-32B\hspace{0.35em}\citep{qwen3vl} & 44.51 & 28.93 & \textbf{8.27} & 28.84 & 18.67 & 10.12 & 6.31 & 0.20 & 6.02 & 73.89 \\
  Qwen3.8-27B\hspace{0.35em}\citep{qwen38} & 65.73 & 42.57 & 3.99 & 41.57 & \textbf{0.00} & 61.27 & 38.97 & 1.70 & 37.24 & 0.28 \\
  Qwen3.5-35B-A3B\hspace{0.35em}\citep{qwen35} & 59.86 & 34.72 & 5.31 & 35.95 & 1.67 & 65.83 & 36.31 & 0.92 & 36.64 & 0.28 \\
  DeepSeek-VL2-Small-16B\hspace{0.35em}\citep{Deepseekvl2} & N/A\textsuperscript{*} & N/A\textsuperscript{*} & N/A\textsuperscript{*} & N/A\textsuperscript{*} & N/A\textsuperscript{*} & N/A\textsuperscript{*} & N/A\textsuperscript{*} & N/A\textsuperscript{*} & N/A\textsuperscript{*} & N/A\textsuperscript{*} \\
  DeepSeek-VL2-27B\hspace{0.35em}\citep{Deepseekvl2} & N/A\textsuperscript{*} & N/A\textsuperscript{*} & N/A\textsuperscript{*} & N/A\textsuperscript{*} & N/A\textsuperscript{*} & N/A\textsuperscript{*} & N/A\textsuperscript{*} & N/A\textsuperscript{*} & N/A\textsuperscript{*} & N/A\textsuperscript{*} \\
  SEED1.5-VL\textsuperscript{*}\hspace{0.35em}\citep{SEED15VL} & 35.00 & -- & 0.30 & 19.50 & -- & 51.90 & -- & 0.80 & 28.10 & -- \\
  \addlinespace[2pt]
  \multicolumn{11}{@{}l}{\hspace{0.4em}\strut\textbf{Vision-Language Models (>1T)}} \\
  Qwen3.7-Max\hspace{0.35em}\citep{qwen37} & 72.58 & 52.18 & 7.92 & 48.64 & \textbf{0.00} & 57.72 & 44.25 & 1.50 & 38.54 & \textbf{0.00} \\
  Kimi-K2.6 & 67.66\textsuperscript{\textdagger} & 40.90\textsuperscript{\textdagger} & 5.52\textsuperscript{\textdagger} & 40.99\textsuperscript{\textdagger} & \textbf{0.00}\textsuperscript{\textdagger} & 76.94\textsuperscript{\textdagger} & 49.43\textsuperscript{\textdagger} & 1.92\textsuperscript{\textdagger} & 46.83\textsuperscript{\textdagger} & \textbf{0.00}\textsuperscript{\textdagger} \\
  Kimi-K3\hspace{0.35em}\citep{KimiK3} & N/A\textsuperscript{*} & N/A\textsuperscript{*} & N/A\textsuperscript{*} & N/A\textsuperscript{*} & N/A\textsuperscript{*} & N/A\textsuperscript{*} & N/A\textsuperscript{*} & N/A\textsuperscript{*} & N/A\textsuperscript{*} & N/A\textsuperscript{*} \\
  GPT-6 Astra & \textbf{74.61} & \textbf{62.13} & 4.96 & \textbf{53.55} & \textbf{0.00} & 70.81 & 62.54 & 5.27 & 53.57 & \textbf{0.00} \\
  \addlinespace[2pt]
  \bottomrule
  
  \end{NiceTabular}%
  }
  \endgroup
\end{table}

\subsection{GUI Grounding}
\label{app:gui_grounding}

\paragraph{ScreenSpot-Pro.}
GroundAnything improves overall action accuracy from 65.34\% for GroundAnything-VLM to 75.96\%, with gains in every reported text/icon domain subset.
For example, CAD icon accuracy rises from 48.44\% to 71.88\%, and scientific-interface icon accuracy from 61.82\% to 81.82\%.
These gains support precise parallel localization in complex interfaces, although GPT-6 Astra remains stronger overall at 93.17\%.
Uncertain prompt/protocol results are not used to infer intrinsic GUI capability.
\begin{table}[htbp]
  \centering
  \BenchmarkTableFont
  \caption{\textbf{ScreenSpot-Pro.} Action accuracy is broken down by domain and target type, followed by overall action accuracy and parse-error rate. GroundingDINO is N/A because GUI grounding is unsupported. BAGEL is N/A because its outputs do not satisfy the unified protocol. The starred JEDI, UI-R1, and UI-TARS scores are from Table~8 of \citet{rexomni}; GUI-Owl-32B scores are from Table~3 of \citet{locateanything}.}
  \label{tab:app-gui-pro}
  \begingroup
  \fontsize{7}{8.7}\selectfont
  \setlength{\tabcolsep}{2pt}
  \renewcommand{\arraystretch}{1.15}
  \resizebox{\linewidth}{!}{%
  \begin{NiceTabular}{@{}>{\raggedright\arraybackslash}p{161pt}cccccccccccccc@{}}
  \CodeBefore
    \rowcolor{benchmarktype}{3}
    \rowcolor{benchmarktype}{5}
    \rowcolor{benchmarkpurple}{25}
    \rowcolor{benchmarkgreen}{26}
    \rowcolor{benchmarktype}{27}
    \rowcolor{benchmarktype}{35}
  \Body
  \toprule
  \textbf{Model} & \multicolumn{2}{c}{\textbf{Dev}} & \multicolumn{2}{c}{\textbf{Creative}} & \multicolumn{2}{c}{\textbf{CAD}} & \multicolumn{2}{c}{\textbf{Sci}} & \multicolumn{2}{c}{\textbf{Office}} & \multicolumn{2}{c}{\textbf{OS}} & \multicolumn{2}{c}{\textbf{Overall}} \\
  \cmidrule(lr){2-3}\cmidrule(lr){4-5}\cmidrule(lr){6-7}\cmidrule(lr){8-9}\cmidrule(lr){10-11}\cmidrule(lr){12-13}\cmidrule(lr){14-15}
   & \BenchHead{Text} & \BenchHead{Icon} & \BenchHead{Text} & \BenchHead{Icon} & \BenchHead{Text} & \BenchHead{Icon} & \BenchHead{Text} & \BenchHead{Icon} & \BenchHead{Text} & \BenchHead{Icon} & \BenchHead{Text} & \BenchHead{Icon} & \BenchHead{Action\\acc.} & \BenchHead{Parse\\err.} \\
  \midrule
  \multicolumn{15}{@{}l}{\hspace{0.4em}\strut\textbf{Open-set Specialized Detectors}} \\
  GroundingDINO\hspace{0.35em}\citep{groundingdino} & N/A\textsuperscript{*} & N/A\textsuperscript{*} & N/A\textsuperscript{*} & N/A\textsuperscript{*} & N/A\textsuperscript{*} & N/A\textsuperscript{*} & N/A\textsuperscript{*} & N/A\textsuperscript{*} & N/A\textsuperscript{*} & N/A\textsuperscript{*} & N/A\textsuperscript{*} & N/A\textsuperscript{*} & N/A\textsuperscript{*} & N/A\textsuperscript{*} \\
  \addlinespace[2pt]
  \multicolumn{15}{@{}l}{\hspace{0.4em}\strut\textbf{Vision-Language Models (<10B)}} \\
  Rex-Omni\hspace{0.35em}\citep{rexomni} & 61.04 & 9.66 & 53.03 & 12.59 & 23.35 & 9.38 & 57.64 & 26.36 & 65.54 & 24.53 & 42.06 & 13.48 & 36.75 & -- \\
  LocateAnything Fast\hspace{0.35em}\citep{locateanything} & 68.18 & 44.83 & 59.60 & 32.87 & 58.38 & 35.94 & 71.53 & 53.64 & 75.14 & 56.60 & 51.40 & 37.08 & 56.04 & -- \\
  LocateAnything Hybrid\hspace{0.35em}\citep{locateanything} & 70.13 & 44.83 & 59.60 & 36.36 & 58.88 & 37.50 & 71.53 & 51.82 & 72.88 & 58.49 & 58.88 & 40.45 & 57.05 & -- \\
  LocateAnything Slow NTP\hspace{0.35em}\citep{locateanything} & 70.78 & 48.28 & 61.11 & 39.86 & 60.41 & 39.06 & 75.69 & 51.82 & 74.58 & 54.72 & 57.94 & 40.45 & 58.57 & -- \\
  Qwen2.5-VL-7B\hspace{0.35em}\citep{qwen25vl} & 40.91 & 3.45 & 36.36 & 9.09 & 18.27 & 3.12 & 47.22 & 6.36 & 56.50 & 13.21 & 34.58 & 11.24 & 26.57 & -- \\
  Qwen3-VL-2B\hspace{0.35em}\citep{qwen3vl} & 47.40 & 7.59 & 29.29 & 8.39 & 23.86 & 7.81 & 38.89 & 18.18 & 49.15 & 22.64 & 37.38 & 20.22 & 27.77 & -- \\
  Qwen3-VL-4B\hspace{0.35em}\citep{qwen3vl} & 72.73 & 31.72 & 66.67 & 25.17 & 57.87 & 26.56 & 77.08 & 35.45 & 84.75 & 47.17 & 78.50 & 34.83 & 56.74 & -- \\
  Qwen3-VL-8B\hspace{0.35em}\citep{qwen3vl} & 75.32 & 30.34 & 71.21 & 20.98 & 60.91 & 26.56 & 76.39 & 40.00 & 83.62 & 37.74 & 73.83 & 33.71 & 56.86 & -- \\
  Qwen3.5-4B\hspace{0.35em}\citep{qwen35} & 79.22 & 42.07 & 67.68 & 25.17 & 65.99 & 35.94 & 74.31 & 34.55 & 82.49 & 50.94 & 69.16 & 43.82 & 59.27 & -- \\
  Qwen3.5-9B\hspace{0.35em}\citep{qwen35} & 70.78 & 37.93 & 64.14 & 30.77 & 40.10 & 28.12 & 72.92 & 33.64 & 74.58 & 49.06 & 69.16 & 38.20 & 53.13 & -- \\
  RynnBrain1.1\hspace{0.35em}\citep{RynnBrain112B} & 52.60 & 15.86 & 45.96 & 12.59 & 21.32 & 10.94 & 52.08 & 21.82 & 60.45 & 20.75 & 47.66 & 20.22 & 34.66 & 0.38 \\
  SenseNova-Vision\hspace{0.35em}\citep{SenseNovaVision7BMoT} & N/A\textsuperscript{\textdagger} & N/A\textsuperscript{\textdagger} & N/A\textsuperscript{\textdagger} & N/A\textsuperscript{\textdagger} & N/A\textsuperscript{\textdagger} & N/A\textsuperscript{\textdagger} & N/A\textsuperscript{\textdagger} & N/A\textsuperscript{\textdagger} & N/A\textsuperscript{\textdagger} & N/A\textsuperscript{\textdagger} & N/A\textsuperscript{\textdagger} & N/A\textsuperscript{\textdagger} & N/A\textsuperscript{\textdagger} & N/A\textsuperscript{\textdagger} \\
  MiMo-VL-7B-SFT\hspace{0.35em}\citep{MimoVL} & 28.57 & 4.83 & 31.82 & 4.20 & 19.29 & 7.81 & 52.78 & 12.73 & 40.11 & 20.75 & 24.30 & 6.74 & 23.21 & 6.83 \\
  MiMo-VL-7B-RL\hspace{0.35em}\citep{MimoVL} & 27.92 & 2.07 & 38.38 & 4.20 & 27.41 & 9.38 & 57.64 & 17.27 & 53.11 & 24.53 & 29.91 & 7.87 & 27.58 & 9.30 \\
  BAGEL\hspace{0.35em}\citep{BAGEL7BMoT} & N/A\textsuperscript{*} & N/A\textsuperscript{*} & N/A\textsuperscript{*} & N/A\textsuperscript{*} & N/A\textsuperscript{*} & N/A\textsuperscript{*} & N/A\textsuperscript{*} & N/A\textsuperscript{*} & N/A\textsuperscript{*} & N/A\textsuperscript{*} & N/A\textsuperscript{*} & N/A\textsuperscript{*} & N/A\textsuperscript{*} & N/A\textsuperscript{*} \\
  RynnBrain\hspace{0.35em}\citep{RynnBrain2B} & 44.81 & 7.59 & 35.35 & 5.59 & 15.23 & 7.81 & 40.28 & 13.64 & 54.80 & 18.87 & 42.06 & 11.24 & 27.07 & 2.47 \\
  JEDI\textsuperscript{*}\hspace{0.35em}\citep{JEDI3B} & 61.00 & 13.80 & 53.50 & 8.40 & 27.40 & 9.40 & 54.20 & 18.20 & 64.40 & 32.10 & 38.30 & 9.00 & 36.10 & -- \\
  UI-R1\textsuperscript{*}\hspace{0.35em}\citep{UIR13B} & 22.70 & 4.10 & 27.30 & 3.50 & 11.20 & 6.30 & 42.40 & 11.80 & 32.20 & 11.30 & 13.10 & 4.50 & 17.80 & -- \\
  UI-TARS\textsuperscript{*}\hspace{0.35em}\citep{UITARS2B} & 47.40 & 4.10 & 42.90 & 6.30 & 17.80 & 4.70 & 56.90 & 17.30 & 50.30 & 17.00 & 21.50 & 5.60 & 27.70 & -- \\
  GroundAnything-VLM & 75.97 & 55.86 & 72.73 & 58.04 & 55.84 & 48.44 & 86.11 & 61.82 & 74.58 & 62.26 & 62.62 & 48.31 & 65.34 & 0.06 \\
  GroundAnything & 77.92 & 70.34 & 85.86 & 66.43 & 71.57 & 71.88 & 90.97 & 81.82 & 83.62 & 66.04 & 64.49 & 60.67 & 75.96 & -- \\
  \addlinespace[2pt]
  \multicolumn{15}{@{}l}{\hspace{0.4em}\strut\textbf{Vision-Language Models (10B--1T)}} \\
  Qwen3.6-27B\hspace{0.35em}\citep{qwen3627b} & 87.01 & 53.79 & 76.77 & 44.76 & 75.13 & 46.88 & 81.25 & 48.18 & 87.57 & 64.15 & 79.44 & 57.30 & 69.64 & -- \\
  Qwen3-VL-32B\hspace{0.35em}\citep{qwen3vl} & 73.38 & 20.00 & 76.26 & 24.48 & 59.39 & 34.38 & 86.11 & 30.91 & 83.62 & 41.51 & 60.75 & 22.47 & 55.66 & 0.13 \\
  Qwen3.8-27B\hspace{0.35em}\citep{qwen38} & 85.71 & 64.83 & 45.45 & 41.96 & 50.76 & 14.06 & 67.36 & 36.36 & 87.57 & 56.60 & 74.77 & 47.19 & 58.76 & 0.76 \\
  Qwen3.5-35B-A3B\hspace{0.35em}\citep{qwen35} & 82.47 & 44.83 & 56.57 & 27.97 & 54.82 & 14.06 & 53.47 & 29.09 & 61.58 & 41.51 & 53.27 & 20.22 & 49.08 & 0.38 \\
  DeepSeek-VL2-Small-16B\hspace{0.35em}\citep{Deepseekvl2} & 0.00\textsuperscript{\textdagger} & 0.00\textsuperscript{\textdagger} & 0.00\textsuperscript{\textdagger} & 0.00\textsuperscript{\textdagger} & 0.00\textsuperscript{\textdagger} & 0.00\textsuperscript{\textdagger} & 0.00\textsuperscript{\textdagger} & 0.00\textsuperscript{\textdagger} & 0.00\textsuperscript{\textdagger} & 1.89\textsuperscript{\textdagger} & 0.93\textsuperscript{\textdagger} & 0.00\textsuperscript{\textdagger} & 0.13\textsuperscript{\textdagger} & 25.62\textsuperscript{\textdagger} \\
  DeepSeek-VL2-27B\hspace{0.35em}\citep{Deepseekvl2} & 0.00\textsuperscript{\textdagger} & 0.00\textsuperscript{\textdagger} & 0.00\textsuperscript{\textdagger} & 0.00\textsuperscript{\textdagger} & 0.51\textsuperscript{\textdagger} & 0.00\textsuperscript{\textdagger} & 0.00\textsuperscript{\textdagger} & 0.00\textsuperscript{\textdagger} & 0.00\textsuperscript{\textdagger} & 0.00\textsuperscript{\textdagger} & 0.00\textsuperscript{\textdagger} & 0.00\textsuperscript{\textdagger} & 0.06\textsuperscript{\textdagger} & 34.28\textsuperscript{\textdagger} \\
  GUI-Owl\textsuperscript{*}\hspace{0.35em}\citep{GUIOwl32B} & 84.40 & 39.30 & 65.20 & 18.20 & 62.40 & 28.10 & 82.60 & 39.10 & 81.40 & 39.60 & 70.10 & 36.00 & 58.00 & -- \\
  \addlinespace[2pt]
  \multicolumn{15}{@{}l}{\hspace{0.4em}\strut\textbf{Vision-Language Models (>1T)}} \\
  Qwen3.7-Max\hspace{0.35em}\citep{qwen37} & 79.22 & 37.50 & 62.63 & 34.27 & 46.19 & 31.25 & 77.08 & 31.82 & 78.53 & 47.17 & 61.68 & 38.20 & 55.06 & 10.57 \\
  Kimi-K2.6 & -- & -- & -- & -- & -- & -- & -- & -- & -- & -- & -- & -- & 6.07\textsuperscript{\textdagger} & 8.22\textsuperscript{\textdagger} \\
  Kimi-K3\hspace{0.35em}\citep{KimiK3} & 29.87\textsuperscript{\textdagger} & 20.00\textsuperscript{\textdagger} & 43.94\textsuperscript{\textdagger} & 25.17\textsuperscript{\textdagger} & 25.38\textsuperscript{\textdagger} & 3.12\textsuperscript{\textdagger} & 29.86\textsuperscript{\textdagger} & 12.73\textsuperscript{\textdagger} & 27.12\textsuperscript{\textdagger} & 13.21\textsuperscript{\textdagger} & 28.97\textsuperscript{\textdagger} & 8.99\textsuperscript{\textdagger} & 25.36\textsuperscript{\textdagger} & 3.73\textsuperscript{\textdagger} \\
  GPT-6 Astra & \textbf{96.10} & \textbf{84.83} & \textbf{95.45} & \textbf{89.51} & \textbf{95.43} & \textbf{85.94} & \textbf{94.44} & \textbf{87.27} & \textbf{98.87} & \textbf{96.23} & \textbf{97.20} & \textbf{89.89} & \textbf{93.17} & \textbf{0.00} \\
  \addlinespace[2pt]
  \bottomrule
  
  \end{NiceTabular}%
  }
  \endgroup
\end{table}

\paragraph{ScreenSpot-V2 and OSWorld-G.}
GroundAnything reaches 95.60\% ScreenSpot-V2 accuracy and 81.21\% OSWorld-G exact accuracy, improving over GroundAnything-VLM by 0.71 and 6.56 pp.
The smaller gain on ScreenSpot-V2 reflects a setting where the AR baseline already scores 94.89\%; OSWorld-G leaves more room for improvement.
GroundAnything does not improve every individual text/icon subset, and GPT-6 Astra still leads both overall metrics.
Unreported DLM parse-error rates remain unavailable.
\begin{table}[htbp]
  \centering
  \BenchmarkTableFont
  \caption{\textbf{ScreenSpot-V2 and OSWorld-G.} ScreenSpot-V2 includes text/icon results for mobile, desktop, and web environments, overall action accuracy, and parse-error rate; OSWorld-G reports exact accuracy and parse-error rate. GroundingDINO is N/A because GUI grounding is unsupported. BAGEL is N/A because its outputs do not satisfy the unified protocol. The starred ScreenSpot-V2 scores are from Table~8 of \citet{rexomni}.}
  \label{tab:app-gui-v2-osworld}
  \begingroup
  \fontsize{8}{9.7}\selectfont
  \setlength{\tabcolsep}{2.5pt}
  \renewcommand{\arraystretch}{1.15}
  \resizebox{\linewidth}{!}{%
  \begin{NiceTabular}{@{}>{\raggedright\arraybackslash}p{177pt}cccccccccc@{}}
  \CodeBefore
    \rowcolor{benchmarktype}{3}
    \rowcolor{benchmarktype}{5}
    \rowcolor{benchmarkpurple}{25}
    \rowcolor{benchmarkgreen}{26}
    \rowcolor{benchmarktype}{27}
    \rowcolor{benchmarktype}{34}
  \Body
  \toprule
  \textbf{Model} & \multicolumn{8}{c}{\textbf{ScreenSpot-V2}} & \multicolumn{2}{c}{\textbf{OSWorld-G}} \\
  \cmidrule(lr){2-9}\cmidrule(lr){10-11}
   & \BenchHead{Mobile\\text} & \BenchHead{Mobile\\icon} & \BenchHead{Desktop\\text} & \BenchHead{Desktop\\icon} & \BenchHead{Web\\text} & \BenchHead{Web\\icon} & \BenchHead{Action\\acc.} & \BenchHead{Parse\\err.} & \BenchHead{Exact\\acc.} & \BenchHead{Parse\\err.} \\
  \midrule
  \multicolumn{11}{@{}l}{\hspace{0.4em}\strut\textbf{Open-set Specialized Detectors}} \\
  GroundingDINO\hspace{0.35em}\citep{groundingdino} & N/A\textsuperscript{*} & N/A\textsuperscript{*} & N/A\textsuperscript{*} & N/A\textsuperscript{*} & N/A\textsuperscript{*} & N/A\textsuperscript{*} & N/A\textsuperscript{*} & N/A\textsuperscript{*} & N/A\textsuperscript{*} & N/A\textsuperscript{*} \\
  \addlinespace[2pt]
  \multicolumn{11}{@{}l}{\hspace{0.4em}\strut\textbf{Vision-Language Models (<10B)}} \\
  Rex-Omni\hspace{0.35em}\citep{rexomni} & 96.90 & 82.46 & 97.94 & 80.71 & 89.74 & 76.35 & 88.29 & -- & 46.10 & -- \\
  LocateAnything Fast\hspace{0.35em}\citep{locateanything} & 94.48 & 81.04 & 93.81 & 86.43 & 88.89 & 83.25 & 88.44 & -- & 59.93 & -- \\
  LocateAnything Hybrid\hspace{0.35em}\citep{locateanything} & 96.21 & 83.89 & 92.78 & 89.29 & 88.89 & 86.21 & 89.94 & -- & 60.46 & -- \\
  LocateAnything Slow NTP\hspace{0.35em}\citep{locateanything} & 95.86 & 83.41 & 93.30 & 89.29 & 90.60 & 85.22 & 90.02 & -- & 61.17 & -- \\
  Qwen2.5-VL-7B\hspace{0.35em}\citep{qwen25vl} & 98.28 & 82.94 & 91.75 & 67.86 & 92.74 & 79.31 & 87.34 & -- & 34.22 & -- \\
  Qwen3-VL-2B\hspace{0.35em}\citep{qwen3vl} & 93.10 & 70.62 & 79.90 & 61.43 & 79.91 & 62.07 & 76.49 & -- & 33.51 & -- \\
  Qwen3-VL-4B\hspace{0.35em}\citep{qwen3vl} & 97.93 & 87.20 & 96.91 & 87.14 & 94.44 & 86.21 & 92.30 & -- & 56.91 & -- \\
  Qwen3-VL-8B\hspace{0.35em}\citep{qwen3vl} & 98.62 & 89.57 & 98.45 & 87.86 & 94.87 & 89.16 & 93.71 & -- & 56.91 & -- \\
  Qwen3.5-4B\hspace{0.35em}\citep{qwen35} & \textbf{98.97} & 88.15 & 97.94 & 89.29 & 95.73 & 90.15 & 93.95 & -- & 54.61 & -- \\
  Qwen3.5-9B\hspace{0.35em}\citep{qwen35} & 96.55 & 88.15 & 97.94 & 93.57 & 87.61 & 78.82 & 90.57 & -- & 60.99 & -- \\
  RynnBrain1.1\hspace{0.35em}\citep{RynnBrain112B} & 82.07 & 71.09 & 75.26 & 52.86 & 73.08 & 57.64 & 70.44 & 4.25 & 33.33 & 4.96 \\
  SenseNova-Vision\hspace{0.35em}\citep{SenseNovaVision7BMoT} & N/A\textsuperscript{\textdagger} & N/A\textsuperscript{\textdagger} & N/A\textsuperscript{\textdagger} & N/A\textsuperscript{\textdagger} & N/A\textsuperscript{\textdagger} & N/A\textsuperscript{\textdagger} & N/A\textsuperscript{\textdagger} & N/A\textsuperscript{\textdagger} & N/A\textsuperscript{\textdagger} & N/A\textsuperscript{\textdagger} \\
  MiMo-VL-7B-SFT\hspace{0.35em}\citep{MimoVL} & 91.03 & 68.25 & 86.60 & 50.00 & 63.68 & 56.65 & 71.54 & 8.18 & 34.04 & 6.38 \\
  MiMo-VL-7B-RL\hspace{0.35em}\citep{MimoVL} & 90.69 & 67.30 & 85.57 & 49.29 & 78.21 & 62.56 & 74.69 & 5.66 & 37.06 & 2.30 \\
  BAGEL\hspace{0.35em}\citep{BAGEL7BMoT} & N/A\textsuperscript{*} & N/A\textsuperscript{*} & N/A\textsuperscript{*} & N/A\textsuperscript{*} & N/A\textsuperscript{*} & N/A\textsuperscript{*} & N/A\textsuperscript{*} & N/A\textsuperscript{*} & N/A\textsuperscript{*} & N/A\textsuperscript{*} \\
  RynnBrain\hspace{0.35em}\citep{RynnBrain2B} & 89.31 & 69.19 & 80.93 & 56.43 & 79.91 & 60.10 & 74.69 & 5.27 & 35.99 & 1.77 \\
  JEDI\textsuperscript{*}\hspace{0.35em}\citep{JEDI3B} & 96.60 & 81.50 & 96.90 & 78.60 & 88.50 & 83.70 & 88.60 & -- & -- & -- \\
  UI-R1\textsuperscript{*}\hspace{0.35em}\citep{UIR13B} & 84.30 & \textbf{96.20} & 75.40 & 89.20 & 63.60 & 92.30 & 85.40 & -- & -- & -- \\
  UI-TARS\textsuperscript{*}\hspace{0.35em}\citep{UITARS2B} & 95.20 & 79.10 & 90.70 & 68.60 & 87.20 & 78.30 & 84.70 & -- & -- & -- \\
  GroundAnything-VLM & 97.24 & 89.10 & 98.45 & 94.29 & 95.73 & 93.60 & 94.89 & \textbf{0.00} & 74.65 & \textbf{0.00} \\
  GroundAnything & 98.62 & 91.00 & 97.94 & 95.00 & 94.87 & 95.07 & 95.60 & -- & 81.21 & -- \\
  \addlinespace[2pt]
  \multicolumn{11}{@{}l}{\hspace{0.4em}\strut\textbf{Vision-Language Models (10B--1T)}} \\
  Qwen3.6-27B\hspace{0.35em}\citep{qwen3627b} & \textbf{98.97} & 92.42 & 98.45 & 92.86 & 96.58 & 89.16 & 95.13 & -- & 68.44 & -- \\
  Qwen3-VL-32B\hspace{0.35em}\citep{qwen3vl} & 98.62 & 90.05 & 98.45 & 86.43 & 96.58 & 91.63 & 94.34 & 0.08 & 60.28 & 0.71 \\
  Qwen3.8-27B\hspace{0.35em}\citep{qwen38} & 97.93 & 90.52 & 98.45 & 94.29 & 94.87 & 89.66 & 94.50 & 0.24 & 63.48 & \textbf{0.00} \\
  Qwen3.5-35B-A3B\hspace{0.35em}\citep{qwen35} & 97.59 & 89.10 & 96.39 & 92.14 & 94.02 & 86.70 & 93.00 & 1.42 & 62.23 & 0.18 \\
  DeepSeek-VL2-Small-16B\hspace{0.35em}\citep{Deepseekvl2} & 2.07\textsuperscript{\textdagger} & 0.00\textsuperscript{\textdagger} & 4.64\textsuperscript{\textdagger} & 1.43\textsuperscript{\textdagger} & 0.85\textsuperscript{\textdagger} & 0.49\textsuperscript{\textdagger} & 1.57\textsuperscript{\textdagger} & 41.82\textsuperscript{\textdagger} & 0.89\textsuperscript{\textdagger} & 11.52\textsuperscript{\textdagger} \\
  DeepSeek-VL2-27B\hspace{0.35em}\citep{Deepseekvl2} & 6.90\textsuperscript{\textdagger} & 1.42\textsuperscript{\textdagger} & 2.58\textsuperscript{\textdagger} & 0.00\textsuperscript{\textdagger} & 2.56\textsuperscript{\textdagger} & 1.48\textsuperscript{\textdagger} & 2.91\textsuperscript{\textdagger} & 25.63\textsuperscript{\textdagger} & 0.00\textsuperscript{\textdagger} & 32.62\textsuperscript{\textdagger} \\
  \addlinespace[2pt]
  \multicolumn{11}{@{}l}{\hspace{0.4em}\strut\textbf{Vision-Language Models (>1T)}} \\
  Qwen3.7-Max\hspace{0.35em}\citep{qwen37} & 89.31 & 82.94 & 71.13 & 76.43 & 81.62 & 79.80 & 81.13 & 13.21 & 49.29 & 24.47 \\
  Kimi-K2.6 & -- & -- & -- & -- & -- & -- & 52.36\textsuperscript{\textdagger} & 7.15\textsuperscript{\textdagger} & 10.11\textsuperscript{\textdagger} & 7.98\textsuperscript{\textdagger} \\
  Kimi-K3\hspace{0.35em}\citep{KimiK3} & 95.86\textsuperscript{\textdagger} & 88.63\textsuperscript{\textdagger} & 97.42\textsuperscript{\textdagger} & 89.29\textsuperscript{\textdagger} & 63.68\textsuperscript{\textdagger} & 61.08\textsuperscript{\textdagger} & 82.70\textsuperscript{\textdagger} & 1.57\textsuperscript{\textdagger} & 68.26\textsuperscript{\textdagger} & 4.96\textsuperscript{\textdagger} \\
  GPT-6 Astra & \textbf{98.97} & 95.26 & \textbf{98.97} & \textbf{98.57} & \textbf{98.29} & \textbf{97.04} & \textbf{97.88} & 0.24 & \textbf{86.70} & 0.18 \\
  \addlinespace[2pt]
  \bottomrule
  
  \end{NiceTabular}%
  }
  \endgroup
\end{table}

\subsection{Layout Grounding}
\label{app:layout_grounding}

Layout grounding predicts labeled document regions and uses the box evaluation convention of \citet{rexomni}.

\paragraph{DocLayNet.}
GroundAnything-VLM reaches 85.78 F1mIoU, slightly above SenseNova-Vision (85.53) and above the external DocLayout-YOLO reference (81.10).
GroundAnything scores 68.38, close to Rex-Omni (68.06) but below LocateAnything Hybrid (77.34).
Its 37.78 F1 at IoU 0.95 nevertheless exceeds GPT-6 Astra's 34.93, illustrating that strict boundary accuracy and overall region detection can rank models differently.
\begin{table}[htbp]
  \centering
  \BenchmarkTableFont
  \caption{\textbf{DocLayNet.} Complete box-grounding metrics. GroundingDINO lacks a compatible document-region interface; it, Kimi-K3, and both DeepSeek variants are N/A. Starred MiMo and BAGEL scores retain observations under suspected output-protocol incompatibility and are not formal capability measurements. The starred DocLayout-YOLO and SEED1.5-VL scores are from Table~9 of \citet{rexomni}.}
  \label{tab:app-doclaynet}
  \begingroup
  \fontsize{8}{9.7}\selectfont
  \setlength{\tabcolsep}{3pt}
  \renewcommand{\arraystretch}{1.15}
  \resizebox{\linewidth}{!}{%
  \begin{NiceTabular}{@{}>{\raggedright\arraybackslash}p{177pt}ccccccccc@{}}
  \CodeBefore
    \rowcolor{benchmarktype}{3}
    \rowcolor{benchmarktype}{5}
    \rowcolor{benchmarktype}{7}
    \rowcolor{benchmarkpurple}{24}
    \rowcolor{benchmarkgreen}{25}
    \rowcolor{benchmarktype}{26}
    \rowcolor{benchmarktype}{34}
  \Body
  \toprule
  \textbf{Model} & \multicolumn{3}{c}{\textbf{IoU 0.50}} & \multicolumn{3}{c}{\textbf{IoU 0.95}} & \multicolumn{3}{c}{\textbf{mIoU}} \\
  \cmidrule(lr){2-4}\cmidrule(lr){5-7}\cmidrule(lr){8-10}
   & \BenchHead{R} & \BenchHead{P} & \BenchHead{F1} & \BenchHead{R} & \BenchHead{P} & \BenchHead{F1} & \BenchHead{R} & \BenchHead{P} & \BenchHead{F1} \\
  \midrule
  \multicolumn{10}{@{}l}{\hspace{0.4em}\strut\textbf{Closed-set Specialized Detectors}} \\
  DocLayout-YOLO\textsuperscript{*}\hspace{0.35em}\citep{DocLayoutYOLO} & -- & -- & 91.20 & -- & -- & \textbf{52.10} & -- & -- & 81.10 \\
  \addlinespace[2pt]
  \multicolumn{10}{@{}l}{\hspace{0.4em}\strut\textbf{Open-set Specialized Detectors}} \\
  GroundingDINO\hspace{0.35em}\citep{groundingdino} & N/A\textsuperscript{*} & N/A\textsuperscript{*} & N/A\textsuperscript{*} & N/A\textsuperscript{*} & N/A\textsuperscript{*} & N/A\textsuperscript{*} & N/A\textsuperscript{*} & N/A\textsuperscript{*} & N/A\textsuperscript{*} \\
  \addlinespace[2pt]
  \multicolumn{10}{@{}l}{\hspace{0.4em}\strut\textbf{Vision-Language Models (<10B)}} \\
  Rex-Omni\hspace{0.35em}\citep{rexomni} & 83.48 & 88.85 & 86.08 & 27.13 & 27.95 & 27.53 & 66.25 & 69.98 & 68.06 \\
  LocateAnything Fast\hspace{0.35em}\citep{locateanything} & 55.10 & 59.26 & 57.11 & 25.03 & 26.55 & 25.77 & 47.73 & 51.12 & 49.37 \\
  LocateAnything Hybrid\hspace{0.35em}\citep{locateanything} & 87.95 & 90.90 & 89.40 & 38.72 & 39.82 & 39.26 & 76.14 & 78.57 & 77.34 \\
  LocateAnything Slow NTP\hspace{0.35em}\citep{locateanything} & 91.55 & 94.58 & 93.04 & 39.56 & 40.64 & 40.09 & 78.98 & 81.45 & 80.19 \\
  Qwen2.5-VL-7B\hspace{0.35em}\citep{qwen25vl} & 30.23 & 29.07 & 29.64 & 2.52 & 2.92 & 2.70 & 16.41 & 16.72 & 16.55 \\
  Qwen3-VL-2B\hspace{0.35em}\citep{qwen3vl} & 40.38 & 35.70 & 37.89 & 2.80 & 2.83 & 2.81 & 21.10 & 19.19 & 20.09 \\
  Qwen3-VL-4B\hspace{0.35em}\citep{qwen3vl} & 64.86 & 69.50 & 67.10 & 8.40 & 8.68 & 8.54 & 39.90 & 41.76 & 40.81 \\
  Qwen3-VL-8B\hspace{0.35em}\citep{qwen3vl} & 60.09 & 57.80 & 58.92 & 7.13 & 7.18 & 7.15 & 37.76 & 36.58 & 37.16 \\
  Qwen3.5-4B\hspace{0.35em}\citep{qwen35} & 59.37 & 47.02 & 52.48 & 5.04 & 4.47 & 4.74 & 35.03 & 28.51 & 31.43 \\
  Qwen3.5-9B\hspace{0.35em}\citep{qwen35} & 63.44 & 49.12 & 55.37 & 6.96 & 6.09 & 6.50 & 39.12 & 31.10 & 34.65 \\
  RynnBrain1.1\hspace{0.35em}\citep{RynnBrain112B} & 5.50 & 24.01 & 8.96 & 1.52 & 6.68 & 2.48 & 3.70 & 16.21 & 6.02 \\
  SenseNova-Vision\hspace{0.35em}\citep{SenseNovaVision7BMoT} & 95.45 & 95.78 & 95.61 & \textbf{49.23} & \textbf{49.43} & 49.33 & 85.38 & 85.67 & 85.53 \\
  MiMo-VL-7B-SFT\hspace{0.35em}\citep{MimoVL} & 3.37\textsuperscript{*} & 5.01\textsuperscript{*} & 4.03\textsuperscript{*} & 0.17\textsuperscript{*} & 0.24\textsuperscript{*} & 0.20\textsuperscript{*} & 1.21\textsuperscript{*} & 1.99\textsuperscript{*} & 1.50\textsuperscript{*} \\
  MiMo-VL-7B-RL\hspace{0.35em}\citep{MimoVL} & 5.43\textsuperscript{*} & 8.01\textsuperscript{*} & 6.47\textsuperscript{*} & 0.11\textsuperscript{*} & 0.19\textsuperscript{*} & 0.14\textsuperscript{*} & 1.98\textsuperscript{*} & 3.15\textsuperscript{*} & 2.43\textsuperscript{*} \\
  BAGEL\hspace{0.35em}\citep{BAGEL7BMoT} & 10.83\textsuperscript{*} & 10.93\textsuperscript{*} & 10.88\textsuperscript{*} & 0.60\textsuperscript{*} & 0.69\textsuperscript{*} & 0.64\textsuperscript{*} & 4.34\textsuperscript{*} & 4.64\textsuperscript{*} & 4.48\textsuperscript{*} \\
  RynnBrain\hspace{0.35em}\citep{RynnBrain2B} & 7.16 & 17.74 & 10.20 & 0.98 & 1.88 & 1.29 & 4.01 & 9.40 & 5.62 \\
  GroundAnything-VLM & 96.29 & \textbf{96.22} & 96.26 & 44.80 & 44.75 & 44.77 & \textbf{85.83} & \textbf{85.73} & \textbf{85.78} \\
  GroundAnything & 78.77 & 72.76 & 75.65 & 38.07 & 37.50 & 37.78 & 71.25 & 65.74 & 68.38 \\
  \addlinespace[2pt]
  \multicolumn{10}{@{}l}{\hspace{0.4em}\strut\textbf{Vision-Language Models (10B--1T)}} \\
  Qwen3.6-27B\hspace{0.35em}\citep{qwen3627b} & 73.85 & 72.14 & 72.98 & 10.44 & 10.10 & 10.27 & 47.15 & 46.10 & 46.62 \\
  Qwen3-VL-32B\hspace{0.35em}\citep{qwen3vl} & 40.07 & 36.22 & 38.05 & 4.08 & 4.22 & 4.15 & 21.49 & 20.37 & 20.91 \\
  Qwen3.8-27B\hspace{0.35em}\citep{qwen38} & 69.25 & 70.21 & 69.73 & 8.62 & 8.74 & 8.68 & 41.97 & 42.46 & 42.21 \\
  Qwen3.5-35B-A3B\hspace{0.35em}\citep{qwen35} & 63.83 & 56.73 & 60.07 & 7.40 & 6.99 & 7.19 & 40.02 & 36.03 & 37.92 \\
  DeepSeek-VL2-Small-16B\hspace{0.35em}\citep{Deepseekvl2} & N/A\textsuperscript{*} & N/A\textsuperscript{*} & N/A\textsuperscript{*} & N/A\textsuperscript{*} & N/A\textsuperscript{*} & N/A\textsuperscript{*} & N/A\textsuperscript{*} & N/A\textsuperscript{*} & N/A\textsuperscript{*} \\
  DeepSeek-VL2-27B\hspace{0.35em}\citep{Deepseekvl2} & N/A\textsuperscript{*} & N/A\textsuperscript{*} & N/A\textsuperscript{*} & N/A\textsuperscript{*} & N/A\textsuperscript{*} & N/A\textsuperscript{*} & N/A\textsuperscript{*} & N/A\textsuperscript{*} & N/A\textsuperscript{*} \\
  SEED1.5-VL\textsuperscript{*}\hspace{0.35em}\citep{SEED15VL} & -- & -- & 54.90 & -- & -- & 4.30 & -- & -- & 28.70 \\
  \addlinespace[2pt]
  \multicolumn{10}{@{}l}{\hspace{0.4em}\strut\textbf{Vision-Language Models (>1T)}} \\
  Qwen3.7-Max\hspace{0.35em}\citep{qwen37} & 75.85 & 65.60 & 70.35 & 12.76 & 12.08 & 12.41 & 49.90 & 44.31 & 46.93 \\
  Kimi-K2.6 & 29.23\textsuperscript{\textdagger} & 28.21\textsuperscript{\textdagger} & 28.71\textsuperscript{\textdagger} & 2.29\textsuperscript{\textdagger} & 2.43\textsuperscript{\textdagger} & 2.36\textsuperscript{\textdagger} & 15.64\textsuperscript{\textdagger} & 15.55\textsuperscript{\textdagger} & 15.59\textsuperscript{\textdagger} \\
  Kimi-K3\hspace{0.35em}\citep{KimiK3} & N/A\textsuperscript{*} & N/A\textsuperscript{*} & N/A\textsuperscript{*} & N/A\textsuperscript{*} & N/A\textsuperscript{*} & N/A\textsuperscript{*} & N/A\textsuperscript{*} & N/A\textsuperscript{*} & N/A\textsuperscript{*} \\
  GPT-6 Astra & \textbf{97.45} & 95.72 & \textbf{96.58} & 35.04 & 34.83 & 34.93 & 78.24 & 76.86 & 77.54 \\
  \addlinespace[2pt]
  \bottomrule
  
  \end{NiceTabular}%
  }
  \endgroup
\end{table}

\paragraph{M6Doc.}
GroundAnything-VLM obtains 74.76 F1mIoU, exceeding LocateAnything Slow NTP (68.35).
GroundAnything reaches 56.69, above Rex-Omni (54.95) but below LocateAnything Hybrid (65.94).
Together with DocLayNet, these results establish coverage of document-region grounding while identifying a substantial diffusion--AR gap in structured layout extraction.
\begin{table}[htbp]
  \centering
  \BenchmarkTableFont
  \caption{\textbf{M6Doc.} Complete box-grounding metrics. GroundingDINO lacks a compatible document-region interface; it, Kimi-K3, and both DeepSeek variants are N/A. Starred MiMo and BAGEL scores retain observations under suspected output-protocol incompatibility and are not formal capability measurements. The starred SEED1.5-VL scores are from Table~9 of \citet{rexomni}.}
  \label{tab:app-m6doc}
  \begingroup
  \fontsize{8}{9.7}\selectfont
  \setlength{\tabcolsep}{3pt}
  \renewcommand{\arraystretch}{1.15}
  \resizebox{\linewidth}{!}{%
  \begin{NiceTabular}{@{}>{\raggedright\arraybackslash}p{177pt}ccccccccc@{}}
  \CodeBefore
    \rowcolor{benchmarktype}{3}
    \rowcolor{benchmarktype}{5}
    \rowcolor{benchmarkpurple}{22}
    \rowcolor{benchmarkgreen}{23}
    \rowcolor{benchmarktype}{24}
    \rowcolor{benchmarktype}{32}
  \Body
  \toprule
  \textbf{Model} & \multicolumn{3}{c}{\textbf{IoU 0.50}} & \multicolumn{3}{c}{\textbf{IoU 0.95}} & \multicolumn{3}{c}{\textbf{mIoU}} \\
  \cmidrule(lr){2-4}\cmidrule(lr){5-7}\cmidrule(lr){8-10}
   & \BenchHead{R} & \BenchHead{P} & \BenchHead{F1} & \BenchHead{R} & \BenchHead{P} & \BenchHead{F1} & \BenchHead{R} & \BenchHead{P} & \BenchHead{F1} \\
  \midrule
  \multicolumn{10}{@{}l}{\hspace{0.4em}\strut\textbf{Open-set Specialized Detectors}} \\
  GroundingDINO\hspace{0.35em}\citep{groundingdino} & N/A\textsuperscript{*} & N/A\textsuperscript{*} & N/A\textsuperscript{*} & N/A\textsuperscript{*} & N/A\textsuperscript{*} & N/A\textsuperscript{*} & N/A\textsuperscript{*} & N/A\textsuperscript{*} & N/A\textsuperscript{*} \\
  \addlinespace[2pt]
  \multicolumn{10}{@{}l}{\hspace{0.4em}\strut\textbf{Vision-Language Models (<10B)}} \\
  Rex-Omni\hspace{0.35em}\citep{rexomni} & 73.09 & 78.27 & 75.59 & 18.16 & 18.78 & 18.46 & 53.36 & 56.64 & 54.95 \\
  LocateAnything Fast\hspace{0.35em}\citep{locateanything} & 59.82 & 61.54 & 60.67 & 18.06 & 18.50 & 18.28 & 46.39 & 47.69 & 47.03 \\
  LocateAnything Hybrid\hspace{0.35em}\citep{locateanything} & 83.54 & 84.94 & 84.23 & 26.18 & 26.59 & 26.39 & 65.38 & 66.50 & 65.94 \\
  LocateAnything Slow NTP\hspace{0.35em}\citep{locateanything} & 88.16 & 89.64 & 88.89 & 25.62 & 26.08 & 25.85 & 67.78 & 68.94 & 68.35 \\
  Qwen2.5-VL-7B\hspace{0.35em}\citep{qwen25vl} & 19.20 & 22.91 & 20.89 & 2.24 & 2.65 & 2.43 & 11.86 & 14.26 & 12.95 \\
  Qwen3-VL-2B\hspace{0.35em}\citep{qwen3vl} & 22.31 & 24.68 & 23.44 & 2.24 & 2.58 & 2.40 & 12.61 & 14.08 & 13.31 \\
  Qwen3-VL-4B\hspace{0.35em}\citep{qwen3vl} & 37.94 & 44.13 & 40.80 & 4.88 & 5.62 & 5.22 & 23.03 & 26.69 & 24.73 \\
  Qwen3-VL-8B\hspace{0.35em}\citep{qwen3vl} & 35.84 & 36.64 & 36.23 & 4.73 & 5.01 & 4.87 & 21.56 & 22.36 & 21.95 \\
  Qwen3.5-4B\hspace{0.35em}\citep{qwen35} & 17.32 & 15.67 & 16.45 & 1.47 & 1.54 & 1.50 & 8.44 & 8.13 & 8.28 \\
  Qwen3.5-9B\hspace{0.35em}\citep{qwen35} & 30.92 & 28.08 & 29.43 & 4.08 & 4.17 & 4.12 & 17.80 & 16.94 & 17.35 \\
  RynnBrain1.1\hspace{0.35em}\citep{RynnBrain112B} & 4.11 & 26.77 & 7.13 & 0.89 & 6.20 & 1.56 & 2.60 & 17.51 & 4.53 \\
  SenseNova-Vision\hspace{0.35em}\citep{SenseNovaVision7BMoT} & 50.01 & 63.15 & 55.82 & 8.87 & 10.60 & 9.66 & 32.12 & 39.99 & 35.62 \\
  MiMo-VL-7B-SFT\hspace{0.35em}\citep{MimoVL} & 2.88\textsuperscript{*} & 5.77\textsuperscript{*} & 3.84\textsuperscript{*} & 0.03\textsuperscript{*} & 0.10\textsuperscript{*} & 0.05\textsuperscript{*} & 0.91\textsuperscript{*} & 2.14\textsuperscript{*} & 1.28\textsuperscript{*} \\
  MiMo-VL-7B-RL\hspace{0.35em}\citep{MimoVL} & 4.63\textsuperscript{*} & 9.36\textsuperscript{*} & 6.19\textsuperscript{*} & 0.07\textsuperscript{*} & 0.12\textsuperscript{*} & 0.09\textsuperscript{*} & 1.53\textsuperscript{*} & 3.24\textsuperscript{*} & 2.07\textsuperscript{*} \\
  BAGEL\hspace{0.35em}\citep{BAGEL7BMoT} & 10.62\textsuperscript{*} & 15.75\textsuperscript{*} & 12.68\textsuperscript{*} & 0.40\textsuperscript{*} & 0.62\textsuperscript{*} & 0.48\textsuperscript{*} & 4.94\textsuperscript{*} & 7.61\textsuperscript{*} & 5.99\textsuperscript{*} \\
  RynnBrain\hspace{0.35em}\citep{RynnBrain2B} & 3.98 & 21.54 & 6.72 & 0.32 & 2.06 & 0.55 & 2.00 & 10.88 & 3.38 \\
  GroundAnything-VLM & \textbf{91.91} & \textbf{93.15} & \textbf{92.53} & \textbf{31.57} & \textbf{31.90} & \textbf{31.74} & \textbf{74.29} & \textbf{75.24} & \textbf{74.76} \\
  GroundAnything & 69.65 & 73.29 & 71.42 & 25.96 & 24.09 & 24.99 & 55.91 & 57.49 & 56.69 \\
  \addlinespace[2pt]
  \multicolumn{10}{@{}l}{\hspace{0.4em}\strut\textbf{Vision-Language Models (10B--1T)}} \\
  Qwen3.6-27B\hspace{0.35em}\citep{qwen3627b} & 45.19 & 47.32 & 46.23 & 6.69 & 6.91 & 6.80 & 27.92 & 29.17 & 28.53 \\
  Qwen3-VL-32B\hspace{0.35em}\citep{qwen3vl} & 46.89 & 48.85 & 47.85 & 7.02 & 7.37 & 7.19 & 29.66 & 31.02 & 30.32 \\
  Qwen3.8-27B\hspace{0.35em}\citep{qwen38} & 34.42 & 35.88 & 35.13 & 4.19 & 4.47 & 4.33 & 19.43 & 20.43 & 19.92 \\
  Qwen3.5-35B-A3B\hspace{0.35em}\citep{qwen35} & 44.90 & 46.66 & 45.76 & 6.63 & 6.99 & 6.81 & 28.07 & 29.32 & 28.68 \\
  DeepSeek-VL2-Small-16B\hspace{0.35em}\citep{Deepseekvl2} & N/A\textsuperscript{*} & N/A\textsuperscript{*} & N/A\textsuperscript{*} & N/A\textsuperscript{*} & N/A\textsuperscript{*} & N/A\textsuperscript{*} & N/A\textsuperscript{*} & N/A\textsuperscript{*} & N/A\textsuperscript{*} \\
  DeepSeek-VL2-27B\hspace{0.35em}\citep{Deepseekvl2} & N/A\textsuperscript{*} & N/A\textsuperscript{*} & N/A\textsuperscript{*} & N/A\textsuperscript{*} & N/A\textsuperscript{*} & N/A\textsuperscript{*} & N/A\textsuperscript{*} & N/A\textsuperscript{*} & N/A\textsuperscript{*} \\
  SEED1.5-VL\textsuperscript{*}\hspace{0.35em}\citep{SEED15VL} & -- & -- & 48.00 & -- & -- & 3.40 & -- & -- & 28.00 \\
  \addlinespace[2pt]
  \multicolumn{10}{@{}l}{\hspace{0.4em}\strut\textbf{Vision-Language Models (>1T)}} \\
  Qwen3.7-Max\hspace{0.35em}\citep{qwen37} & 48.23 & 50.84 & 49.50 & 9.36 & 9.87 & 9.61 & 31.50 & 33.16 & 32.31 \\
  Kimi-K2.6 & 21.96\textsuperscript{\textdagger} & 23.87\textsuperscript{\textdagger} & 22.87\textsuperscript{\textdagger} & 2.80\textsuperscript{\textdagger} & 3.16\textsuperscript{\textdagger} & 2.97\textsuperscript{\textdagger} & 12.89\textsuperscript{\textdagger} & 14.16\textsuperscript{\textdagger} & 13.50\textsuperscript{\textdagger} \\
  Kimi-K3\hspace{0.35em}\citep{KimiK3} & N/A\textsuperscript{*} & N/A\textsuperscript{*} & N/A\textsuperscript{*} & N/A\textsuperscript{*} & N/A\textsuperscript{*} & N/A\textsuperscript{*} & N/A\textsuperscript{*} & N/A\textsuperscript{*} & N/A\textsuperscript{*} \\
  GPT-6 Astra & 78.30 & 83.51 & 80.84 & 19.68 & 21.54 & 20.58 & 58.50 & 62.78 & 60.59 \\
  \addlinespace[2pt]
  \bottomrule
  
  \end{NiceTabular}%
  }
  \endgroup
\end{table}

\subsection{Visual Prompting}
\label{app:visual_prompting}

\paragraph{FSC147.}
GroundAnything-VLM and GroundAnything achieve 60.29 and 52.43 F1mIoU, respectively, below SenseNova-Vision's 62.51.
The DLM variant has higher F1 at IoU 0.95 than its AR counterpart (9.37 versus 4.48), despite the lower aggregate.
This threshold dependence again cautions against equating a single strict-overlap score with overall grounding quality.
LocateAnything's N/A entries indicate an unavailable visual-prompt interface, not zero accuracy.
\begin{table}[htbp]
  \centering
  \BenchmarkTableFont
  \caption{\textbf{FSC147 visual prompting.} Complete box-grounding metrics. LocateAnything variants lack a supported visual-prompt interface; both DeepSeek variants have incompatible output protocols. Their entries are N/A. Starred GroundingDINO scores come from an unsupported visual-prompt task interface; starred MiMo and BAGEL scores are retained observations under suspected output-protocol incompatibility.}
  \label{tab:app-fsc147}
  \begingroup
  \fontsize{8}{9.7}\selectfont
  \setlength{\tabcolsep}{3pt}
  \renewcommand{\arraystretch}{1.15}
  \resizebox{\linewidth}{!}{%
  \begin{NiceTabular}{@{}>{\raggedright\arraybackslash}p{177pt}ccccccccc@{}}
  \CodeBefore
    \rowcolor{benchmarktype}{3}
    \rowcolor{benchmarktype}{5}
    \rowcolor{benchmarkpurple}{22}
    \rowcolor{benchmarkgreen}{23}
    \rowcolor{benchmarktype}{24}
    \rowcolor{benchmarktype}{31}
  \Body
  \toprule
  \textbf{Model} & \multicolumn{3}{c}{\textbf{IoU 0.50}} & \multicolumn{3}{c}{\textbf{IoU 0.95}} & \multicolumn{3}{c}{\textbf{mIoU}} \\
  \cmidrule(lr){2-4}\cmidrule(lr){5-7}\cmidrule(lr){8-10}
   & \BenchHead{R} & \BenchHead{P} & \BenchHead{F1} & \BenchHead{R} & \BenchHead{P} & \BenchHead{F1} & \BenchHead{R} & \BenchHead{P} & \BenchHead{F1} \\
  \midrule
  \multicolumn{10}{@{}l}{\hspace{0.4em}\strut\textbf{Open-set Specialized Detectors}} \\
  GroundingDINO\hspace{0.35em}\citep{groundingdino} & 7.94\textsuperscript{*} & 20.16\textsuperscript{*} & 11.39\textsuperscript{*} & 1.56\textsuperscript{*} & 4.94\textsuperscript{*} & 2.37\textsuperscript{*} & 6.34\textsuperscript{*} & 16.01\textsuperscript{*} & 9.08\textsuperscript{*} \\
  \addlinespace[2pt]
  \multicolumn{10}{@{}l}{\hspace{0.4em}\strut\textbf{Vision-Language Models (<10B)}} \\
  Rex-Omni\hspace{0.35em}\citep{rexomni} & 78.97 & 77.92 & 78.44 & 9.13 & 9.01 & 9.07 & 57.56 & 56.75 & 57.15 \\
  LocateAnything Fast\hspace{0.35em}\citep{locateanything} & N/A\textsuperscript{*} & N/A\textsuperscript{*} & N/A\textsuperscript{*} & N/A\textsuperscript{*} & N/A\textsuperscript{*} & N/A\textsuperscript{*} & N/A\textsuperscript{*} & N/A\textsuperscript{*} & N/A\textsuperscript{*} \\
  LocateAnything Hybrid\hspace{0.35em}\citep{locateanything} & N/A\textsuperscript{*} & N/A\textsuperscript{*} & N/A\textsuperscript{*} & N/A\textsuperscript{*} & N/A\textsuperscript{*} & N/A\textsuperscript{*} & N/A\textsuperscript{*} & N/A\textsuperscript{*} & N/A\textsuperscript{*} \\
  LocateAnything Slow NTP\hspace{0.35em}\citep{locateanything} & N/A\textsuperscript{*} & N/A\textsuperscript{*} & N/A\textsuperscript{*} & N/A\textsuperscript{*} & N/A\textsuperscript{*} & N/A\textsuperscript{*} & N/A\textsuperscript{*} & N/A\textsuperscript{*} & N/A\textsuperscript{*} \\
  Qwen2.5-VL-7B\hspace{0.35em}\citep{qwen25vl} & 14.70 & 34.29 & 20.57 & 0.09 & 1.20 & 0.16 & 7.54 & 17.39 & 10.49 \\
  Qwen3-VL-2B\hspace{0.35em}\citep{qwen3vl} & 22.55 & 39.85 & 28.80 & 0.30 & 0.73 & 0.42 & 11.21 & 19.07 & 14.11 \\
  Qwen3-VL-4B\hspace{0.35em}\citep{qwen3vl} & 18.74 & 83.98 & 30.65 & 0.59 & 1.60 & 0.86 & 11.60 & 51.16 & 18.90 \\
  Qwen3-VL-8B\hspace{0.35em}\citep{qwen3vl} & 19.80 & 74.93 & 31.33 & 0.39 & 0.85 & 0.53 & 11.09 & 40.67 & 17.41 \\
  Qwen3.5-4B\hspace{0.35em}\citep{qwen35} & 27.06 & 41.98 & 32.91 & 0.14 & 0.21 & 0.17 & 11.70 & 18.62 & 14.37 \\
  Qwen3.5-9B\hspace{0.35em}\citep{qwen35} & 29.37 & 60.29 & 39.49 & 0.64 & 1.21 & 0.84 & 15.31 & 31.30 & 20.56 \\
  RynnBrain1.1\hspace{0.35em}\citep{RynnBrain112B} & 5.25 & 47.70 & 9.47 & 0.02 & 0.14 & 0.03 & 2.68 & 22.33 & 4.78 \\
  SenseNova-Vision\hspace{0.35em}\citep{SenseNovaVision7BMoT} & 78.56 & 88.88 & 83.40 & \textbf{10.25} & \textbf{10.78} & \textbf{10.51} & 59.46 & \textbf{65.89} & \textbf{62.51} \\
  MiMo-VL-7B-SFT\hspace{0.35em}\citep{MimoVL} & 13.59\textsuperscript{*} & 16.68\textsuperscript{*} & 14.98\textsuperscript{*} & 0.04\textsuperscript{*} & 0.04\textsuperscript{*} & 0.04\textsuperscript{*} & 4.65\textsuperscript{*} & 5.79\textsuperscript{*} & 5.15\textsuperscript{*} \\
  MiMo-VL-7B-RL\hspace{0.35em}\citep{MimoVL} & 13.58\textsuperscript{*} & 21.88\textsuperscript{*} & 16.76\textsuperscript{*} & 0.01\textsuperscript{*} & 0.02\textsuperscript{*} & 0.01\textsuperscript{*} & 4.51\textsuperscript{*} & 7.23\textsuperscript{*} & 5.56\textsuperscript{*} \\
  BAGEL\hspace{0.35em}\citep{BAGEL7BMoT} & 10.94\textsuperscript{*} & 52.39\textsuperscript{*} & 18.09\textsuperscript{*} & 0.31\textsuperscript{*} & 1.06\textsuperscript{*} & 0.48\textsuperscript{*} & 6.62\textsuperscript{*} & 31.67\textsuperscript{*} & 10.95\textsuperscript{*} \\
  RynnBrain\hspace{0.35em}\citep{RynnBrain2B} & 2.04 & 37.23 & 3.87 & 0.02 & 0.17 & 0.03 & 1.00 & 17.12 & 1.88 \\
  GroundAnything-VLM & 87.40 & 88.52 & 87.95 & 4.50 & 4.46 & 4.48 & 60.09 & 60.49 & 60.29 \\
  GroundAnything & 66.73 & 72.09 & 69.31 & 9.24 & 9.51 & 9.37 & 51.03 & 53.90 & 52.43 \\
  \addlinespace[2pt]
  \multicolumn{10}{@{}l}{\hspace{0.4em}\strut\textbf{Vision-Language Models (10B--1T)}} \\
  Qwen3.6-27B\hspace{0.35em}\citep{qwen3627b} & 28.56 & 53.78 & 37.31 & 1.60 & 3.45 & 2.19 & 19.05 & 35.48 & 24.78 \\
  Qwen3-VL-32B\hspace{0.35em}\citep{qwen3vl} & 29.33 & 79.24 & 42.82 & 1.25 & 1.90 & 1.51 & 18.24 & 47.65 & 26.36 \\
  Qwen3.8-27B\hspace{0.35em}\citep{qwen38} & 14.80 & 86.51 & 25.27 & 0.60 & 2.42 & 0.97 & 9.84 & 57.57 & 16.80 \\
  Qwen3.5-35B-A3B\hspace{0.35em}\citep{qwen35} & 37.48 & 80.34 & 51.12 & 1.44 & 2.31 & 1.77 & 23.36 & 49.13 & 31.65 \\
  DeepSeek-VL2-Small-16B\hspace{0.35em}\citep{Deepseekvl2} & N/A\textsuperscript{*} & N/A\textsuperscript{*} & N/A\textsuperscript{*} & N/A\textsuperscript{*} & N/A\textsuperscript{*} & N/A\textsuperscript{*} & N/A\textsuperscript{*} & N/A\textsuperscript{*} & N/A\textsuperscript{*} \\
  DeepSeek-VL2-27B\hspace{0.35em}\citep{Deepseekvl2} & N/A\textsuperscript{*} & N/A\textsuperscript{*} & N/A\textsuperscript{*} & N/A\textsuperscript{*} & N/A\textsuperscript{*} & N/A\textsuperscript{*} & N/A\textsuperscript{*} & N/A\textsuperscript{*} & N/A\textsuperscript{*} \\
  \addlinespace[2pt]
  \multicolumn{10}{@{}l}{\hspace{0.4em}\strut\textbf{Vision-Language Models (>1T)}} \\
  Qwen3.7-Max\hspace{0.35em}\citep{qwen37} & 19.91 & \textbf{90.30} & 32.62 & 0.82 & 1.92 & 1.15 & 13.31 & 57.98 & 21.62 \\
  Kimi-K2.6 & 52.17 & 55.60 & 53.83 & 2.75 & 3.06 & 2.89 & 31.33 & 33.69 & 32.47 \\
  Kimi-K3\hspace{0.35em}\citep{KimiK3} & 67.29 & 76.03 & 71.39 & 3.98 & 4.16 & 4.07 & 40.56 & 44.83 & 42.59 \\
  GPT-6 Astra & \textbf{89.83} & 87.12 & \textbf{88.48} & 8.34 & 8.03 & 8.18 & \textbf{61.83} & 60.25 & 61.04 \\
  \addlinespace[2pt]
  \bottomrule
  
  \end{NiceTabular}%
  }
  \endgroup
\end{table}

\paragraph{Dense200 with exemplars.}
GroundAnything-VLM scores 75.26 F1mIoU, while GroundAnything reaches 63.52, above Rex-Omni (55.50) and SenseNova-Vision (62.84).
GroundAnything's F1 at IoU 0.95 is 25.97 versus 12.29 for GPT-6 Astra, but its IoU-0.50 recall is lower than the AR variant's (67.13 versus 88.68).
Thus, the DLM model can localize exemplar-matched instances precisely, while recovering the full instance set remains more difficult.
\begin{table}[htbp]
  \centering
  \BenchmarkTableFont
  \caption{\textbf{Dense200 visual prompting.} Complete box-grounding metrics. LocateAnything variants lack a supported visual-prompt interface; both DeepSeek variants have incompatible output protocols. Their entries are N/A. Kimi-K3 is N/A because the required prompt format is unsupported. Starred GroundingDINO scores come from an unsupported visual-prompt task interface; starred MiMo and BAGEL scores are retained observations under suspected output-protocol incompatibility.}
  \label{tab:app-visual-dense200}
  \begingroup
  \fontsize{8}{9.7}\selectfont
  \setlength{\tabcolsep}{3pt}
  \renewcommand{\arraystretch}{1.15}
  \resizebox{\linewidth}{!}{%
  \begin{NiceTabular}{@{}>{\raggedright\arraybackslash}p{177pt}ccccccccc@{}}
  \CodeBefore
    \rowcolor{benchmarktype}{3}
    \rowcolor{benchmarktype}{5}
    \rowcolor{benchmarkpurple}{22}
    \rowcolor{benchmarkgreen}{23}
    \rowcolor{benchmarktype}{24}
    \rowcolor{benchmarktype}{31}
  \Body
  \toprule
  \textbf{Model} & \multicolumn{3}{c}{\textbf{IoU 0.50}} & \multicolumn{3}{c}{\textbf{IoU 0.95}} & \multicolumn{3}{c}{\textbf{mIoU}} \\
  \cmidrule(lr){2-4}\cmidrule(lr){5-7}\cmidrule(lr){8-10}
   & \BenchHead{R} & \BenchHead{P} & \BenchHead{F1} & \BenchHead{R} & \BenchHead{P} & \BenchHead{F1} & \BenchHead{R} & \BenchHead{P} & \BenchHead{F1} \\
  \midrule
  \multicolumn{10}{@{}l}{\hspace{0.4em}\strut\textbf{Open-set Specialized Detectors}} \\
  GroundingDINO\hspace{0.35em}\citep{groundingdino} & 2.03\textsuperscript{*} & 9.98\textsuperscript{*} & 3.38\textsuperscript{*} & 1.35\textsuperscript{*} & 5.64\textsuperscript{*} & 2.18\textsuperscript{*} & 1.90\textsuperscript{*} & 9.00\textsuperscript{*} & 3.14\textsuperscript{*} \\
  \addlinespace[2pt]
  \multicolumn{10}{@{}l}{\hspace{0.4em}\strut\textbf{Vision-Language Models (<10B)}} \\
  Rex-Omni\hspace{0.35em}\citep{rexomni} & 73.05 & 74.81 & 73.92 & 10.89 & 11.12 & 11.00 & 54.93 & 56.08 & 55.50 \\
  LocateAnything Fast\hspace{0.35em}\citep{locateanything} & N/A\textsuperscript{*} & N/A\textsuperscript{*} & N/A\textsuperscript{*} & N/A\textsuperscript{*} & N/A\textsuperscript{*} & N/A\textsuperscript{*} & N/A\textsuperscript{*} & N/A\textsuperscript{*} & N/A\textsuperscript{*} \\
  LocateAnything Hybrid\hspace{0.35em}\citep{locateanything} & N/A\textsuperscript{*} & N/A\textsuperscript{*} & N/A\textsuperscript{*} & N/A\textsuperscript{*} & N/A\textsuperscript{*} & N/A\textsuperscript{*} & N/A\textsuperscript{*} & N/A\textsuperscript{*} & N/A\textsuperscript{*} \\
  LocateAnything Slow NTP\hspace{0.35em}\citep{locateanything} & N/A\textsuperscript{*} & N/A\textsuperscript{*} & N/A\textsuperscript{*} & N/A\textsuperscript{*} & N/A\textsuperscript{*} & N/A\textsuperscript{*} & N/A\textsuperscript{*} & N/A\textsuperscript{*} & N/A\textsuperscript{*} \\
  Qwen2.5-VL-7B\hspace{0.35em}\citep{qwen25vl} & 0.75 & 23.66 & 1.46 & 0.00 & 0.00 & 0.00 & 0.44 & 13.33 & 0.86 \\
  Qwen3-VL-2B\hspace{0.35em}\citep{qwen3vl} & 7.76 & 60.82 & 13.76 & 0.69 & 1.78 & 1.00 & 5.08 & 33.85 & 8.79 \\
  Qwen3-VL-4B\hspace{0.35em}\citep{qwen3vl} & 1.76 & \textbf{93.00} & 3.45 & 0.21 & 8.00 & 0.41 & 1.27 & 62.90 & 2.49 \\
  Qwen3-VL-8B\hspace{0.35em}\citep{qwen3vl} & 8.26 & 87.71 & 15.10 & 0.99 & 3.91 & 1.58 & 5.79 & 55.05 & 10.45 \\
  Qwen3.5-4B\hspace{0.35em}\citep{qwen35} & 2.82 & 78.69 & 5.44 & 0.34 & 7.71 & 0.65 & 2.01 & 54.25 & 3.87 \\
  Qwen3.5-9B\hspace{0.35em}\citep{qwen35} & 26.66 & 65.61 & 37.91 & 4.19 & 10.33 & 5.96 & 19.00 & 46.27 & 26.93 \\
  RynnBrain1.1\hspace{0.35em}\citep{RynnBrain112B} & 1.16 & 62.50 & 2.28 & 0.07 & 4.00 & 0.14 & 0.74 & 38.25 & 1.45 \\
  SenseNova-Vision\hspace{0.35em}\citep{SenseNovaVision7BMoT} & 69.56 & 84.68 & 76.38 & 19.15 & 22.06 & 20.50 & 57.21 & 69.70 & 62.84 \\
  MiMo-VL-7B-SFT\hspace{0.35em}\citep{MimoVL} & 2.87\textsuperscript{*} & 20.40\textsuperscript{*} & 5.03\textsuperscript{*} & 0.01\textsuperscript{*} & 0.00\textsuperscript{*} & 0.00\textsuperscript{*} & 1.08\textsuperscript{*} & 6.38\textsuperscript{*} & 1.83\textsuperscript{*} \\
  MiMo-VL-7B-RL\hspace{0.35em}\citep{MimoVL} & 2.58\textsuperscript{*} & 14.41\textsuperscript{*} & 4.38\textsuperscript{*} & 0.01\textsuperscript{*} & 0.01\textsuperscript{*} & 0.01\textsuperscript{*} & 0.89\textsuperscript{*} & 5.40\textsuperscript{*} & 1.51\textsuperscript{*} \\
  BAGEL\hspace{0.35em}\citep{BAGEL7BMoT} & 1.28\textsuperscript{*} & 56.29\textsuperscript{*} & 2.50\textsuperscript{*} & 0.02\textsuperscript{*} & 1.25\textsuperscript{*} & 0.05\textsuperscript{*} & 0.74\textsuperscript{*} & 29.87\textsuperscript{*} & 1.45\textsuperscript{*} \\
  RynnBrain\hspace{0.35em}\citep{RynnBrain2B} & 1.26 & 64.50 & 2.47 & 0.02 & 1.00 & 0.03 & 0.74 & 36.35 & 1.44 \\
  GroundAnything-VLM & 88.68 & 92.85 & 90.72 & \textbf{28.51} & \textbf{30.01} & \textbf{29.24} & \textbf{73.59} & \textbf{77.00} & \textbf{75.26} \\
  GroundAnything & 67.13 & 79.38 & 72.74 & 24.97 & 27.06 & 25.97 & 58.16 & 69.96 & 63.52 \\
  \addlinespace[2pt]
  \multicolumn{10}{@{}l}{\hspace{0.4em}\strut\textbf{Vision-Language Models (10B--1T)}} \\
  Qwen3.6-27B\hspace{0.35em}\citep{qwen3627b} & 31.92 & 59.87 & 41.64 & 5.79 & 7.69 & 6.61 & 23.81 & 45.89 & 31.33 \\
  Qwen3-VL-32B\hspace{0.35em}\citep{qwen3vl} & 18.96 & 55.73 & 28.30 & 2.51 & 6.05 & 3.55 & 13.84 & 41.11 & 20.69 \\
  Qwen3.8-27B\hspace{0.35em}\citep{qwen38} & 25.08 & 68.06 & 36.65 & 4.61 & 14.62 & 7.01 & 19.15 & 53.40 & 28.19 \\
  Qwen3.5-35B-A3B\hspace{0.35em}\citep{qwen35} & 41.09 & 71.35 & 52.15 & 5.85 & 7.14 & 6.43 & 29.95 & 51.04 & 37.73 \\
  DeepSeek-VL2-Small-16B\hspace{0.35em}\citep{Deepseekvl2} & N/A\textsuperscript{*} & N/A\textsuperscript{*} & N/A\textsuperscript{*} & N/A\textsuperscript{*} & N/A\textsuperscript{*} & N/A\textsuperscript{*} & N/A\textsuperscript{*} & N/A\textsuperscript{*} & N/A\textsuperscript{*} \\
  DeepSeek-VL2-27B\hspace{0.35em}\citep{Deepseekvl2} & N/A\textsuperscript{*} & N/A\textsuperscript{*} & N/A\textsuperscript{*} & N/A\textsuperscript{*} & N/A\textsuperscript{*} & N/A\textsuperscript{*} & N/A\textsuperscript{*} & N/A\textsuperscript{*} & N/A\textsuperscript{*} \\
  \addlinespace[2pt]
  \multicolumn{10}{@{}l}{\hspace{0.4em}\strut\textbf{Vision-Language Models (>1T)}} \\
  Qwen3.7-Max\hspace{0.35em}\citep{qwen37} & 28.05 & 72.46 & 40.44 & 5.40 & 9.51 & 6.89 & 22.37 & 55.16 & 31.81 \\
  Kimi-K2.6 & 43.14 & 51.19 & 46.82 & 5.48 & 6.84 & 6.09 & 29.90 & 35.39 & 32.41 \\
  Kimi-K3\hspace{0.35em}\citep{KimiK3} & N/A\textsuperscript{*} & N/A\textsuperscript{*} & N/A\textsuperscript{*} & N/A\textsuperscript{*} & N/A\textsuperscript{*} & N/A\textsuperscript{*} & N/A\textsuperscript{*} & N/A\textsuperscript{*} & N/A\textsuperscript{*} \\
  GPT-6 Astra & \textbf{95.44} & 88.71 & \textbf{92.04} & 12.84 & 11.75 & 12.29 & 71.48 & 66.44 & 68.94 \\
  \addlinespace[2pt]
  \bottomrule
  
  \end{NiceTabular}%
  }
  \endgroup
\end{table}

\subsection{Potential Application Scenarios}
\label{app:applications}

The following prospective workflows combine changing queries with substantial spatial output; they are opportunities for evaluation rather than demonstrated deployments.

\paragraph{High-throughput industrial inspection.}
Language or exemplar queries can specify defects, components, assembly checks, and labels across changing product lines.
Parallel output is particularly relevant when each image requires many localized results.

\paragraph{Embodied and driving data annotation.}
Batch processing of robot views, manipulation keyframes, and road images can produce candidate object, part, text, and relation-conditioned spatial annotations.
Lower decoding cost could expand annotation and review capacity as images, queries, and instances multiply.

\paragraph{Dense remote sensing.}
Image tiles of ports, car parks, airports, and urban areas require locating many queried instances.
Open queries and long coordinate lists make this a useful setting for evaluating parallel spatial extraction.

\paragraph{Batch OCR and document processing.}
Receipts, forms, archives, and complex pages require text together with region locations.
Long transcriptions and multi-region outputs make this a practical test of decoding savings at matched extraction quality.

\paragraph{Medical and microscopy annotation.}
Candidate boxes or points for cells, nuclei, and repeated structures could assist research counting and expert review.
Domain-specific reliability requires separate evaluation before using these annotations in specialist workflows.

\paragraph{Sports and video analysis.}
Sampled frames can be queried for players, balls, officials, and jersey numbers, including appearance or spatial conditions.
Cheaper frame-level extraction could support denser sampling; temporal association remains a downstream task.

\paragraph{Retail inventory and agricultural counting.}
Crowded shelves, logistics bins, orchards, and nurseries combine repeated instances with changing appearance or region queries.
Exemplar conditioning provides a useful interface when category names alone are insufficient.

\paragraph{Interactive annotation and GUI grounding.}
Users can revise descriptions and confirm candidate boxes, points, or click locations through repeated queries.
Short responses should be assessed by end-to-end interaction latency, including visual processing and prefill.

\subsection{Qualitative Analysis}
\label{app:qualitative_analysis}

We examine selected visualizations across grounding tasks, emphasizing the spatial and semantic demands visible in each example. Source images and their displayed annotations are preserved. These examples provide qualitative context, not additional estimates of accuracy or recall. A panel marked N/A denotes an unavailable comparison.

\subsubsection{General Object Grounding}

\begin{figure}[htbp]
\centering
\includegraphics[width=\linewidth]{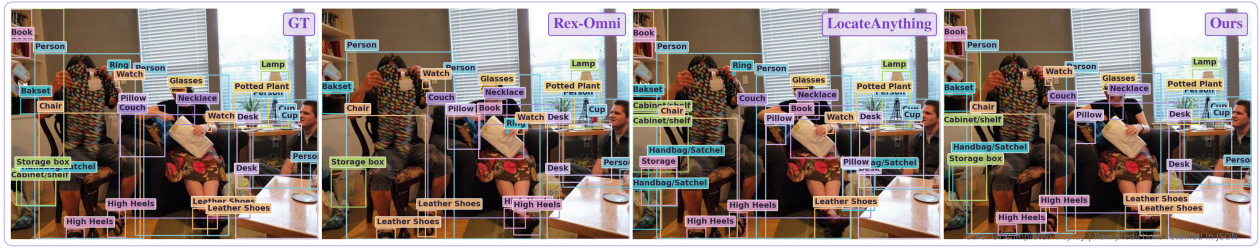}
\caption{Multi-scale object grounding in a cluttered indoor scene.}
\label{fig:qual-01}
\end{figure}

\paragraph{Grounding in a cluttered indoor scene.} The indoor scene in \Cref{fig:qual-01} combines people, furniture, books, containers, and small accessories. Large objects provide scene context, while partially visible items and accessories require finer spatial discrimination. The displayed annotations illustrate the need to retain instance identity across scale and occlusion, rather than replacing a collection of nearby objects with one broad region. In particular, distinguishing an accessory from its wearer requires semantic association as well as localization.

\subsubsection{Dense Object Grounding}

\begin{figure}[htbp]
\centering
\includegraphics[width=\linewidth]{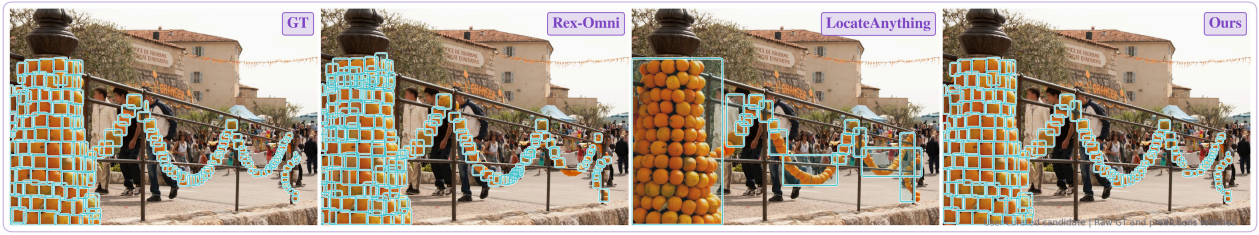}
\caption{Dense instance grounding of fruit decorations.}
\label{fig:qual-03}
\end{figure}

\paragraph{Dense instances under occlusion.} The repeated fruit decorations in \Cref{fig:qual-03} vary in apparent size and overlap along a central support and surrounding garlands. The comparison illustrates a distinction between localizing individual fruits and enclosing an entire decorated structure. Several broad boxes in the comparison panels merge multiple instances, whereas the displayed Ours panel retains finer instance granularity. Adjacent fruits with similar color make both duplicate suppression and boundary separation difficult. The curated visualization illustrates these error modes without assigning additional detection scores.

\begin{figure}[htbp]
\centering
\includegraphics[width=\linewidth]{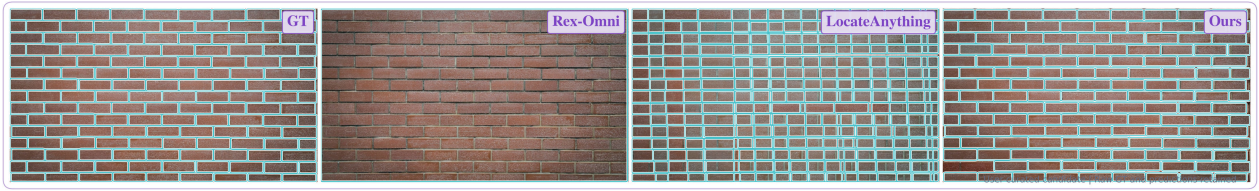}
\caption{Dense grounding of repeated bricks.}
\label{fig:qual-04}
\end{figure}

\paragraph{Repeated boundaries and fine spatial structure.} \Cref{fig:qual-04} uses a brick wall to probe repeated structures with small appearance differences. The displayed Ours boxes follow individual bricks and preserve the staggered rows, whereas crossing or oversized boxes in the comparison panels mix neighboring units. Success here requires aligning the queried unit with local mortar boundaries while maintaining consistency across a large set of nearly interchangeable instances. The figure is a qualitative illustration of spatial granularity; an unannotated comparison panel alone does not identify the underlying cause of a missing display.

\subsubsection{Referring Grounding and Complex Visual Configurations}

\begin{figure}[htbp]
\centering
\includegraphics[width=\linewidth]{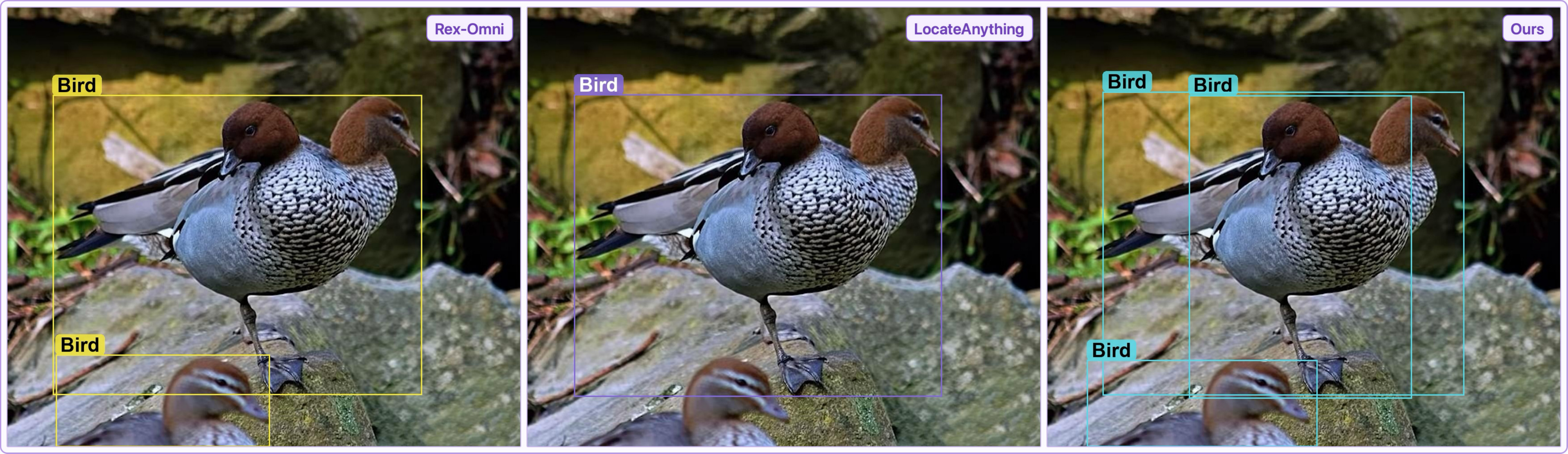}
\caption{Separating overlapping birds with deceptive silhouettes.}
\label{fig:qual-06}
\end{figure}

\paragraph{Complex case: visually deceptive overlap.} \Cref{fig:qual-06} presents a visually deceptive configuration: two birds overlap so closely that their similar feather textures and aligned body contours can resemble a single two-headed bird. A third bird is only partly visible at the lower image boundary. The Rex-Omni panel encloses the overlapping pair in one large box and marks the foreground bird separately; LocateAnything shows one broad box; the Ours panel distinguishes the two overlapping instances and the truncated foreground instance. The relevant evidence includes two distinct heads and beaks, differently oriented necks, and the relationship between each head and its body contour. A purely salient-region interpretation can merge those cues into one object. Resolving the ambiguity calls for instance-level semantic understanding, occlusion reasoning, and spatial part association, motivating a strong vision--language foundation rather than localization based on isolated texture or outline alone. This example illustrates a demanding capability; a single selected case does not establish a particular internal reasoning mechanism or a universal advantage over all competing models.

\begin{figure}[htbp]
\centering
\includegraphics[width=\linewidth]{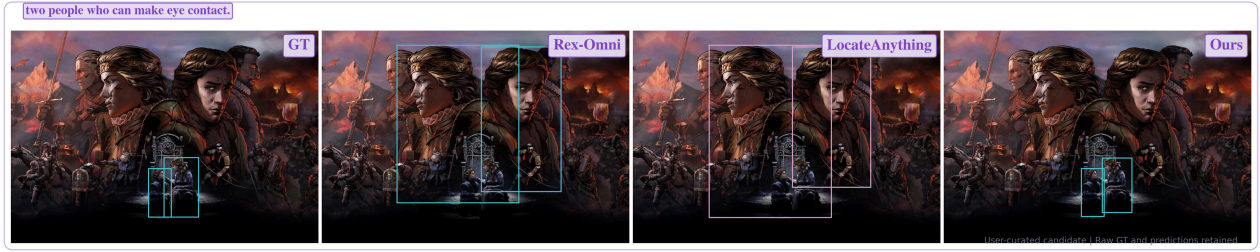}
\caption{Grounding a pair specified by a visual relationship.}
\label{fig:qual-07}
\end{figure}

\paragraph{Relational referring beyond visual salience.} The query in \Cref{fig:qual-07} asks for two people who can make eye contact. The relevant pair occupies a small area near the bottom of a poster dominated by much larger portraits. The displayed Ours regions select that pair, while other panels emphasize the larger faces. The challenge is to interpret a relation between two instances and their orientation, rather than rank people by size or salience. This example concerns the visible relational configuration, not an inference about the depicted people's identity or mental state.

\begin{figure}[htbp]
\centering
\includegraphics[width=\linewidth]{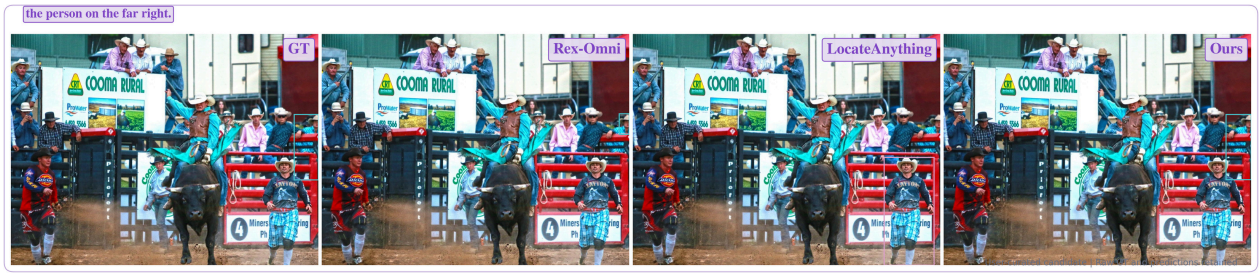}
\caption{Grounding a spatially specified, partially visible person.}
\label{fig:qual-08}
\end{figure}

\paragraph{Referring at the image boundary.} In \Cref{fig:qual-08}, the phrase ``the person on the far right'' refers to a partially truncated spectator at the image boundary, rather than the prominent foreground participant. The scene contains many people with similar hats and clothing, making category recognition alone insufficient. The displayed Ours box follows the boundary instance. The example illustrates the need to resolve a relative spatial expression over all visible candidates, including small and partly occluded instances, before predicting the target extent.

\subsubsection{Referring Point-in-Mask Grounding}

\begin{figure}[htbp]
\centering
\includegraphics[width=\linewidth]{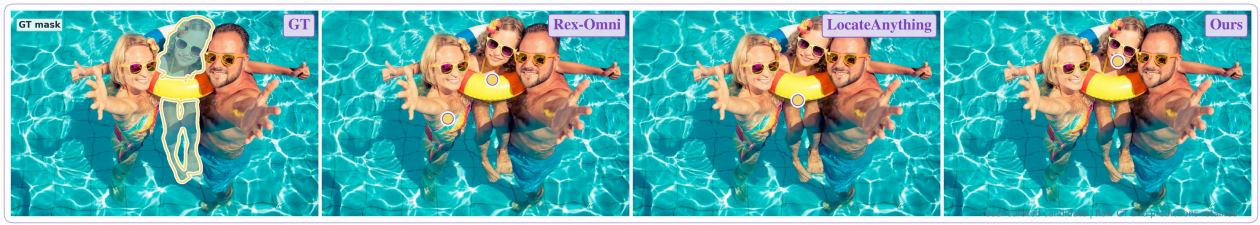}
\caption{Point grounding within an occluded person's visible mask.}
\label{fig:qual-10}
\end{figure}

\paragraph{Referring points inside visible target regions.} \Cref{fig:qual-10} separates semantic target selection from the geometric requirement that a point lie inside the target's visible mask. The child is partially hidden by a float, so the enclosing box center can fall on an occluder. The displayed Ours point lies on the visible face, while the comparison points fall on the float or another person. A valid point therefore requires both identifying the intended instance and selecting visible evidence belonging to it, rather than using a generic scene center or box-center heuristic.

\subsubsection{Dense Point Grounding}

\begin{figure}[htbp]
\centering
\includegraphics[width=\linewidth]{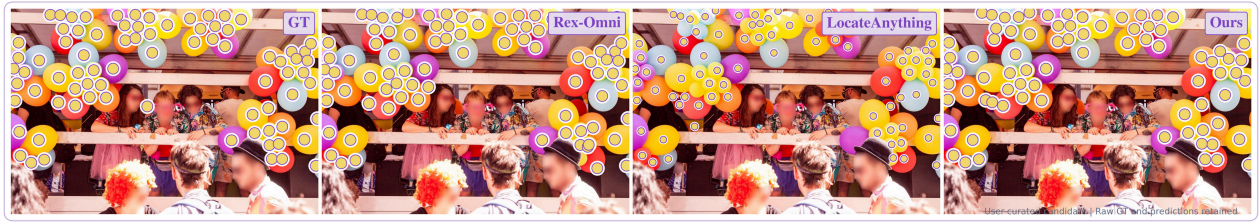}
\caption{Dense point grounding of overlapping balloons.}
\label{fig:qual-12}
\end{figure}

\paragraph{Dense pointing with repeated appearances.} \Cref{fig:qual-12} contains many balloons with repeated colors, touching outlines, partial occlusions, and instances cut by the image boundary. The displayed points span the arch rather than collapsing the group to a few salient locations. The task combines set coverage with one-point-per-instance consistency: missing small balloons and placing repeated points on the same balloon are distinct errors. Marker size is part of the visualization and should not be interpreted as localization uncertainty or predicted object size.

\subsubsection{Dense Point-in-Mask Grounding}

\begin{figure}[htbp]
\centering
\includegraphics[width=\linewidth]{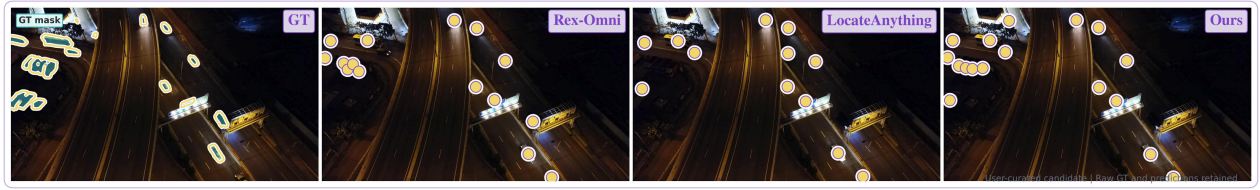}
\caption{Dense point grounding of vehicles in a night scene.}
\label{fig:qual-14}
\end{figure}

\paragraph{Dense point placement in low light.} \Cref{fig:qual-14} shows vehicles from above under low illumination, including moving vehicles on the road and a compact parked group. Ground-truth masks identify the visible vehicle regions, while point predictions must remain within those regions despite shadows and bright headlights. The displayed Ours points cover both isolated and clustered vehicles. The example illustrates why coverage and interior-point placement must be evaluated together: a point on a light streak or adjacent road surface can be close to a vehicle while missing its visible mask.

\subsubsection{GUI Grounding}

\begin{figure}[htbp]
\centering
\includegraphics[width=\linewidth]{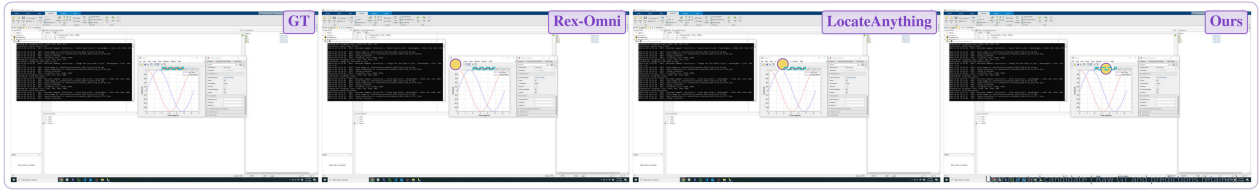}
\caption{GUI target grounding in a multi-window desktop.}
\label{fig:qual-15}
\end{figure}

\paragraph{GUI grounding among overlapping windows.} The desktop in \Cref{fig:qual-15} contains an editor, a terminal, a plot window, and a property inspector. The marked reference region is the small plot-title area, surrounded by nearby menu and toolbar controls. The displayed Ours point lies in that region, whereas the comparison points are displaced toward neighboring controls. This case requires assigning a target to the correct window and interpreting local interface structure. It illustrates visual target grounding, not successful execution of an unshown action sequence.

\subsubsection{OCR and Artistic Text Understanding}

\begin{figure}[htbp]
\centering
\includegraphics[width=\linewidth]{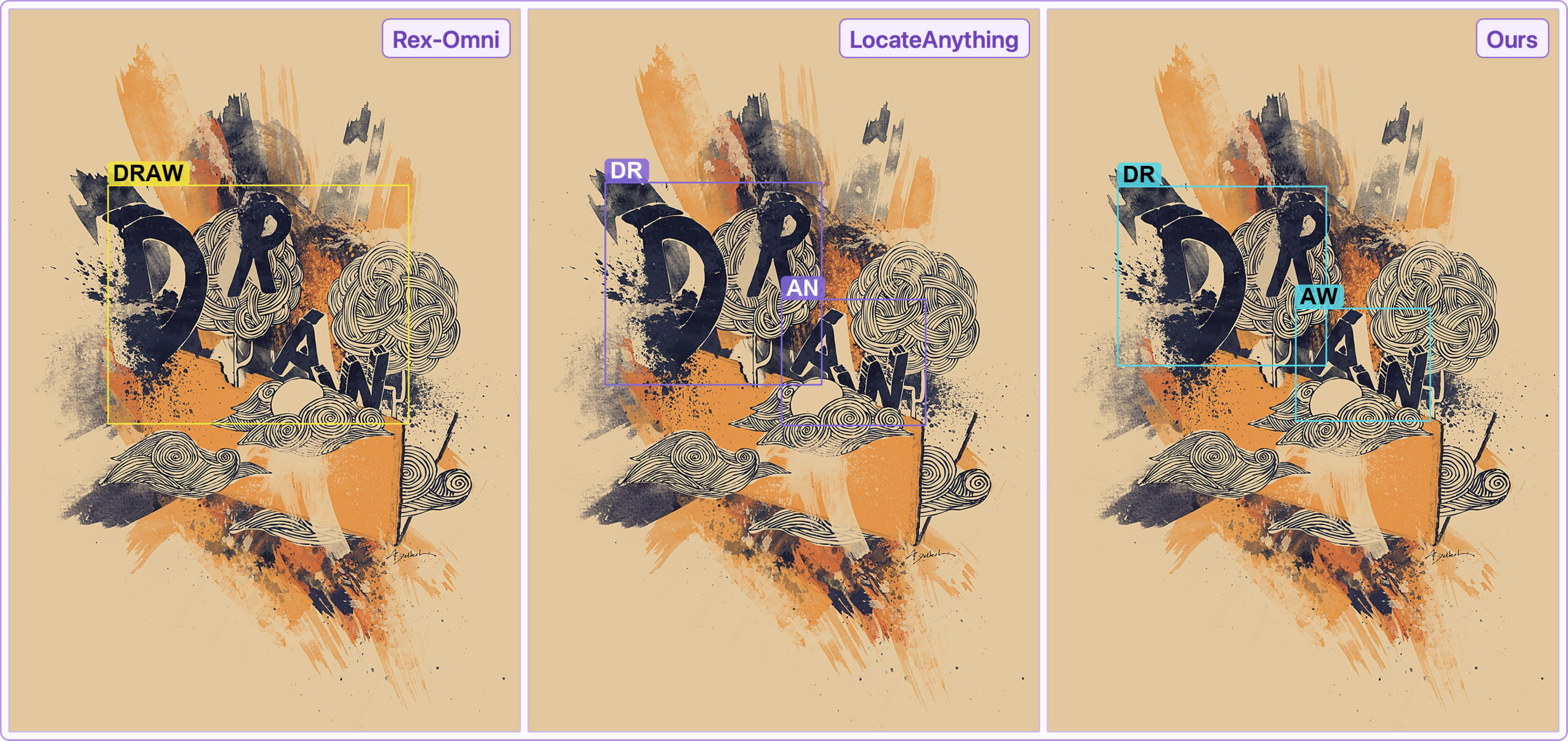}
\caption{OCR of artistic lettering embedded in illustration.}
\label{fig:qual-17}
\end{figure}

\paragraph{Complex case: artistic letters and decorative texture.} The word ``DRAW'' in \Cref{fig:qual-17} is distributed across staggered, tilted letters embedded in brush strokes and dense line art. Decorative contours resemble character strokes, while the letters do not share a conventional horizontal baseline. The Ours panel reads the two spatial groups as ``DR'' and ``AW''; LocateAnything shows ``DR'' and ``AN'', and Rex-Omni uses a single ``DRAW'' region. This distinction matters: a complete word box and two correct fragments can both be reasonable annotation granularities, so the split alone is not evidence of superior recognition. The informative feature is maintaining the correct letter identity and its spatial support despite stylization, especially the final W. Such reading requires integrating local strokes with word-level context and distinguishing typography from illustration. It motivates strong visual--linguistic representations while also showing why OCR comparisons must specify both transcription and grouping conventions.

\begin{figure}[htbp]
\centering
\includegraphics[width=\linewidth]{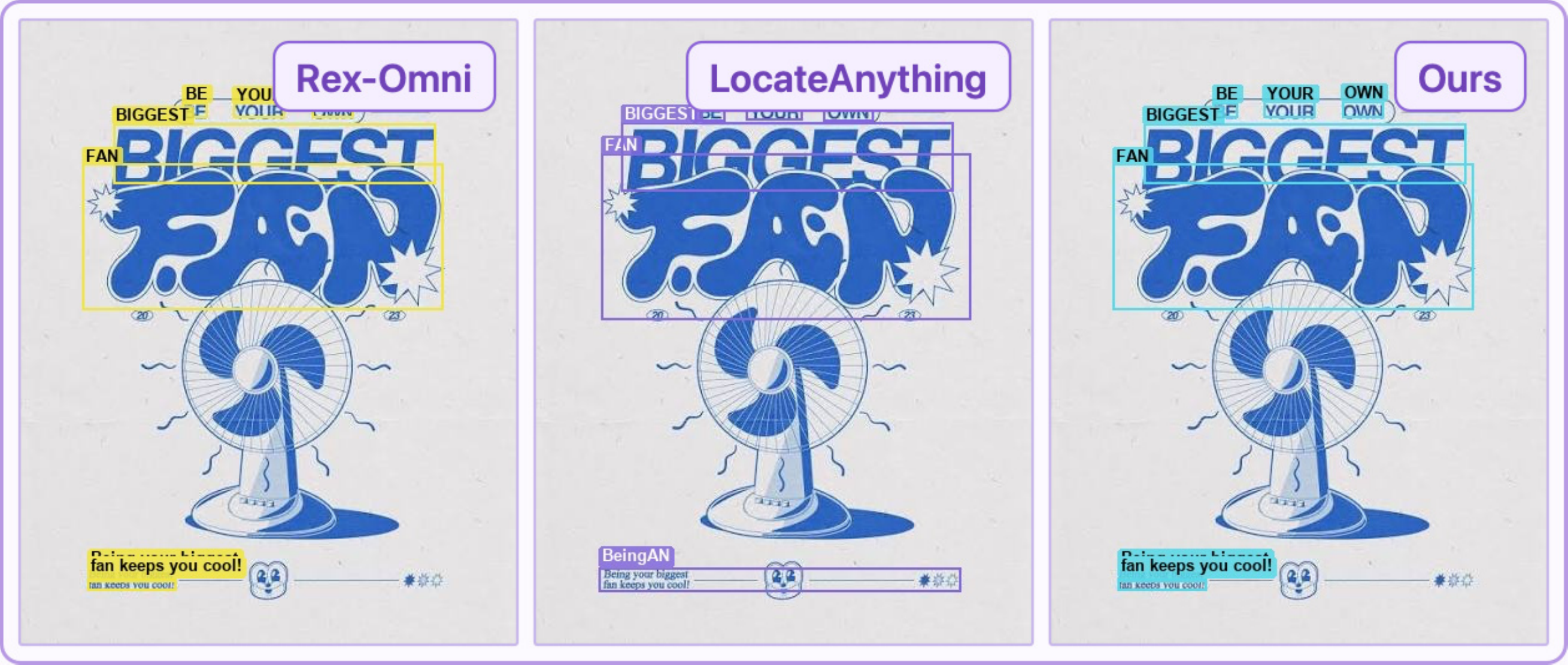}
\caption{OCR of highly stylized poster lettering.}
\label{fig:qual-18}
\end{figure}

\paragraph{Complex case: shape-based typography.} In \Cref{fig:qual-18}, the poster combines small conventional words with the oversized, heavily deformed word ``FAN'' and an illustrated electric fan. Rounded, interlocking glyphs and decorative stars blur the boundary between text and drawing. The Ours panel preserves the large ``FAN'' region while also identifying smaller words such as ``BE'', ``YOUR'', and ``OWN''. Reading the slogan requires tracking character order across sizes and styles rather than treating all blue shapes as one object. The fan illustration supplies useful semantic context, but a grounded transcription must still be supported by visible letter strokes. This is a demanding vision--language reading example; it does not imply that contextual plausibility can substitute for character evidence or that every small printed line is recovered perfectly.

\begin{figure}[htbp]
\centering
\includegraphics[width=\linewidth]{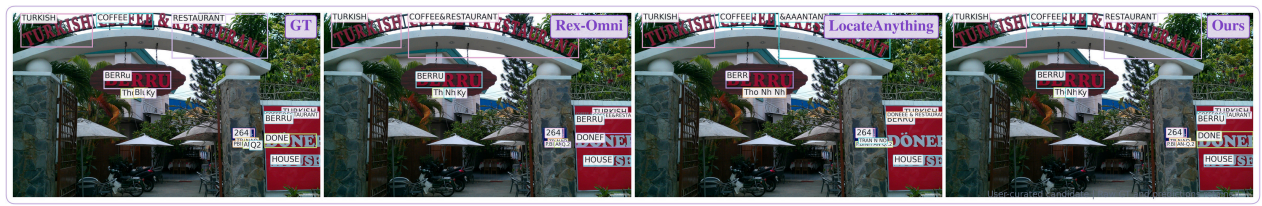}
\caption{OCR of outdoor signs with perspective distortion.}
\label{fig:qual-21}
\end{figure}

\paragraph{Scene text with perspective and scale variation.} \Cref{fig:qual-21} includes large lettering along a curved entrance, a hanging business sign, and smaller roadside text. Perspective changes character orientation and spacing, while foliage and uneven contrast complicate localization. The displayed Ours regions retain separate text groups across these scales; comparison panels illustrate merged phrases and transcription changes. The case highlights joint text--geometry consistency: a plausible phrase in a region that spans multiple signs is different from a transcription aligned to the intended word or line.

\begin{figure}[htbp]
\centering
\includegraphics[width=\linewidth]{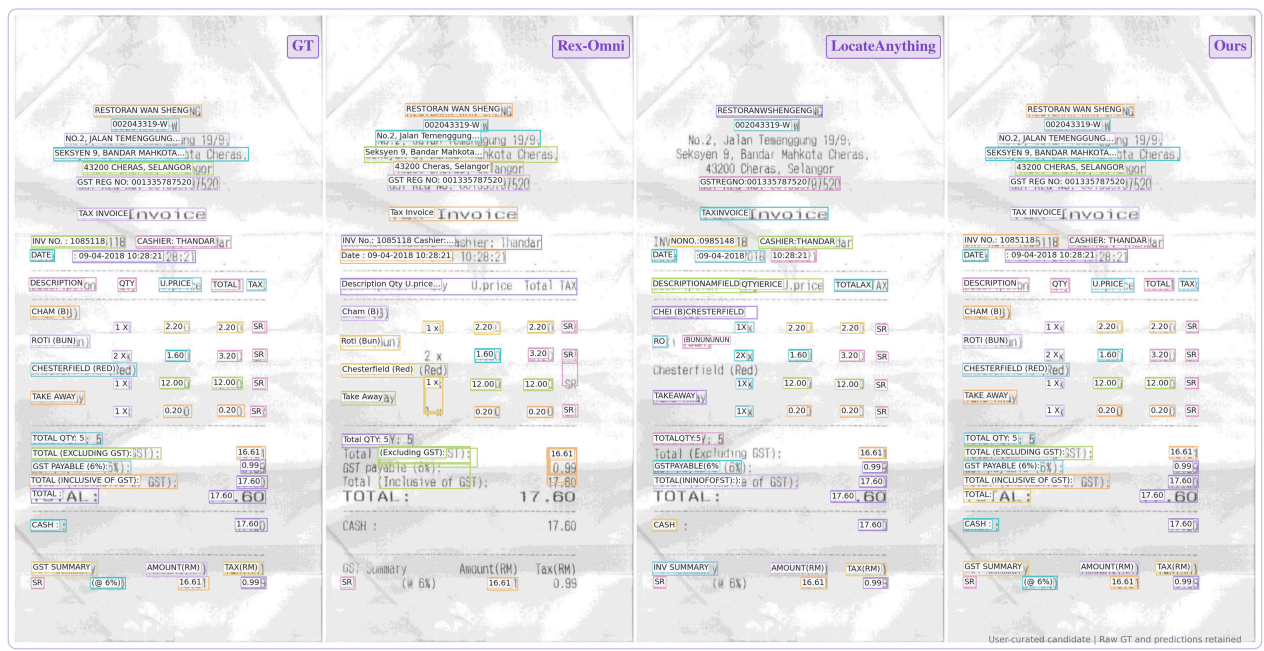}
\caption{OCR of a low-contrast receipt with aligned fields.}
\label{fig:qual-22}
\end{figure}

\begin{figure}[htbp]
\centering
\includegraphics[width=\linewidth]{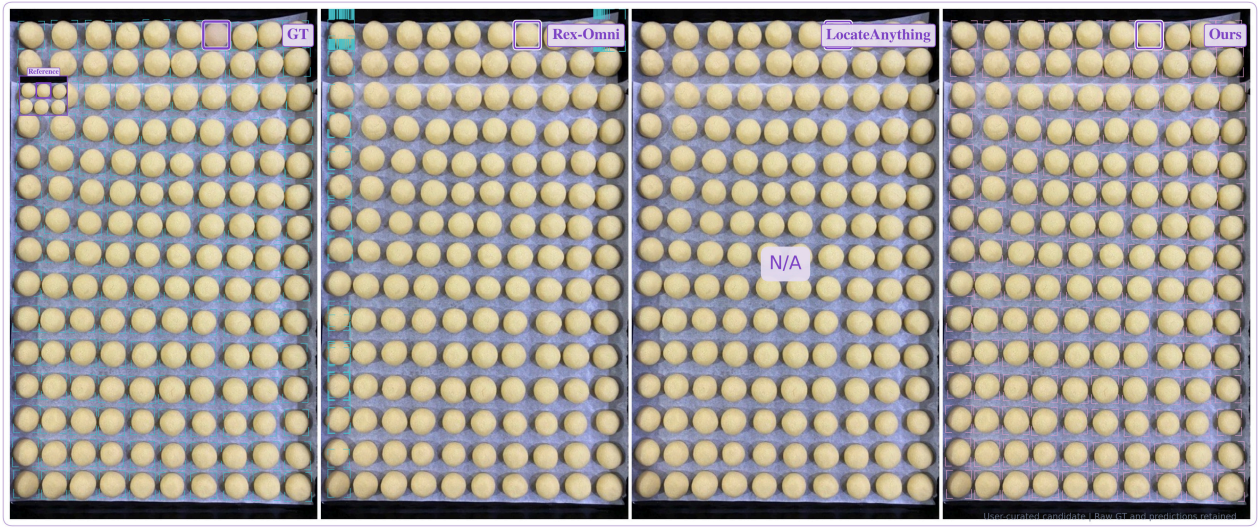}
\caption{Exemplar-guided grounding of repeated rounded objects.}
\label{fig:qual-24}
\end{figure}

\begin{figure}[htbp]
\centering
\includegraphics[width=\linewidth]{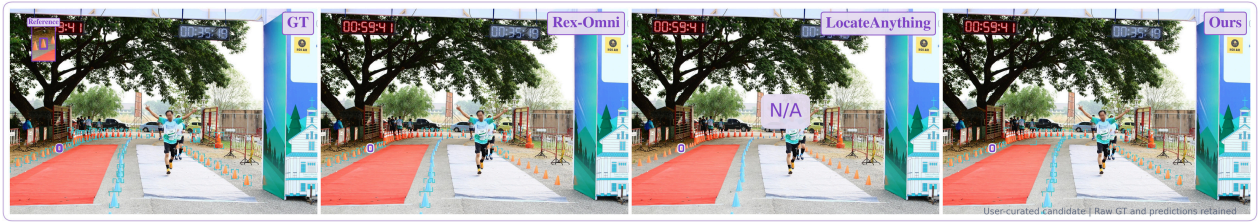}
\caption{Exemplar-guided grounding of traffic cones.}
\label{fig:qual-25}
\end{figure}

\paragraph{Document text and tabular alignment.} The faded receipt in \Cref{fig:qual-22} contains headings, item descriptions, quantities, unit prices, totals, and tax fields. The displayed Ours regions preserve separate text cells and repeated numerical entries across the document, while comparison panels illustrate merged headers and missed or fragmented lines. Useful OCR must retain both character identity and the layout that associates each value with its row and column. The visualization concerns recognition and localization; it does not establish arithmetic verification or downstream financial correctness.

\subsubsection{Visual-Prompt Grounding}

\paragraph{Visual exemplars in highly repetitive scenes.} In \Cref{fig:qual-24}, a boxed visual exemplar specifies the repeated rounded objects without relying on a category name. The target instances form dense rows with mild shape and size variation. The displayed Ours boxes cover instances across the tray, while the Rex-Omni panel contains sparse and repeated edge-aligned proposals. The LocateAnything panel is explicitly marked N/A and is not treated as a scored failure. The example illustrates transferring an exemplar's visual identity to many instances while avoiding duplicate localization.

\paragraph{Visual prompting with perspective variation.} \Cref{fig:qual-25} uses a local visual reference to specify cones distributed around a running course. Their apparent size changes strongly with depth, and the scene includes both orange and blue cones. The displayed Ours boxes follow instances from the foreground toward the distant gate, illustrating visual category transfer beyond the exact reference crop. The comparison highlights partial coverage and repeated localization. The panel marked N/A denotes an unavailable comparison, not an observed zero score.

\end{document}